\documentclass[12pt]{article}

\usepackage{PRIMEarxiv}

\usepackage[utf8]{inputenc} %
\usepackage[T1]{fontenc}    %
\usepackage{url}            %
\usepackage{booktabs}       %
\usepackage{amsfonts}       %
\usepackage{nicefrac}       %
\usepackage{microtype}      %
\usepackage{lipsum}
\usepackage{fancyhdr}       %
\usepackage{graphicx}       %
\graphicspath{{media/}}     %
\usepackage{etoolbox}       %
\usepackage{makecell}  %
\usepackage{times}
\usepackage{soul}
\usepackage{url}
\usepackage[hidelinks]{hyperref}
\usepackage[table]{xcolor}
\usepackage{booktabs}
\usepackage[utf8]{inputenc}
\usepackage[small]{caption}
\usepackage{graphicx}
\usepackage{amsmath}    
\usepackage{amssymb}
\usepackage{amsthm}
\usepackage{wasysym}
\usepackage{booktabs}
\usepackage{algorithm}
\usepackage{algorithmic}
\usepackage[switch]{lineno}
\usepackage{xcolor}
\usepackage{listings}
\usepackage{tikz}
\usepackage{xparse}
\usepackage{enumitem}
\usepackage{fontawesome5}
\definecolor{mygreen}{RGB}{30, 128, 20}
\colorlet{shadecolor}{gray!20}

\tikzset{chatstyle/.style={text width=2.8in,rounded corners=2pt}}

\newtheorem{definition}{Definition}[section]

\definecolor{mygreen}{HTML}{88EABB}
\definecolor{OliveGreen}{HTML}{00693E}

\usepackage{wrapfig}
\usepackage{adjustbox}
\usepackage{xspace}
\usepackage{listings}
\usepackage{subcaption}
\usepackage[most]{tcolorbox}

\usepackage{pifont}

\usepackage{multicol}
\usepackage{multirow}
\usepackage{bbding}
\usepackage{circledsteps}

\usepackage{fancyhdr}
\usetikzlibrary{mindmap,shadows}

\definecolor{LightCyan}{RGB}{232,241,255}
\definecolor{LightRed}{RGB}{255,235,235}
\definecolor{LightPink}{RGB}{255,235,255}
\definecolor{LightGreen}{RGB}{218,255,234}
\definecolor{LightYellow}{RGB}{255,255,235}
\definecolor{LightGray}{RGB}{242,242,242}
\definecolor{Red}{RGB}{253, 239, 242}
\definecolor{Yellow}{RGB}{255, 255, 204}
\definecolor{Pink}{RGB}{255, 243, 254}
\definecolor{Gray}{RGB}{249, 249, 249}
\definecolor{Green}{RGB}{230, 255, 241}
\definecolor{Blue1}{RGB}{218, 232, 245}
\definecolor{Blue2}{RGB}{239, 248, 253}
\definecolor{Blue3}{RGB}{136, 190, 220}
\definecolor{Blue4}{RGB}{83, 157, 204}
\definecolor{Blue5}{RGB}{42, 122, 185}
\definecolor{Blue6}{RGB}{11, 85, 159}
\definecolor{GreenCheck}{RGB}{0, 102, 51}
\definecolor{LightBack}{RGB}{247,249,251}
\tcbset{
    colback=LightBack,colframe=black,boxsep=3pt, arc=3pt,boxrule=1pt,
    width=\linewidth,enhanced,frame hidden, overlay={\draw[line width=1pt] (frame.north west)--([xshift=\linewidth]frame.north west);\draw[line width=1pt] (frame.south west)--([xshift=\linewidth]frame.south west);}
}

\usepackage{amsmath,amsfonts,bm}

\def\eqref#1{equation~\ref{#1}}

\def\1{\bm{1}}

\def\vo{{\bm{o}}}

\DeclareMathAlphabet{\mathsfit}{\encodingdefault}{\sfdefault}{m}{sl}
\SetMathAlphabet{\mathsfit}{bold}{\encodingdefault}{\sfdefault}{bx}{n}

\def\gC{{\mathcal{C}}}
\def\gD{{\mathcal{D}}}
\def\gE{{\mathcal{E}}}

\def\gG{{\mathcal{G}}}

\def\gL{{\mathcal{L}}}

\def\gV{{\mathcal{V}}}

\def\gX{{\mathcal{X}}}
\def\gY{{\mathcal{Y}}}

\DeclareMathOperator*{\argmax}{arg\,max}
\DeclareMathOperator*{\argmin}{arg\,min}

\input{mydef.sty}

\newcommand{\squishlist}{
\begin{list}{{{\small{$\bullet$}}}}
{\setlength{\itemsep}{1pt}      \setlength{\parsep}{5pt}
\setlength{\topsep}{-2pt}       \setlength{\partopsep}{0pt}
\setlength{\leftmargin}{2.5em} \setlength{\labelwidth}{1em}
\setlength{\labelsep}{1em} } }

\newcommand{\squishend}{  \end{list}  }
\definecolor{aigold}{RGB}{244,210, 1} 
\definecolor{aigreen}{RGB}{210,244,211} 

\sethlcolor{aigreen}
\definecolor{aired}{RGB}{255,180,181} 
\definecolor{lighterseafoam}{RGB}{194,218,184}

\usepackage{datetime}
\usepackage{CJKutf8}
\title{Graph Foundation Models: A Comprehensive Survey}

\author{
{\bfseries Zehong Wang$^{\dagger,*,\circledast,1}$}\quad
{\bfseries Zheyuan Liu$^{*,1}$}\quad
{\bfseries Tianyi Ma$^{*,1}$}\quad
{\bfseries Jiazheng Li$^{*,2}$}\quad
{\bfseries Zheyuan Zhang$^{*,1}$}\quad
\\
{\bfseries Xingbo Fu$^{*,3}$}\quad
{\bfseries Yiyang Li$^{*,1}$}\quad
{\bfseries Zhengqing Yuan$^{*,1}$}\quad
{\bfseries Wei Song$^{1}$}\quad
{\bfseries Yijun Ma$^{1}$}\quad
{\bfseries Qingkai Zeng$^{1}$}\quad
\\
{\bfseries Xiusi Chen$^{4}$}\quad
{\bfseries Jianan Zhao$^{6,7}$}\quad
{\bfseries Jundong Li$^{3}$}\quad
{\bfseries Meng Jiang$^{1}$}\quad 
{\bfseries Pietro Li\`o$^{5}$}\quad
\\
{\bfseries Nitesh Chawla$^{1}$}\quad
{\bfseries Chuxu Zhang$^{2}$}\quad
{\bfseries Yanfang Ye$^{\circledast,1}$}
\\
\\
{\bfseries $^\dagger$Project Leader \quad $^{*}$Major Student Contributors}
\\
{\bfseries $^\circledast$Correspondance}: Zehong Wang <\texttt{zwang43@nd.edu}>,  Yanfang Ye <\texttt{yye7@nd.edu}>
\\
\\
$^1$University of Notre Dame,
$^2$University of Connecticut,
$^3$University of Virginia,\\
$^4$University of Illinois Urbana-Champaign,
$^5$University of Cambridge, \\
$^6$Mila - Qu\'ebec AI Institute,
$^7$Universit\'e de Montr\'eal
}

\begin{document}
\maketitle

\begin{abstract}

Graph-structured data pervades domains such as social networks, biological systems, knowledge graphs, and recommender systems. While foundation models have transformed natural language processing, vision, and multimodal learning through large-scale pretraining and generalization, extending these capabilities to graphs—characterized by non-Euclidean structures and complex relational semantics—poses unique challenges and opens new opportunities. To this end, \textit{Graph Foundation Models} (GFMs) aim to bring scalable, general-purpose intelligence to structured data, enabling broad transfer across graph-centric tasks and domains. This survey provides a comprehensive overview of GFMs, unifying diverse efforts under a modular framework comprising three key components: backbone architectures, pretraining strategies, and adaptation mechanisms. We categorize GFMs by their generalization scope—universal, task-specific, and domain-specific—and review representative methods, key innovations, and theoretical insights within each category. Beyond methodology, we examine theoretical foundations including transferability and emergent capabilities, and highlight key challenges such as structural alignment, heterogeneity, scalability, and evaluation. Positioned at the intersection of graph learning and general-purpose AI, GFMs are poised to become foundational infrastructure for open-ended reasoning over structured data. This survey consolidates current progress and outlines future directions to guide research in this rapidly evolving field. \textbf{Resources are available at \url{https://github.com/Zehong-Wang/Awesome-Foundation-Models-on-Graphs}.}

\end{abstract}

\newpage
\tableofcontents
\newpage

\section{Introduction}
\label{sec:introduction}

\begin{figure*}[!b]
      \centering
      \includegraphics[width=0.9\linewidth]{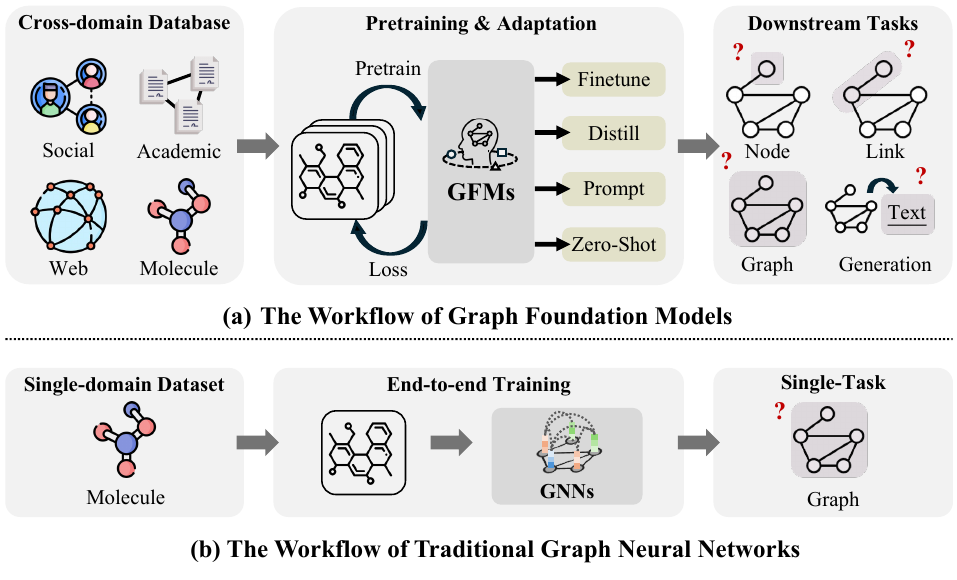}
      \caption{
            \textbf{From Task-Specific Graph Models to General-Purpose Graph Foundation Models.} This figure contrasts the paradigm shift from traditional Graph Neural Networks (GNNs) to Graph Foundation Models (GFMs). \textbf{(a)} GFMs are pretrained on large-scale graph corpora spanning multiple domains (e.g., social, web, academic, molecular) to acquire broadly transferable representations. Through various adaptation techniques—such as fine-tuning, distillation, prompting, or zero-shot inference—they can generalize across a wide spectrum of downstream tasks, including node classification, link prediction, graph classification, and graph-to-text generation. \textbf{(b)} In contrast, traditional GNNs are typically trained in an end-to-end manner on a single-domain dataset for a specific task, often lacking the scalability and generalization capabilities required for open-world settings. This shift mirrors the transition observed in language and vision domains, where foundation models have redefined the standard for general-purpose intelligence.
      }
      \label{fig:intro}
\end{figure*}

The pursuit of a \textit{one-model-fits-all} paradigm stands as one of the most ambitious and transformative goals in machine learning. This vision aspires to develop highly generalizable models capable of performing a wide spectrum of tasks across diverse domains, without requiring extensive task-specific architecture design or training. Historically, machine learning has been dominated by specialized models tailored to specific data modalities and objectives~\cite{mahesh2020machine}, often requiring handcrafted features~\cite{dong2018feature} and domain-dependent optimization strategies~\cite{bhavsar2012comparative}. From early rule-based systems and linear classifiers to the rise of deep learning, the evolution of machine learning has been marked by progressive gains in representation learning, scalability, and task performance~\cite{alpaydin2021machine, lecun2015deep}. Classical models such as decision trees, support vector machines (SVMs), and k-nearest neighbors (KNN) demonstrated success in low-dimensional and structured settings, but faced challenges when applied to high-dimensional, unstructured, or multimodal data. The emergence of deep learning models—such as convolutional neural networks (CNNs) for vision~\cite{he2016deep} and recurrent neural networks (RNNs) for sequential data~\cite{hochreiter1997long, chung2014empirical}—significantly advanced performance in perceptual tasks. Nonetheless, these models still required task-specific tuning, architecture adjustments, and large-scale labeled datasets to achieve robust generalization. A paradigm shift occurred with the development of \textit{transfer learning}~\cite{zhuang2020comprehensive} and \textit{self-supervised learning}~\cite{liu2021self}, which enabled models to learn broadly transferable representations from large-scale unlabeled data. These developments laid the groundwork for the emergence of \textit{foundation models}~\cite{bommasani2021opportunities}, which are trained on massive datasets with the objective of acquiring universal knowledge that can be readily adapted to a wide array of downstream tasks.

Foundation models are characterized by their scale, general-purpose nature, and pretraining across heterogeneous data sources. They are built to capture transferable inductive biases, enabling strong performance with minimal task-specific supervision. Scaling laws~\cite{kaplan2020scaling, snell2024scaling} and data-driven learning paradigms have driven their success across numerous domains, including natural language processing, computer vision, and robotics. For instance, \textit{Large Language Models} (LLMs)~\cite{achiam2023gpt, zhou2023comprehensive} process text by tokenizing input sequences and formulating tasks such as translation, summarization, or reasoning as autoregressive next-token prediction problems. Similarly, \textit{Large Vision Models} (LVMs)~\cite{yuan2021florence, yu2023language, bai2023sequential} treat visual inputs as sequences of tokens and apply Transformer-based architectures for visual question answering, captioning, or image generation. These models exhibit remarkable zero-shot and few-shot generalization capabilities, enabling rapid adaptation to novel tasks without requiring substantial fine-tuning.

In this context, the rise of \textit{Graph Foundation Models} (GFMs), as illustrated in Figure \ref{fig:intro}, seeks to extend these capabilities to graph-structured data—an essential yet fundamentally different modality characterized by relational dependencies, permutation invariance, and non-Euclidean geometry \cite{kipf2017semisupervised,velickovic2018graph,hamilton2017inductive}. GFMs aspire to offer a unified, pretrainable, and adaptable solution for a wide range of graph-based applications, spanning from molecular property prediction and knowledge graph reasoning to social network analysis and recommendation systems. For instance, OFA~\cite{liu2024one} operates on eight text-attributed graphs (TAGs), spanning citation networks, Wikipedia networks, knowledge graphs, and molecular graphs, where each node is associated with a textual description. By employing a shared textual encoder, OFA projects these node descriptions into a unified embedding space, thereby aligning node features across graphs. To bridge the gap between pretraining and downstream tasks, it further introduces a \textit{prompt graph} mechanism tailored to facilitate task adaptation. Similarly, GFT~\cite{wang2024gft} identifies transferable patterns in graph data by modeling them as computation trees. It aligns node representations across graphs via a tree reconstruction task designed to capture cross-domain generalization. A key innovation of GFT lies in its construction of a transferable tree vocabulary, which encodes structural patterns shared across diverse graph domains. Beyond these general-purpose models, various GFMs have been proposed for specific tasks—such as node classification~\cite{li2024zerog, zhao2024all}, anomaly detection~\cite{qiao_anomalygfm_2025}, and recommendation systems~\cite{gong2023unified}—or are specialized for particular domains, including knowledge graphs~\cite{galkin2023towards, galkin2024foundation}, molecular graphs~\cite{shoghi2023molecules, zhang2024unimot}, and computation graphs~\cite{chen2024graphwiz, guo2023gpt4graph}.

\noindent\textbf{Existing Surveys.}
Despite the rapid progress and growing interest in GFMs, the literature still lacks a comprehensive and unified survey that systematically covers the breadth and depth of this emerging field. Existing reviews tend to focus on isolated aspects of GFMs, offering fragmented insights without capturing the full landscape of foundational techniques, design challenges, and research directions. For instance, Liu et al.~\cite{liu2023towards} propose a taxonomy of GFMs based on backbone architectures—categorizing them into GNN-based, LLM-based, and hybrid GNN+LLM models—but their discussion remains limited to methodologies, without delving into applications and theoretical understandings. Zhao et al.~\cite{zhao2024survey} center their analysis around pretraining objectives, offering valuable insights into learning paradigms. However, their scope excludes broader system design and theoretical insights. Mao et al.~\cite{mao2024graph} provide a theoretical perspective on transferability within GFMs, shedding light on generalization capacity but omitting concrete methodological advances and empirical systematization. Wang et al.~\cite{wang2025towards} similarly emphasize transferability and emergent abilities, but without encompassing the full architectural, algorithmic, and application-driven spectrum of GFMs. In a complementary direction, Zhao et al.~\cite{zhao2025crossdomain} survey methods for cross-domain graph learning, a crucial but singular facet of GFM design. However, effective foundation models must also address cross-task generalization and structural alignment across diverse graph types. Other works such as Wu et al.~\cite{wu2025graph} explore the use of GFMs in specific domains like recommender systems, while recent reviews~\cite{jin2024large, li2023survey, fan2024graph, ren2024survey} focus on the integration of GNNs and LLMs, treating them as a subfield rather than part of a cohesive GFM framework.

\noindent\textbf{Our Position.}
In contrast, our survey aims to bridge these gaps by offering a holistic and systematic review of graph foundation models. We begin by outlining the historical development and foundational challenges behind GFMs, followed by a unified framework that decomposes GFMs into their core components—backbone architectures, pretraining strategies, and adaptation mechanisms. We introduce a comprehensive taxonomy that classifies GFMs into universal, domain-specific, and task-specific paradigms. Moreover, we analyze theoretical foundations (e.g., transferability, emergent ability), benchmark resources, and current limitations. Finally, we synthesize open research challenges and future directions to guide the continued advancement of the field. Our key contributions are summarized as follows:

\begin{itemize}[leftmargin=10pt]

      \item \textbf{Challenges in Designing GFMs (Section~\ref{sec:challenges}).}
            We identify and categorize the fundamental challenges in building graph foundation models into three core dimensions: \textit{feature heterogeneity}, \textit{structural heterogeneity}, and \textit{task heterogeneity}. These challenges highlight the unique complexities of learning from graph-structured data at scale.

      \item \textbf{A Unified Framework (Section~\ref{sec:techniques}).}
            We propose a unified modular framework that decomposes GFMs into three key components: \textit{backbone architectures}, \textit{pretraining strategies}, and \textit{adaptation mechanisms}. This abstraction facilitates a systematic understanding of diverse design choices and supports composability across methods.

      \item \textbf{Taxonomy and Comprehensive Review (Sections~\ref{sec:universal gfm},~\ref{sec:task gfm},~\ref{sec:domain gfm}).}
            We introduce a principled taxonomy that classifies GFMs into three categories based on their scope and generalization capacity: \textit{universal GFMs}, \textit{domain-specific GFMs}, and \textit{task-specific GFMs}. For each category, we conduct an extensive literature review\footnote{As of April 1, 2025}, detailing design philosophies and summarizing representative models.

      \item \textbf{Theoretical Foundations (Section~\ref{sec:theory}).}
            We explore the theoretical foundations underpinning GFMs, with a focus on scaling laws, transferability theory, and emerging understanding of generalization in graph-based pretraining. These insights provide formal grounding for the empirical success of GFMs.

      \item \textbf{Resources and GitHub Repository (Section~\ref{sec:resource}).}
            To support reproducibility and accelerate ongoing research, we compile and release a curated repository of resources, including benchmark datasets, open-source implementations, pretrained models, and a living GitHub collection: \url{https://github.com/Zehong-Wang/Awesome-Foundation-Models-on-Graphs}.

      \item \textbf{Open Questions (Section~\ref{sec:open question}).}
            We conclude by outlining key open problems in the development of GFMs, such as effective alignment across heterogeneous graphs, scalable and efficient adaptation mechanisms, robust evaluation protocols, and deeper theoretical insights. These challenges point to promising avenues for advancing the next generation of general-purpose graph learning systems.

\end{itemize}

\noindent\textbf{Summary of Future Directions in Building Graph Foundation Models.}
Despite recent progress, the development of GFMs remains in its infancy, with numerous open challenges spanning scalability, data availability, evaluation, utilization, and theoretical understanding. First, unlike LLMs and VLMs that benefit from established scaling laws, GFMs require more scalable architectures, high-level generative objectives, and unified learning instances to unlock similar performance gains. Second, addressing the data scarcity inherent to graphs calls for automated graph data collection, high-fidelity synthetic generation, and quality-centric dataset curation strategies. Third, evaluating GFMs demands benchmarks that reflect real-world tasks, alongside metrics that capture generalization, robustness, and trustworthiness. Fourth, effectively utilizing GFMs involves improving adaptation mechanisms (e.g., zero-shot and prompt-based learning), identifying high-impact applications beyond traditional graph tasks, and integrating multimodal knowledge representations. Finally, theoretical foundations remain underexplored—key issues include understanding the limits of transferability, resolving cross-domain pattern conflicts, ensuring robustness under distribution shifts, and deriving generalization guarantees. Addressing these open questions is essential to realizing the full potential of GFMs across diverse domains. The comprehensive discussion is provided in Section \ref{sec:open question}.


\newpage

\section{Background}
\label{sec:background}

\subsection{A Brief History of Graph Learning}

Similar to the trajectory observed in natural language processing (NLP) and computer vision (CV), graph machine learning is undergoing a paradigm shift—from highly specialized, task-specific models toward more unified and general-purpose frameworks. This evolution has progressed through several key milestones, as outlined below:

\begin{figure*}[!b]
    \centering
    \includegraphics[width=\linewidth]{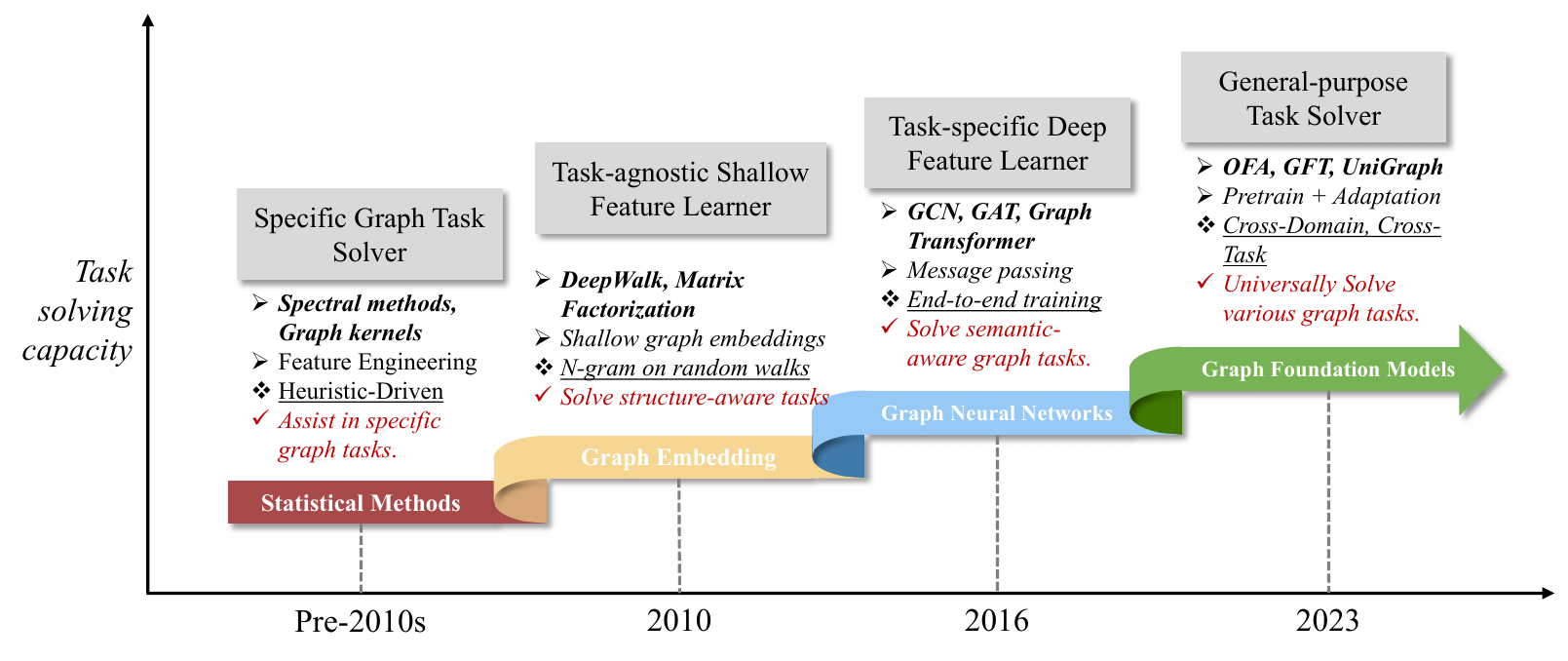}
    \caption{
        \textbf{The Evolution of Graph Learning Paradigms.}
        This figure illustrates the historical trajectory of graph learning, highlighting the increasing task-solving capacity over time.
        \textbf{(1)} \textit{Statistical methods} (pre-2010s) relied on heuristic-driven techniques, such as spectral analysis and graph kernels, to solve narrowly scoped graph tasks.
        \textbf{(2)} \textit{Graph embeddings} (circa 2010) introduced shallow, task-agnostic representations via random walks or matrix factorization, enabling better structural understanding.
        \textbf{(3)} \textit{Graph neural networks} (2016 onward) adopted deep learning principles—particularly message passing—to build end-to-end task-specific models capable of capturing semantic dependencies.
        \textbf{(4)} \textit{Graph foundation models} (post-2023) represent the latest paradigm, aiming for universal, general-purpose solvers that are pretrained on large-scale graphs and adapted to diverse downstream tasks across domains.
        This timeline reflects a broader shift from handcrafted, task-bound solutions to scalable, generalizable graph intelligence.
    }
    \label{fig:development}
\end{figure*}

\begin{itemize}[leftmargin=10pt]

    \item \textbf{Traditional Graph Learning Methods.}
          Early approaches to graph learning were deeply rooted in classical graph theory and combinatorial optimization~\cite{bondy2008graph}. Algorithms such as shortest path computation~\cite{dreyfus1969appraisal}, spectral clustering~\cite{von2007tutorial}, and graph kernels~\cite{vishwanathan2010graph} enabled important applications in network analysis, community detection, and graph matching. However, these methods often relied on handcrafted features, struggled with scalability, and lacked the capacity to learn rich, transferable representations.

    \item \textbf{Graph Embedding.}
          The integration of representation learning into graphs led to the emergence of graph embedding techniques. Methods such as DeepWalk~\cite{perozzi2014deepwalk}, node2vec~\cite{grover2016node2vec}, and LINE~\cite{tang2015line} introduced the idea of mapping nodes into low-dimensional continuous vector spaces via random walks or neighborhood sampling. These embeddings proved effective for downstream tasks like node classification, clustering, and link prediction. Nonetheless, they were largely transductive, lacked inductive generalization, and captured structural patterns without accommodating node or edge attributes.

    \item \textbf{Graph Neural Networks.}
          A transformative step occurred with the introduction of Graph Neural Networks (GNNs)~\cite{wu2020comprehensive}, which brought deep learning principles to non-Euclidean graph structures. GNNs utilize message-passing mechanisms~\cite{gilmer2017neural} to iteratively aggregate and update node representations based on their neighbors. Key models include Graph Convolutional Networks (GCNs)~\cite{kipf2017semisupervised}, Graph Attention Networks (GATs)~\cite{velickovic2018graph}, and GraphSAGE~\cite{hamilton2017inductive}, each improving expressive power and generalization in different contexts. While GNNs advanced the field significantly, they remained constrained by challenges such as oversmoothing, limited receptive fields, and the need for extensive task-specific architecture tuning.

    \item \textbf{Graph Foundation Models.}
          Inspired by the success of foundation models in NLP and vision, graph learning has recently entered the era of \textit{Graph Foundation Models}. These models~\cite{wang2024gft,shoghi2023molecules,kong2025gofa} are pretrained on large-scale graphs using self-supervised objectives~\cite{liu2022graph}, enabling them to learn universal representations that transfer across tasks and domains. GFMs integrate both structural dependencies and semantic content, exhibiting strong zero-shot and few-shot generalization. Applications span diverse domains, including molecular property prediction, social network analysis, recommendation systems, and knowledge graphs. By decoupling model training from specific tasks, GFMs reduce reliance on labeled data and domain-specific heuristics, moving the field closer to general-purpose graph intelligence.

\end{itemize}

\subsection{Background of Foundation Models}

\textbf{Foundation Models.} Foundation models have emerged as a cornerstone in modern artificial intelligence, representing a paradigm shift from narrowly designed task-specific models to highly generalizable systems. These models are defined by large-scale pretraining on diverse and heterogeneous datasets—ranging from web text and books to images, code, and multimodal content—which enables them to develop broad, transferable capabilities across a wide array of domains. The term "foundation model" was popularized by the Stanford Institute for Human-Centered Artificial Intelligence \cite{bommasani2021opportunities}, underscoring the shared trends across modalities such as language, vision, code, and audio. Architecturally, foundation models are typically built upon the transformer framework \cite{vaswani2017attention}, which leverages self-attention mechanisms to effectively capture long-range dependencies and scale across billions of parameters. Prominent examples include GPT \cite{brown2020language}, BERT \cite{devlin2019bert}, PaLM \cite{anil2023palm}, and LLaMA \cite{touvron2023llama} in the language domain, as well as CLIP \cite{radford2021learning} and DALL·E \cite{radford2021learning,ramesh2021zero} for vision-language tasks.

A defining strength of foundation models lies in their ability to learn general-purpose representations that are highly adaptable to new tasks with minimal additional supervision. Through exposure to large and diverse pretraining data, these models acquire versatile semantic, syntactic, and structural knowledge, which can be transferred to a wide range of downstream tasks via fine-tuning, prompting, or instruction tuning. Their transformer-based architecture further enhances their capacity to model complex and structured data, making them suitable for tasks involving reasoning, generation, and classification. Notably, foundation models demonstrate strong few-shot and zero-shot learning capabilities by leveraging in-context learning—where the model performs new tasks based solely on input prompts without additional gradient updates. This opens up a flexible interface for users to guide model behavior through natural language or structured prompts.

As these models scale in size and training data, they begin to exhibit emergent behaviors—capabilities that were not explicitly programmed or observed in smaller models~\cite{wei2022emergent,kaplan2020scaling}. These include logical reasoning, chain-of-thought generation, tool use, and other complex cognitive functions that arise organically during training. While these behaviors highlight the vast potential of foundation models as general-purpose AI systems, they also introduce new challenges. Issues such as data and societal bias, interpretability, safety, and environmental cost become increasingly relevant as these models are integrated into real-world applications. Nonetheless, the ability of foundation models to unify learning across tasks and domains continues to reshape the AI landscape and drive innovation across disciplines.

\textbf{Graph Foundation Models.} Inspired by the success of foundation models on other domains, graph foundation models are proposed as specialized foundation models designed to understand and reason over graph-structured data. Graphs—comprising nodes (entities) and edges (relationships)—are widely used to represent complex systems such as social networks, molecular structures, and knowledge graphs. Same as other foundation models, GFMs are pre-trained on large, diverse graph datasets to learn general-purpose representations that can be fine-tuned or adapted for various downstream tasks like node classification, link prediction, and graph generation. By capturing the structural and relational properties inherent in graphs, GFMs enable more effective and scalable analysis across domains where interconnected data plays a central role. We summarize the key properties of GFMs in the following.

\begin{itemize}[leftmargin=10pt]
    \item \textbf{Pretraining on Large-Scale Graph Data} GFMs are trained on extensive and diverse graph datasets, enabling them to learn generalizable patterns and structural semantics across various domains (e.g., biology, social networks, and knowledge graphs).

    \item \textbf{General-Purpose Representations} GFMs learn universal node, edge, and graph-level embeddings that can be adapted to a wide range of tasks with minimal fine-tuning or prompting efforts.

    \item \textbf{Structural Awareness} GFMs inherently capture topological features of graphs—such as connectivity, neighborhood structure, and global graph properties—making them effective in modeling complex relationships.

    \item \textbf{Transferability} Similar to other foundation models, GFMs can transfer knowledge across tasks and domains, allowing for effective performance even with limited task-specific data.

    \item \textbf{Few-shot and Zero-shot Capabilities} Once pre-trained, GFMs can often perform new tasks with few or even no labeled examples by leveraging their rich internal representations.
\end{itemize}

\begin{table}[!b]
    \centering
    \caption{Summary of Notations}
    \label{tab:notation}
    \resizebox{0.9\linewidth}{!}{
        \begin{tabular}{p{4cm}p{10cm}}
            \toprule
            \textbf{Symbol}                              & \textbf{Description}                                        \\ \midrule
            $\mathcal{G}$                                & A graph                                                     \\ \hline
            $\mathcal{V}, \mathcal{E}$                   & Sets of nodes and edges in graph $\mathcal{G}$              \\ \hline
            $N, M$                                       & Number of nodes and edges                                   \\ \hline
            $v_i \in \mathcal{V}$                        & A node in graph $\mathcal{G}$                               \\ \hline
            $e_{ij} \in \mathcal{E}$                     & An edge in graph $\mathcal{G}$                              \\ \hline
            $\mathbf{X} \in \mathbb{R}^{N \times D}$     & Node attribute matrix for graph $\mathcal{G}$               \\ \hline
            $\mathbf{x}_i \in \mathbb{R}^{D}$            & Feature vector for node $v_i \in \mathcal{V}$               \\ \hline
            $\mathbf{E} \in \mathbb{R}^{M \times D}$     & Edge attribute matrix for graph $\mathcal{G}$               \\ \hline
            $\mathbf{e}_{ij} \in \mathbb{R}^{D}$         & Feature vector for edge $e_{ij} \in \mathcal{E}$            \\ \hline
            $\mathbf{A} \in \{0, 1\}^{N \times N}$       & Adjacency matrix of graph $\mathcal{G}$                     \\ \hline
            $\mathbf{D}$                                 & Textual information on graphs                               \\ \hline
            $\mathbf{d}_{v_i}$                           & Text description associated with node $v_i$                 \\ \hline
            $\mathbf{d}_{e_{ij}}$                        & Text description associated with edge $e_{ij}$              \\ \hline
            $\mathbf{d}_{\mathcal{G}}$                   & Textual description associated with the entire graph        \\ \hline
            $\mathbf{Z} \in \mathbb{R}^{N \times D^{'}}$ & Learned node representations                                \\ \hline
            $\mathbf{z}_i \in \mathbb{R}^{D^{'}}$        & Learned representation of node $v_i \in \mathcal{V}$        \\ \hline
            $\mathcal{N}_{v}$                            & Neighborhood set of node $v \in \mathcal{V}$                \\ \hline
            $\mathcal{T}$                                & Set of augmentation functions                               \\ \hline
            $\mathbf{W}, \mathbf{\Theta}, w, \theta$     & Learnable parameters of the model                           \\ \hline
            $t \sim \mathcal{T}$                         & A specific augmentation function sampled from $\mathcal{T}$ \\ \hline
            $| \cdot |$                                  & Cardinality of a set                                        \\ \hline
            $\|$                                         & Concatenation operator                                      \\ \hline
            $\operatorname{GNN}(\cdot)$                  & Graph Neural Network (GNN) encoder                          \\ \hline
            $\operatorname{LLM}(\cdot)$                  & Large Language Model (LLM) encoder                          \\
            \bottomrule
        \end{tabular}
    }
\end{table}

\subsection{Definitions \& Notations}

We introduce key notations and concepts used throughout this paper. Table \ref{tab:notation} summarizes the primary symbols and their respective meanings. Throughout the paper, bold uppercase letters denote matrices, while bold lowercase letters represent vectors.

\begin{definition}[Graph]
    A graph is represented as $\mathcal{G} = (\mathcal{V}, \mathcal{E})$, where $\mathcal{V}$ and $\mathcal{E} \subseteq \mathcal{V} \times \mathcal{V}$ are node and edge sets, where $N = |\mathcal{V}|$ and $M = |\mathcal{E}|$ denote the number of nodes and edges. This defines the adjacency matrix $\mathbf{A}$. Iff $(i, j) \in \mathcal{E}$, then $\mathbf{A}_{ij} = 1$; otherwise, $\mathbf{A}_{ij} = 0$. This allows us to define the neighborhoods for each node in the graph as $\mathcal{N}_v = \{u \in \mathcal{V} | (u, v) \in \mathcal{E} \}$.
\end{definition}

\begin{definition}[Attributed Graph]
    If a graph associates to node features or edge features, the graph is denoted as attributed graph. This is represented as $\mathcal{G} = (\mathbf{X}, \mathbf{A})$. The $\mathbf{X} \in \mathbb{R}^{N \times D}$ and $\mathbf{A} \in \{0,1\}^{N \times N}$ denote the node attributes and adjacent matrix, respectively. The raw attribute of node $v_i \in \mathcal{V}$ is represented by $\mathbf{x}_i \in \mathbb{R}^D$.
\end{definition}

\begin{definition}[Text-Attributed Graph (TAG)]
    A text-attributed graph is formally defined as $\mathcal{G} = (\mathbf{X}, \mathbf{A}, \mathbf{D})$, where $\mathbf{X}$ denotes node attributes, $\mathbf{A}$ represents the adjacency matrix, and $\mathbf{D}$ encapsulates textual descriptions associated with nodes, edges, or the entire graph. Specifically, the textual description linked to a node $v_i$ is denoted as $\mathbf{d}_{v_i}$, while $\mathbf{d}_{e_{ij}}$ corresponds to the textual description of edge $e_{ij}$. Additionally, $\mathbf{d}_{\mathcal{G}}$ represents the textual information describing the entire graph $\mathcal{G}$.
\end{definition}

\begin{definition}[Graph Neural Network (GNN)]
    Graph neural networks are a class of neural architectures specifically designed to operate on graph-structured data. Given an attributed graph $\mathcal{G} = (\mathbf{X}, \mathbf{A})$, GNNs learn node, edge, or graph-level representations by recursively aggregating and transforming information from local neighborhoods. GNNs capture the relational and topological structure of graphs and are widely used in tasks such as node classification, link prediction, and graph classification.
\end{definition}

\begin{definition}[Large Language Model (LLM)]
    Large language models are deep neural architectures, typically based on the Transformer framework, that are pretrained on massive textual corpora to learn general-purpose language representations. Given an input text sequence or query $q$, $\operatorname{LLM}(\cdot)$ produces a contextual output, often modeled via next-token prediction. LLMs exhibit strong capabilities in text understanding, generation, and reasoning, and support a wide range of downstream applications without requiring explicit fine-tuning. When applied to graph-structured data, LLMs can process text-attributed graphs by leveraging associated textual information for structure-aware inference.
\end{definition}

\begin{definition}[Graph Foundation Model (GFM)]
    Graph foundation models are a class of large-scale models pretrained on extensive cross-domain and cross-task graph datasets. Through pretraining, GFMs acquire transferable knowledge and general-purpose capabilities, demonstrating emergent properties and adaptability across diverse graph-based applications. These include molecular property prediction, recommender systems, social network analysis, and anomaly detection, where GFMs effectively leverage structural and relational information to enhance predictive performance.
\end{definition}

\newpage

\section{Challenges in Designing Graph Foundation Models}
\label{sec:challenges}

\begin{figure*}[!b]
    \centering
    \includegraphics[width=0.8\linewidth]{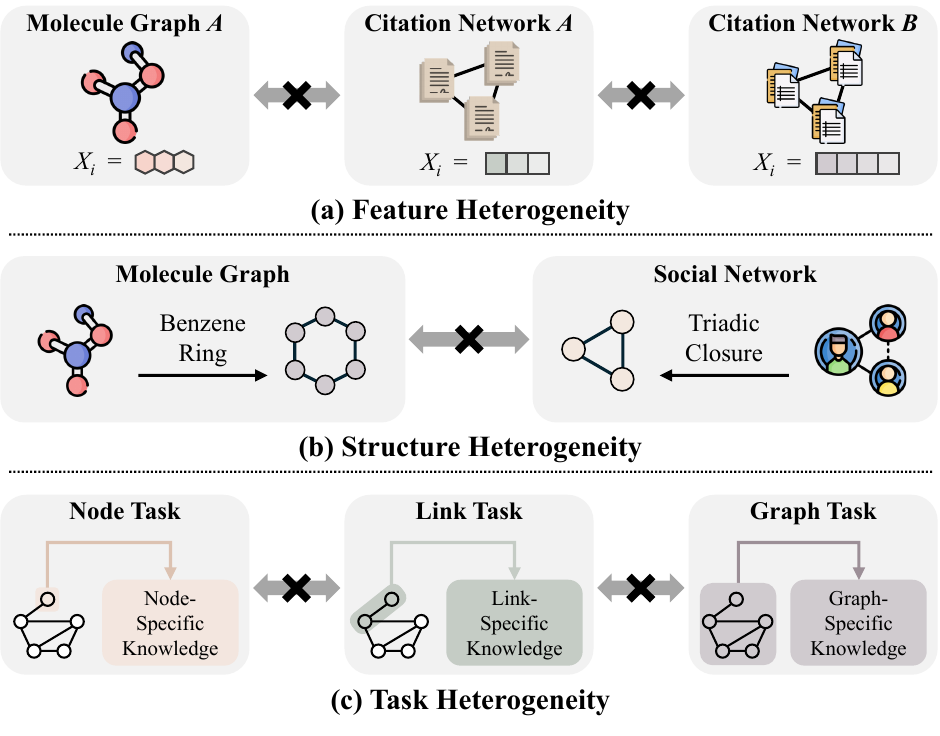}
    \caption{
        \textbf{Key Challenges in Designing Graph Foundation Models.}
        This figure illustrates three fundamental sources of heterogeneity that challenge the development of general-purpose GFMs:
        \textbf{(a) Feature Heterogeneity:} Graph datasets vary widely in their node features due to differences in domain semantics and data preprocessing. For instance, molecular graphs use atom-level descriptors, while citation networks rely on textual or structural attributes. Such diversity complicates unified representation learning.
        \textbf{(b) Structure Heterogeneity:} Graphs from different domains exhibit distinct topological patterns—molecules contain recurring chemical motifs such as benzene rings, whereas social networks emphasize relational structures like triadic closures. Capturing such heterogeneous structures requires models to be structurally adaptive.
        \textbf{(c) Task Heterogeneity:} Graph learning tasks span multiple granularities—including node-level classification, link prediction, and graph-level classification—each requiring different inductive biases. A universal GFM must effectively generalize across these varied task formulations.
        Together, these challenges underscore the need for robust, flexible architectures and training paradigms capable of aligning graphs across both structure and semantics.
    }

    \label{fig:challenges}
\end{figure*}

Graph datasets inherently capture complex real-world phenomena through rich and diverse relational structures. Due to the heterogeneous nature of graph-structured data across domains, designing a single, unified model that generalizes well to various graphs remains a substantial challenge. GFMs are designed to overcome these barriers by learning transferable and adaptable representations across diverse graph settings. In this section, we highlight three fundamental challenges in designing GFMs: \textit{feature heterogeneity}, \textit{structure heterogeneity}, and \textit{task heterogeneity}. These challenges collectively encapsulate the difficulty of building a universally applicable model across diverse graph datasets and learning scenarios.

Addressing feature, structure, and task heterogeneity is central to the development of graph foundation models. We provide a conceptual analysis of each challenge and its root causes below. Strategies to resolve these issues are discussed in depth in Sections~\ref{sec:universal gfm}, \ref{sec:task gfm}, and \ref{sec:domain gfm}.

\subsection{Feature Heterogeneity}
\label{feature heterogenetiy}

Feature heterogeneity refers to differences in node, edge, or graph-level features across datasets. This challenge arises from two primary sources: (1) domain-specific differences and (2) preprocessing inconsistencies, as shown in Figure~\ref{fig:challenges}(a). \textit{Domain-specific differences} occur because graphs from different fields encode different semantics. For instance, citation networks represent academic relationships and feature attributes such as paper titles, authors, and abstracts. In contrast, molecular graphs represent chemical compounds, where features describe atom types, bond orders, and spatial configurations. \textit{Preprocessing inconsistencies} stem from the use of different feature extraction pipelines, even within the same domain. For example, while both \texttt{Cora} and \texttt{Pubmed} are citation networks, \texttt{Cora} encodes nodes with a 1433-dimensional one-hot representation of keywords, whereas \texttt{Pubmed} uses 500-dimensional TF-IDF vectors, leading to different input feature spaces.

\textbf{Remedy.} Traditional GNNs assume a consistent input feature space, making them not suitable for handling feature heterogeneity. Several solutions have been proposed to mitigate this issue, such as using singular value decomposition (SVD) to project features into a shared latent space~\cite{xia2024opengraph,xia2024anygraph}, leveraging large-scale language models for textual encoding~\cite{wang2024gft,liu2024one}, or applying graph-specific feature projection layers~\cite{zhao2024all,yu2025samgpt}. Furthermore, GraphAny \cite{zhao2025fully} poses a promising direction by getting inductive features based on relative (instead of absolute) information on graphs. While promising, these methods often depend on domain knowledge, handcrafted mappings, or introduce scalability bottlenecks. Developing a more general and adaptive solution for heterogeneous feature alignment remains an open research direction.

\subsection{Structure Heterogeneity}
\label{structure heterogeneity}

Structure heterogeneity refers to the differences in topological patterns across graph datasets. These differences significantly impact model performance, as graph structure plays a crucial role in downstream reasoning tasks. For example, social and citation networks often exhibit localized dependencies and motifs such as star-shaped hubs and triangles that capture popularity and community structures. In contrast, molecular graphs exhibit long-range dependencies, characterized by ring structures and $k$-cliques that encode chemical substructures and interactions. These structural patterns vary widely across domains, complicating the development of a GNN model that generalizes effectively across graph types. Increasing model depth or capacity is not a sufficient solution, as it introduces problems such as over-smoothing~\cite{keriven2022not} and over-squashing~\cite{topping2021understanding}, both of which hinder the propagation of discriminative signals. Moreover, the locality bias inherent in traditional message-passing GNNs limits their ability to capture global structures~\cite{ma2021homophily}.

\textbf{Remedy.} To improve structural adaptability, several techniques have been proposed, including structure-aware data augmentation~\cite{tang2024cross}, graph prompt tuning~\cite{yan2024inductive, fu2025edge}, and discrete structural codebooks~\cite{wang2024gft,xiamole}. However, current approaches still struggle to adapt to the full spectrum of graph structures found in real-world data.

\subsection{Task Heterogeneity}

Task heterogeneity captures the diversity of learning objectives in graph-based tasks, each requiring different modeling strategies and inductive biases. Unlike in natural language processing, where many tasks can be reformulated as question-answering, graph tasks vary substantially in formulation and underlying assumptions. \textit{Node-level tasks} aim to classify individual nodes using information from their local neighborhoods. Success in these tasks often depends on modeling homophily or heterophily in node interactions. \textit{Link-level tasks} focus on predicting the presence or type of edges between node pairs. These tasks often rely on proximity metrics, such as common neighbors or shortest paths, to infer relational patterns. \textit{Graph-level tasks} require holistic understanding of the entire graph, necessitating the extraction of subgraph motifs and global dependencies. Additionally, more complex domain-specific tasks, such as reasoning over knowledge graphs or molecule generation, introduce unique modeling requirements, further complicating generalization across task types.

\textbf{Remedy.} Tackling task heterogeneity involves either aligning tasks explicitly or developing task-agnostic methods. \textit{Explicit alignment} strategies reformulate diverse tasks into a unified objective, such as link prediction~\cite{sun2022gppt}, subgraph classification~\cite{liu2024one,he2024unigraph}, or tree classification~\cite{wang2024gft,wang2024learning}. \textit{Implicit alignment} methods aim to learn generalizable representations across tasks without relying on task-specific reformulations~\cite{bevilacqua2025holographic}. Although progress has been made, a universal GFM capable of seamlessly adapting to varied graph tasks remains a challenging and largely unsolved goal.

\newpage

\section{A Unified Framework of Graph Foundation Models}
\label{sec:techniques}

\subsection{Unified Framework}


Traditional graph learning models typically operate under a task-specific, end-to-end paradigm. A single graph dataset is provided as input, and a graph neural network is trained directly to solve a designated downstream task, such as node classification, link prediction, or graph classification. In contrast, graph foundation models embrace a paradigm shift: rather than being narrowly optimized for a specific task or dataset, they are pretrained on a diverse collection of graph datasets and subsequently adapted to a broad spectrum of downstream scenarios. Figure~\ref{fig:intro} illustrates the fundamental distinctions between traditional GNNs and GFMs. Central to this new paradigm is the principle of ``\textit{pretrain-then-adapt}''. Given a large-scale graph database $\gD_{pt}$ as the source of pretraining data, and a parameterized model backbone $\theta$ (which may be a GNN, Transformer, or even a language model), the goal is to learn generalized graph representations by minimizing a pretraining objective:
\begin{equation}
    \theta^* = \argmin_{\theta} \gL_{pt} (\gD_{pt}; \theta).
\end{equation}

Once pretraining is complete, the model parameters $\theta^*$ encapsulate transferable knowledge about graph structure, semantics, and dynamics. This pretrained model can then be directly applied to downstream tasks, or further adapted for improved performance. To enhance task-specific generalization, an additional adaptation stage can be employed. Given downstream data $\gD_{adapt}$ relevant to a target task or domain, the pretrained model parameters $\theta^*$ are fine-tuned by minimizing an adaptation loss:
\begin{equation}
    \hat{\theta} = \argmin_{\theta^*} \gL_{adapt} (\gD_{adapt}; \theta^*).
\end{equation}

This unified framework—comprising a backbone architecture, a pretraining strategy, and an adaptation mechanism—forms the core of GFM methodology. We describe each of these components in detail below:
\begin{itemize}[leftmargin=10pt]
    \item \textbf{Backbone:} The backbone refers to the foundational architecture responsible for processing graph-structured data and learning meaningful representations. It defines how nodes, edges, and global contexts are encoded, integrating structural dependencies and attribute information. Effective backbones can range from GNNs and LLMs to even hybrid architectures. A well-designed backbone is crucial for enabling scalability, multimodal integration, and cross-domain generalization in GFMs.

    \item \textbf{Pretraining:} Pretraining is the stage where the model learns general-purpose graph representations from large-scale, often unlabeled, graph corpora. This is typically achieved via self-supervised objectives such as contrastive learning, graph masking, or predictive modeling of structural and semantic properties. The goal is to endow the model with a rich understanding of universal graph patterns, fostering transferability, robustness, and data efficiency. A powerful pretraining scheme lays the groundwork for zero-shot and few-shot generalization across a multitude of graph tasks and domains.

    \item \textbf{Adaptation:} Adaptation refers to the process of aligning the pretrained model with specific downstream tasks or domains. This can involve fine-tuning all or part of the model parameters, employing lightweight tuning methods, or appending task-specific prediction heads. The adaptation phase ensures that the generalized knowledge from pretraining is effectively leveraged for specialized applications. When designed properly, adaptation enhances task performance, reduces data requirements, and facilitates rapid deployment across different graph scenarios.
\end{itemize}

In subsequent sections, we delve deeper into each component of this framework, surveying representative methods and highlighting emerging trends that shape the design and capabilities of graph foundation models.

\subsection{Backbone Architectures} 

The backbone architectures of GFMs are designed to integrate both structural and semantic information by leveraging the representational capabilities of graph-based models and large language models. Graph models—such as Graph Neural Networks (GNNs) and Graph Transformers—are particularly adept at capturing relational dependencies through message-passing mechanisms\footnote{Graph Transformers can be seen as performing global message passing between any pairs of nodes.} \cite{gilmer2017neural}. However, they often face challenges in modeling long-range dependencies and integrating multimodal data sources. Conversely, LLMs encode graph-structured data as sequences, enabling enhanced generalization and scalability. These models benefit from large-scale pretraining on textual corpora, but typically lack explicit graph inductive biases, which limits their capacity to faithfully represent topological structures \cite{chenllaga}. To mitigate these limitations, hybrid approaches have emerged that combine the strengths of GNNs and LMs. Such methods aim to fuse structural inductive biases with semantic generalization, thereby offering a more holistic understanding of graph-structured data.

\begin{figure*}[!t]
    \centering
    \includegraphics[width=\linewidth]{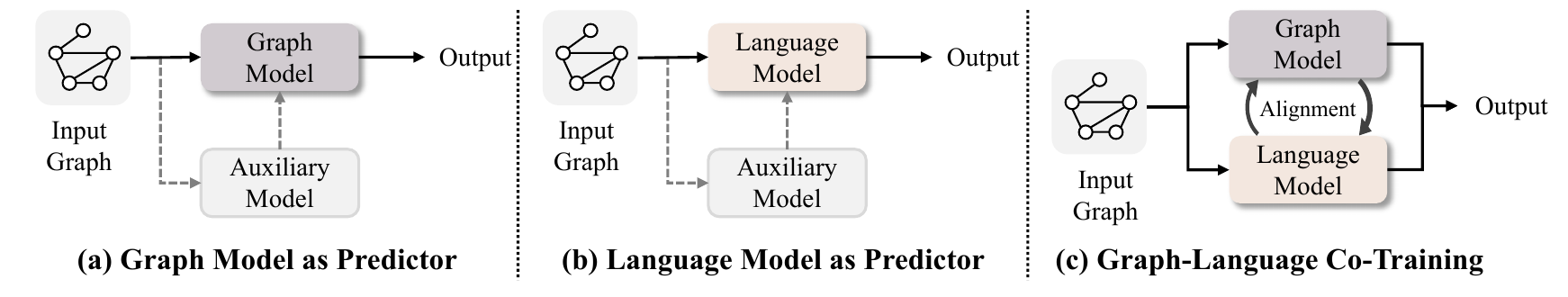}
    \caption{
        \textbf{Backbone Architectures of Graph Foundation Models.}
        We illustrate three representative paradigms:
        \textbf{(a)} Graph Model as Predictor, where GNNs serve as the primary reasoning engine with optional auxiliary modules;
        \textbf{(b)} Language Model as Predictor, where LLMs interpret graph-structured inputs converted into text or structured prompts;
        \textbf{(c)} Graph-Language Co-Training, which jointly optimizes GNNs and LLMs through alignment for enhanced generalization across tasks and modalities.
    }

    \label{fig:backbone}
\end{figure*}

In this section, we provide a systematic overview of the core backbones used in GFMs. We categorize them into three principal paradigms: (i) graph models as predictors, (ii) language models as predictors, and (iii) graph-language co-training frameworks. For each category, we discuss the architectural foundations, underlying assumptions, and representative applications. An illustrative taxonomy of these backbone strategies is presented in Figure~\ref{fig:backbone}. For detailed information, we recommend to read the awesome survey \cite{jin2024large}.

\subsubsection{Graph Model as Predictor}

Graph models play a central role in learning from graph-structured data, where nodes represent entities and edges define relational dependencies. Unlike traditional neural networks operating on Euclidean domains such as images or sequences, graph models extend deep learning paradigms to non-Euclidean spaces, enabling powerful representations across a diverse set of domains, including recommender systems \cite{zhang2024mopi}, social networks \cite{kipf2017semisupervised,wu2022graphbert, wang2024tackling}, molecular property prediction \cite{gilmer2017neural}, knowledge graphs \cite{chen2020review,zhang2024ngqa}, and healthcare analytics \cite{zhang2024diet}.

At the core of graph-based learning lies the message-passing paradigm, where node representations are iteratively updated by aggregating information from their local neighborhoods. Formally, given a graph $\mathcal{G} = (\mathcal{V}, \mathcal{E})$ with node set $\mathcal{V}$ and edge set $\mathcal{E}$, the update rule at the $k$-th layer of a GNN is typically defined as:
\begin{equation}
    \mathbf{h}_v^{(k)} = \text{UPDATE} \left( \mathbf{h}_v^{(k-1)}, \text{AGGREGATE} \left( \{ \mathbf{h}_u^{(k-1)} : u \in \mathcal{N}(v) \} \right) \right),
\end{equation}
where $\mathbf{h}_v^{(k)}$ denotes the feature embedding of node $v$ at layer $k$, $\mathcal{N}(v)$ is the set of its neighbors, and the \texttt{AGGREGATE} and \texttt{UPDATE} functions define the information propagation and transformation.

Different graph models instantiate these functions in unique ways. For instance, in Graph Convolutional Networks (GCNs) \cite{kipf2017semisupervised}, aggregation is performed via normalized summation: $\mathbf{h}_v^{(k)} = \sigma \left( \mathbf{W}^{(k)} \cdot \frac{1}{|\mathcal{N}(v)|} \sum_{u \in \mathcal{N}(v) \cup \{v\}} \mathbf{h}_u^{(k-1)} \right)$, where $\mathbf{W}^{(k)}$ is a learnable weight matrix and $\sigma$ is a non-linear activation. Graph Attention Networks (GATs) \cite{velickovic2018graph}, on the other hand, use attention mechanisms to assign importance scores $\alpha_{vu}$ to neighboring nodes: $\mathbf{h}_v^{(k)} = \sigma \left( \sum_{u \in \mathcal{N}(v)} \alpha_{vu} \cdot \mathbf{W}^{(k)} \mathbf{h}_u^{(k-1)} \right)$, where $\alpha_{vu}$ is computed via a self-attention mechanism, allowing nodes to dynamically focus on the most relevant neighbors.

Recent advances have pushed the expressive capacity of GNNs by incorporating global attention mechanisms inspired by Transformers \cite{vaswani2017attention}. Graph Transformers \cite{dwivedi2020generalization,wang2025gpm,rampasek2022recipe} replace local aggregation with attention over all nodes in the graph, enabling the capture of long-range dependencies:
\begin{equation}
    \mathbf{h}_v^{(k)} = \sum_{u \in \mathcal{V}} \alpha_{vu} \cdot \mathbf{W} \mathbf{h}_u^{(k-1)}, \quad
    \alpha_{vu} = \frac{\exp\left(\phi(\mathbf{h}_v, \mathbf{h}_u)\right)}{\sum_{w \in \mathcal{V}} \exp\left(\phi(\mathbf{h}_v, \mathbf{h}_w)\right)},
\end{equation}
where $\phi(\cdot, \cdot)$ denotes a learnable compatibility function (e.g., dot-product), and $\alpha_{vu}$ reflects the attention weight assigned to node $u$ from the perspective of node $v$.

\noindent\textbf{Graph Models without Auxiliary Modules.} Several GFMs adopt pure graph-based backbones, relying solely on structural signals without external models. These approaches capitalize on the inductive bias of GNNs for relational learning. For example, \textsc{MiniMol}~\cite{klaser2024minimol} introduces a parameter-efficient GNN for molecular property prediction, where atomic features $\mathbf{x}_i$ are encoded via a shared initialization function $\mathbf{h}_i^{(0)} = f_{\theta}(\mathbf{x}_i)$. \textsc{JMP}~\cite{shoghi2023molecules} proposes a hierarchical representation scheme, with node embeddings updated using a degree-normalized formulation $\mathbf{h}_i^{(t)} = \sigma \left( \mathbf{W} \sum_{j \in \mathcal{N}_i} \frac{\mathbf{h}_j^{(t-1)}}{\sqrt{d_i d_j}} \right)$, where $d_i$ and $d_j$ are the degrees of nodes $i$ and $j$, respectively. These approaches demonstrate strong performance in domains with rich relational signals, though they may struggle in scenarios requiring multimodal reasoning or external contextual knowledge.

\noindent\textbf{Graph Models with Auxiliary Language Models.} To address the limitations of pure GNNs, several methods incorporate language models as auxiliary modules. These hybrid approaches enable the integration of unstructured textual knowledge into graph representation learning. Two primary strategies have emerged: (i) enhancing node features using LLMs prior to graph encoding, and (ii) employing LLMs as direct encoders over pure graph structure. In the first category, \textsc{TAPE}~\cite{he2023harnessing} introduces an LLM-to-LM interpreter that extracts task-specific textual explanations, which are then converted into structured node features. Specifically, given a textual explanation $\mathbf{e}_v$ generated by a language model, node embeddings are computed via $\mathbf{h}_v = f_{\text{LM}}(\mathbf{e}_v)$, where $f_{\text{LM}}$ is a frozen or fine-tuned language model. These embeddings are then used within the graph model to enhance learning. In the second strategy, LLMs act as standalone encoders of node-level text. For example, \textsc{OFA}~\cite{liu2024one} proposes a unified textual template to align different graph node descriptions, which are then encoded into a shared embedding space using a pretrained LM. This enables zero-shot generalization across domains by leveraging the linguistic alignment of node semantics. These auxiliary-enhanced graph models highlight the flexibility and effectiveness of hybrid architectures, capable of integrating structural and semantic signals in a unified learning framework.

\subsubsection{Language Model as Predictor}

Language models, initially developed for natural language processing tasks such as machine translation, summarization, and question answering \cite{zhao2023survey}, have recently found increasing utility in graph learning applications \cite{ren2024survey,li2023survey}. By leveraging large-scale pre-trained models \cite{devlin2019bert,brown2020language,touvron2023llama,touvron2023llama2,grattafiori2024llama,bai2023qwen,guo2025deepseek,zhu2024understanding}, these approaches enable structured data to be encoded, reasoned over, and generalized through powerful language-driven representations. A key example is BERT \cite{devlin2019bert}, a bidirectional Transformer that optimizes a masked-token prediction (MTP) objective:
\begin{equation}
    \mathbb{E}_{S \sim \mathcal{D}} \left[ \sum_{s_i \in S} \log p(s_i | s_1, \dots, s_{i-1}, s_{i+1}, \dots, s_{N_S}) \right],
\end{equation}
where $S$ is a sentence sampled from corpus $\mathcal{D}$, $s_i$ denotes a masked token, and $N_S$ is the sentence length. This objective promotes contextual representation learning across both left and right contexts. In contrast, autoregressive language models such as GPT-3 \cite{brown2020language} use next-token prediction (NTP), predicting each token sequentially:
\begin{equation}
    \mathbb{E}_{S \sim \mathcal{D}} \left[ \sum_{s_i \in S} \log p(s_i | s_1, \dots, s_{i-1}) \right].
\end{equation}
Scaling model size and training data has led to emergent capabilities, including in-context learning \cite{dong2022survey}, chain-of-thought reasoning \cite{wei2022chain}, and zero-shot generalization \cite{kojima2022large}. Foundation models such as GPT \cite{achiam2023gpt}, PaLM \cite{anil2023palm}, and LLaMA \cite{touvron2023llama,touvron2023llama2,grattafiori2024llama} exhibit strong compositional reasoning and flexible adaptation, making them promising candidates for graph-based learning when appropriately adapted.

\noindent\textbf{Language Models without Auxiliary Modules.} Standalone language models can be used as graph predictors by transforming graph structures into sequential representations, allowing graph reasoning via text processing without relying on GNNs. This approach serializes the graph structure into a textual input, enabling LLMs to interpret node identities, features, and relationships. The transformation pipeline typically follows:
\begin{equation}
    \vo = \operatorname{LLM}(\mathcal{S}(\mathcal{G})), \quad \mathcal{S}(\mathcal{G}) = \sum_{v_i \in \mathcal{V}} \text{Tokenize}(v_i, \mathbf{x}_i, \mathcal{N}(v_i)),
\end{equation}
where $\text{Tokenize}(\cdot)$ encodes each node $v_i$, its features $\mathbf{x}_i$, and its neighborhood $\mathcal{N}(v_i)$ into a structured sequence. Language models then perform inference based solely on this textual input. LangGFM~\cite{lin2024langgfm} exemplifies this method by converting graphs into natural language templates and demonstrating that LLMs can reason over structural patterns without explicit graph operations. BeyondText~\cite{hu2023beyond} further validates LLMs' ability to recover graph topologies and relational dependencies through textual prompts. To optimize these strategies, GLM~\cite{ye2024language} adopts reinforcement learning to refine prompt engineering, tailoring input formats for graph-aware inference. Despite their flexibility, pure LLM-based methods can struggle to internalize structural inductive biases, particularly for graphs with rich topologies or weak textual signals. To address these challenges, recent work introduces structured tokenization, retrieval mechanisms, and task-specific tuning—leading toward hybrid VLM-style frameworks.

\noindent\textbf{Language Models with Auxiliary Modules.}
VLM-style architectures enhance language models by incorporating explicit graph structure through cross-modal alignment, graph-aware tokenization, and auxiliary encoders. Unlike standalone LLM methods, these hybrid architectures use graph models to inform or condition LLMs, facilitating structure-aware language understanding. A common pipeline encodes graphs via a GNN and projects node embeddings into the LLM token space:
\begin{equation}
    \vo = \operatorname{LLM}(\gC(\gG)), \quad \gC(\gG) = [\mathbf{z}_1, ..., \mathbf{z}_n \| \mathbf{p}_1, ..., \mathbf{p}_m],
\end{equation}
where $\mathbf{z}_i = \rho(\mathbf{h}_i)$ are node embeddings transformed via a projector $\rho$, and $\mathbf{p}_j$ are trainable or handcrafted prompt tokens. This enables the LLM to process structural features in a token-compatible format. GraphGPT~\cite{tang2023graphgpt} adopts this approach, applying instruction tuning and contrastive learning to align graph and language representations. GraphTranslator~\cite{zhang2024graphtranslator} introduces structure-aware regularization to reduce semantic drift between modalities. LLaGA~\cite{chenllaga} further augments reasoning with a retrieval module that extracts informative subgraphs $\mathcal{G}_s \subset \mathcal{G}$ based on task relevance, ensuring the LLM focuses on salient relational patterns. Although these methods are often categorized as LLM-centric, they fundamentally rely on GNN components, and are therefore included in the GNN+LLM hybrid paradigm described in the following sections.

\subsubsection{Graph-Language Co-Training}

Graph-Language Co-Training offers a bidirectional learning framework that integrates graph structural modeling and language-based semantic reasoning. Unlike auxiliary-based approaches where one modality dominates, co-training treats graph and language models as co-equal components, encouraging mutual representation learning. Two major paradigms define this space: (i) \textit{Graph-Language Alignment}, where shared latent spaces are enforced via contrastive learning, and (ii) \textit{Graph-Language Iterative Update}, where graph and language representations are jointly optimized through multi-stage or variational objectives.

\noindent\textbf{Graph-Language Alignment.} This approach maps structured and unstructured representations into a shared embedding space. Inspired by CLIP-like training from vision-language modeling \cite{radford2021learning}, GraphCLIP~\cite{zhu2024graphclip} introduces a dual-encoder framework comprising a GNN encoder $f^{\text{GNN}}$ and a language encoder $f^{\text{LLM}}$. Given a graph $\mathcal{G} = (\mathcal{V}, \mathcal{E}, \mathbf{X})$, and corresponding node descriptions, representations are learned via contrastive loss:
\begin{equation}
    \mathcal{L}_{\text{clip}} = - \sum_{(v_i, v_j) \in \mathcal{P}} \log \frac{\exp (\text{sim}(\mathbf{h}_{v_i}^{\text{GNN}}, \mathbf{h}_{v_j}^{\text{LLM}})/\tau)}
    {\sum_{w \in \mathcal{V}} \exp (\text{sim}(\mathbf{h}_{v_i}^{\text{GNN}}, \mathbf{h}_w^{\text{LLM}})/\tau)},
\end{equation}
where $\text{sim}(\cdot,\cdot)$ denotes similarity (e.g., cosine), $\mathcal{P}$ is the set of aligned node-text pairs, and $\tau$ is a temperature parameter. ConGraT~\cite{brannon2023congrat} extends this idea with dual encoders trained via cross-modal supervision, where the GNN reinforces text-derived features. These methods improve generalization across graph-text pairs by unifying their representation spaces.

\noindent\textbf{Graph-Language Iterative Update.} Beyond contrastive alignment, iterative co-training enforces dynamic interaction between graph and language models through successive refinement. GLEM~\cite{zhao2022learning} exemplifies this by introducing a latent variable $\mathbf{z}_v$ representing the shared semantics of textual and structural inputs. The generative process is modeled as:
\begin{equation}
    p(\mathbf{h}_v | \mathcal{G}) = \int p(\mathbf{h}_v | \mathbf{z}_v, \mathcal{G}) \cdot p(\mathbf{z}_v | \mathcal{G}) \, d\mathbf{z}_v,
\end{equation}
where $\mathbf{z}_v$ serves as a bridge for integrating modalities. The model alternates between updating semantic features via textual pseudo-labeling and refining topological embeddings through graph propagation, following an EM-style optimization. Such iterative mechanisms offer more expressive fusion of modalities, enabling the co-evolution of graph and language representations and enhancing performance on text-attributed graphs.

\subsubsection{Discussion}

The backbone architectures of Graph Foundation Models can be broadly classified into three categories: \textit{Graph Models}, \textit{Language Models}, and \textit{Hybrid Models}. In summary, graph-based backbones offer structural fidelity and efficiency; language-based backbones provide generalization and multimodal flexibility; and hybrid models enable comprehensive reasoning but demand more sophisticated training pipelines.

\begin{itemize}[leftmargin=10pt]
    \item \textbf{Graph-Based Backbones} (e.g., GNNs, Graph Transformers) are inherently aligned with the structure of graph data. By explicitly modeling node neighborhoods through message-passing or attention-based aggregation, these architectures preserve local connectivity and relational information. They are highly effective for tasks such as node classification, link prediction, and graph classification, where structural dependencies are critical. Furthermore, these models tend to be more parameter-efficient when compared to language-based counterparts. However, they face several limitations: first, their inductive bias toward locality makes it difficult to capture long-range dependencies; second, integrating textual or multimodal information remains non-trivial, often requiring auxiliary encoders or ad-hoc fusion mechanisms.

    \item \textbf{Language-Based Backbones} leverage LLMs by encoding graph components—such as node attributes, edge descriptions, or subgraph patterns—as natural language. This graph-to-text formulation enables the transfer of powerful language understanding capabilities to graph learning tasks. Language-based models are especially useful when graphs are enriched with textual metadata, such as in knowledge graphs, social networks, or biomedical corpora. Their generalization ability also makes them suitable for zero-shot and few-shot learning scenarios. Nevertheless, these models lack explicit structural inductive biases and rely on sequential representations, which can obscure topological nuances and lead to suboptimal performance in tasks requiring precise relational reasoning.

    \item \textbf{Hybrid Backbones} integrate graph-based and language-based architectures, aiming to combine the best of both worlds. These models typically involve dual encoders (e.g., GNNs and LLMs) that interact through co-training, cross-modal attention, or alignment objectives. Hybrid approaches have shown impressive performance across diverse tasks by jointly modeling structure and semantics. For instance, graph encoders can capture the connectivity skeleton, while LLMs encode descriptive context or domain-specific knowledge. However, this synergy comes at a cost: hybrid models often require complex architecture design, increased memory and compute overhead, and careful pretraining or alignment strategies to avoid modality collapse or overfitting.
\end{itemize}

\subsection{Pretraining Strategies}

\begin{figure*}[!t]
    \centering
    \includegraphics[width=\linewidth]{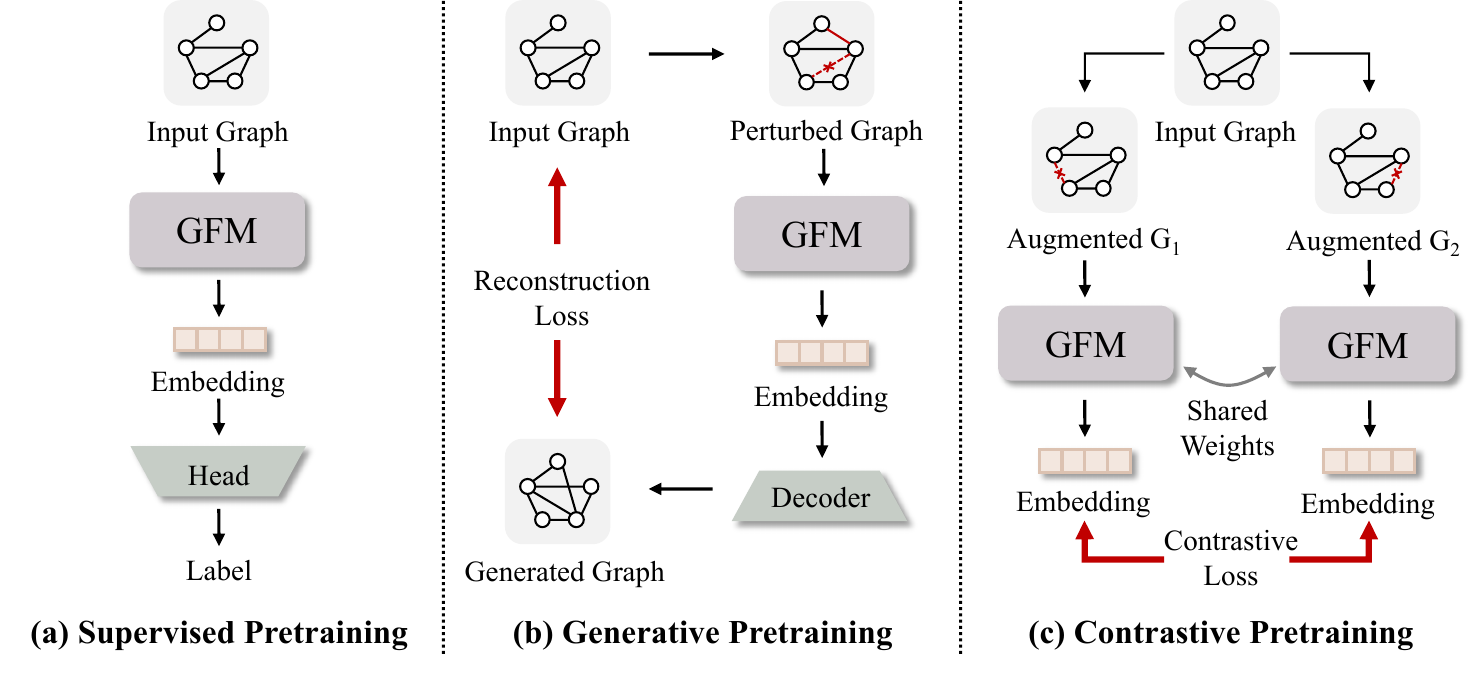}
    \caption{
        \textbf{Pretraining Strategies of Graph Foundation Models.}
        \textbf{(a) Supervised Pretraining:} Models are pretrained using labeled subgraphs derived from large-scale graphs, where task-specific labels guide the learning process via supervised objectives. \textbf{(b) Generative Pretraining:} Models learn to reconstruct masked or corrupted parts of the graph, such as node features or adjacency information, often using autoregressive or autoencoding paradigms to capture high-order dependencies. \textbf{(c) Contrastive Pretraining:} Models are trained to distinguish between similar (positive) and dissimilar (negative) node or subgraph pairs across different views or augmentations of the graph, thereby learning invariant and transferable representations.
    }
    \label{fig:pretrain}
\end{figure*}

Pretraining serves as a foundational step in the development of GFMs, enabling them to acquire transferable knowledge from large-scale unlabeled or weakly-labeled data. In this section, we provide a comprehensive overview of key pretraining paradigms for GFMs, including \textit{supervised pretraining}, \textit{generative pretraining}, \textit{contrastive pretraining}, as illustrated in Figure \ref{fig:pretrain}. For detailed information, we recommend to read the awesome surveys \cite{liu2022graph,wu2021self}.

\subsubsection{Supervised Pretraining}

Supervised pretraining is a strategy that leverages labeled graph data to guide the model pretraining. Unlike self-supervised methods that rely solely on intrinsic graph signals, supervised pretraining explicitly optimizes the model to predict known labels, encouraging it to learn task-relevant representations from the outset. Formally, let $\mathcal{G}_{\text{pt}}$ denote a large-scale pretraining graph with associated supervision $\mathcal{Y}$, which may correspond to node-level, edge-level, or graph-level labels. The model is trained to minimize a supervised loss function over these labeled graph instances:
\begin{equation}
    \theta^* = \arg\min_{\theta} \mathcal{L}_{\text{pt}}(f(\mathcal{G_{\text{pt}}}; \theta), \mathcal{Y}),
\end{equation}
where $\mathcal{L}_{\text{pt}}$ is typically a task-specific loss, such as cross-entropy for classification and mean-average error for regression. A core benefit of supervised pretraining is its direct alignment with downstream objectives, which often results in faster convergence and improved performance on similar tasks. Representative methods following this paradigm include {OFA}~\cite{liu2024one}, which introduces a unified task formulation using nodes-of-interest to standardize various graph prediction tasks, and {Prodigy}~\cite{huang2023prodigy}, which generates subgraph tasks through structural and semantic decomposition. These approaches illustrate the flexibility of supervised pretraining in capturing both local and global signals across graph domains.

Despite its effectiveness, supervised pretraining is often limited by the availability and cost of acquiring large-scale, high-quality labels \cite{wu2021self}. As a result, it is frequently complemented by self-supervised learning to enhance generalization and reduce reliance on labeled data.

\subsubsection{Generative Pretraining}

Generative pretraining is a foundational learning paradigm in foundation models. The core philosophy behind generative pretraining is to learn universal representations by training models to predict or generate data in its raw form without requiring task-specific labels. This approach assumes that by modeling the data distribution itself, the model acquires broad, transferable knowledge that can be adapted to a wide range of downstream tasks via fine-tuning or prompting. Formally, generative pretraining aims to optimize the likelihood of observed data $\mathcal{D} = \{x_1, x_2, \dots, x_N\}$ under a parameterized model $f_\theta$:
\begin{equation}
    \theta^* = \argmax_{\theta} \sum_{i=1}^{N} \log p(x_i; \theta),
\end{equation}
where $x_i$ represents an individual data sample and $p(x_i; \theta)$ denotes the estimated probability of generating $x_i$. In practice, this objective is often instantiated through auto-regressive or auto-encoder objectives, as seen in models like GPT~\cite{brown2020language} and BERT~\cite{devlin2019bert}. We discuss these two approaches in the following.

\noindent\textbf{Auto-Regressive Generation.}
Autoregressive modeling is a foundational paradigm in generative learning, widely applied in domains such as natural language processing \cite{brown2020language} and computer vision \cite{bai2024sequential}. In this framework, the model learns a joint probability distribution over sequential variables by decomposing it into a product of conditional probabilities. Formally, for a sequence of variables $\mathbf{x} = (x_1, x_2, \dots, x_T)$, an autoregressive model parameterized by $\theta$ defines the joint likelihood as:
\begin{equation}
    p_\theta(\mathbf{x}) = \prod_{t=1}^T p_\theta(x_t \mid \mathbf{x}_{<t}),
\end{equation}
where $\mathbf{x}_{<t}$ denotes all preceding elements in the sequence. This formulation aligns with the structure of Bayesian networks~\cite{liu2021self}, where each variable is conditionally dependent on its predecessors. Autoregressive objectives have been central to next-token prediction in NLP (e.g., GPT~\cite{brown2020language}) and patch generation in vision models \cite{bai2024sequential}.

In the graph domain, autoregressive models extend this concept to structured data by mapping graphs into sequences through node ordering schemes. Given an undirected graph $\mathcal{G} = (\mathbf{X}, \mathbf{A})$ with $n$ nodes and a node permutation $\pi$, the graph can be transformed into a sequence $\mathbf{S}^\pi = f_S(\mathcal{G}, \pi) = (\mathbf{S}_1^\pi, \dots, \mathbf{S}_n^\pi)$, where each element $\mathbf{S}_i^\pi$ encodes the connectivity (i.e., edges) between node $\pi(v_i)$ and its preceding nodes $\pi(v_j), j < i$ in the permutation. The generation process then follows an autoregressive factorization:
\begin{equation}
    p(\mathbf{S}^\pi) = \prod_{i=1}^{n+1} p(\mathbf{S}_i \mid \mathbf{S}_{<i}),
    \label{eq: auto-regression sequence}
\end{equation}
where a new node and its connections are generated conditioned on the current graph prefix. At each step $i$, the model maintains a hidden state $h_i$ that summarizes the generation history, and uses it to parameterize the output distribution for the next adjacency slice $\mathbf{S}_i^\pi$:
\begin{equation}
    h_i      = f_{\text{trans}}(h_{i-1}, \mathbf{S}_{i-1}^\pi), \quad \theta_i = f_{\text{out}}(h_i),
\end{equation}
where $f_{\text{trans}}$ is typically implemented using recurrent neural networks (e.g., GRUs~\cite{cho2014learning}, LSTMs~\cite{graves2012long}) or Transformers~\cite{vaswani2017attention}, and $f_{\text{out}}$ defines the distributional output head (e.g., for edge prediction). This architecture enables autoregressive GFMs to sequentially generate nodes and edges in a structured and history-aware manner. Notable examples of this approach include GraphRNN~\cite{you2018graphrnn}, MolecularRNN~\cite{popova2019molecularrnn}, and GraphGPT~\cite{hu2020gpt}, which utilize autoregressive mechanisms for graph generation. Moreover, beyond node-centric generation, several works~\cite{Bacciu_2020, goyal2020graphgen} extend this framework to edge-wise autoregression, preserving the same conditional structure as Equation~\ref{eq: auto-regression sequence}. These methods demonstrate the versatility of autoregressive models in capturing the sequential dependencies embedded within complex graph structures.

\noindent\textbf{Auto-Encoding Generation.}
The auto-encoder paradigm, particularly in the context of foundation models, has evolved from classical reconstruction objectives \cite{kingma2013auto,pu2016variational,kipf2016variational} to masked modeling strategies \cite{devlin2019bert,he2022masked}. Inspired by the success of models like BERT in natural language processing~\cite{devlin2019bert}, the underlying philosophy centers on reconstructing masked or corrupted parts of the input conditioned on the visible context. Formally, given an input sequence $\mathbf{x} = (x_1, x_2, \dots, x_T)$ with a subset of tokens masked, the objective is to maximize the likelihood of the masked components given the observed ones:
\begin{equation}
    p_\theta(\mathbf{x}) = \prod_{t \in \mathcal{M}} p_\theta(x_t \mid \mathbf{x}_{\setminus \mathcal{M}}),
\end{equation}
where $\mathcal{M} \subset \{1, \dots, T\}$ is the set of masked indices, and $\mathbf{x}_{\setminus \mathcal{M}}$ denotes the visible (unmasked) tokens. This conditional modeling objective enables the model to learn contextualized and robust representations by leveraging co-occurrence patterns within the input.

Extending this idea to graphs, Graph Auto-Encoders (GAEs) \cite{kipf2016variational} adopt a similar masked modeling philosophy. Rather than reconstructing the entire graph or feature space directly, the model is trained to predict masked components of a graph—such as node features, edge features, or structure—conditioned on the visible graph context \cite{kipf2016variational,hou2022graphmae,hou2023graphmae2,tan2023s2gae,tian2023heterogeneous,liu2023rethinking,xiamole,hu2020gpt}. Let $\mathcal{G} = (\mathbf{X}, \mathbf{A})$ denote an attributed graph, and let $t \sim \mathcal{T}$ be a masking or augmentation function that produces a corrupted version $\widetilde{\mathcal{G}} = (\widetilde{\mathbf{X}}, \widetilde{\mathbf{A}})$. A GNN-based encoder is applied to the corrupted graph to yield latent representations $\widetilde{\mathbf{Z}} = f_{\text{GNN}}(\widetilde{\mathcal{G}}; \theta)$, which are then used to reconstruct the masked components through specialized prediction heads.

For \emph{masked feature modeling}, the objective is to reconstruct the masked node or edge attributes:
\begin{equation}
    \theta^* = \arg\min_{\theta} \mathcal{L}_{\text{pt}}\left( p(\widetilde{\mathbf{Z}}), \mathbf{X}_{\mathcal{M}} \right),
\end{equation}
where $\mathbf{X}_{\mathcal{M}}$ denotes the masked features and $p(\cdot)$ is the prediction head.

For \emph{masked structure modeling}, the goal is to recover the masked links or adjacency entries:
\begin{equation}
    \theta^* = \arg\min_{\theta} \mathcal{L}_{\text{pt}}\left( p(\widetilde{\mathbf{Z}}), \mathbf{A}_{\mathcal{M}} \right),
\end{equation}
where $\mathbf{A}_{\mathcal{M}}$ represents the masked structural components.

\subsubsection{Contrastive Pretraining}

Contrastive pretraining is a self-supervised learning paradigm that focuses on learning discriminative representations by contrasting positive pairs against negative pairs \cite{chen2020simple,he2020momentum,oord2018representation,grill2020bootstrap,caron2020unsupervised,hjelm2018learning,qian2024dual,tian2020contrastive}. The underlying philosophy is to bring semantically similar representations closer in the embedding space while pushing apart dissimilar ones \cite{wang2020understanding,tian2020makes,khosla2020supervised}, thereby encouraging the model to capture meaningful and generalizable features without relying on manual annotations. Let $\mathcal{D} = \{x_i\}_{i=1}^N$ be a dataset, and let $(x_i, x_i^+)$ denote a positive pair (e.g., different views or augmentations of the same instance), while $(x_i, x_j^-)$ represents a negative pair where $x_j^- \neq x_i$. A widely used objective is the InfoNCE loss~\cite{oord2018representation}, formulated as:
\begin{equation}
    \mathcal{L}_{\text{InfoNCE}} = -\sum_{i=1}^{N} \log \frac{\exp(\text{sim}(f(x_i), f(x_i^+))/\tau)}{\sum_{j=1}^{N} \exp(\text{sim}(f(x_i), f(x_j^-))/\tau)},
    \label{eq:infonce}
\end{equation}
where $f(\cdot)$ is the encoder network, $\text{sim}(\cdot,\cdot)$ denotes a similarity function (e.g., cosine similarity), and $\tau$ is a temperature parameter. By optimizing this objective, the model learns to distinguish between semantically relevant and irrelevant samples, leading to representations that transfer well to a variety of downstream tasks. Based on the contrastive levels, we categorize the existing contrastive pretraining methods into \textit{instance-instance} and \textit{instance-context} contrastive methods.

\noindent\textbf{Instance-Instance Contrastive Learning.}
Instance-instance contrastive learning focuses on comparing individual instances against one another. The key idea is to encourage representations of the same instance under different views (positive pairs) to be close in the embedding space, while pushing apart representations of different instances (negative pairs). This stands in contrast to instance-to-context learning, which aims to align an instance with a broader semantic context (e.g., prototypes, clusters, or class-level embeddings) \cite{velickovic2018deep,caron2020unsupervised}. The instance-instance paradigm treats every instance as its own class, leading to instance-discriminative embeddings that are highly generalizable across downstream tasks.

In the graph domain, many works \cite{you2020graph,zhu2020deep,zhu2021graph,wang2024gft,wang2023heterogeneous,wang2024select,hassani2020contrastive,thakoor2022largescale} illustrates the standard pipeline for instance-instance contrastive learning over graph-structured data. The typical workflow begins by applying two stochastic augmentations $t_1, t_2 \sim \mathcal{T}$ to a given graph $\mathcal{G} = (\mathbf{X}, \mathbf{A})$, producing two views $\widetilde{\mathcal{G}}_1$ and $\widetilde{\mathcal{G}}_2$:
\begin{equation}
    \widetilde{\mathcal{G}}_* = (\widetilde{\mathbf{X}}_*, \widetilde{\mathbf{A}}_*) = t_*(\mathbf{X}, \mathbf{A}), \quad \mathbf{Z}_* = f(\widetilde{\mathcal{G}}_*; \theta),
\end{equation}
where $*$ denotes the view index (either 1 or 2), and $\mathbf{Z}_*$ contains the node-level embeddings for the corresponding augmented graph.

For each node $v_i \in \mathcal{G}$, a set of positive counterparts $\mathbb{P}(v_i)$ is selected—typically consisting of the same node under the alternate view. The training objective aims to maximize agreement between positive pairs while minimizing it for all others using InfoNCE loss~\cite{oord2018representation} in Equation \ref{eq:infonce}.

When negative samples are not explicitly used—as in BGRL~\cite{thakoor2021large}—a bootstrapped alternative is employed, which avoids direct contrast with negative instances. This bootstrapping loss is defined as:
\begin{equation}
    \mathcal{L}_{\text{Bootstrap}} = -\sum_{i=1}^{N} \text{sim}(f(x_i), \operatorname{sg}g([f(x_i^+)])),
    \label{eq:bootstrap}
\end{equation}
where $g(\cdot)$ is a projector and $\operatorname{sg}[\cdot]$ denotes the stop-gradient operation.

While instance-instance contrastive learning has demonstrated impressive results in graph representation learning, a notable limitation arises from the assumption that the positive sample is simply the same node under different augmentations. This assumption can introduce sampling bias, especially in graphs with noisy structure or weak homophily. Addressing the challenge of positive sample selection and exploring more semantically aligned contrastive pairs remain important directions for improving the robustness and generality of graph contrastive learning methods \cite{jin2020self}.

\noindent\textbf{Instance-Context Contrastive Learning.} Beyond instance-instance contrastive learning, an alternative and complementary paradigm is \emph{instance-context contrastive learning}, which aims to maximize mutual information between local (instance-level) and global (context-level) representations \cite{caron2020unsupervised,hassani2020contrastive,jiao2020sub,caron2018deep}. Unlike the instance-instance approach—which distinguishes between individual views of nodes or subgraphs—instance-context contrastive learning encourages alignment between a node (or substructure) and a global summary of the graph, facilitating the capture of holistic graph-level semantics.

In the graph domain, this idea was first introduced by Deep Graph Infomax (DGI)~\cite{velickovic2018deep}, which adapts the Deep InfoMax framework~\cite{hjelm2019learningdeeprepresentationsmutual} to graph-structured data. DGI proposes to maximize the local-global mutual information between node representations and a global summary vector, thereby enabling the encoder to learn node embeddings that are contextually grounded in the structure and semantics of the entire graph. Formally, given a graph $\mathcal{G} = (\mathbf{X}, \mathbf{A})$ and a stochastic augmentation $t \sim \mathcal{T}$ that produces a corrupted graph $\widetilde{\mathcal{G}} = (\widetilde{\mathbf{X}}, \widetilde{\mathbf{A}})$, an encoder $f(\cdot)$ is used to map both the original and corrupted graphs into latent space:
\begin{equation}
    \mathbf{Z}             = f(\mathcal{G}; \theta), \quad \widetilde{\mathbf{Z}} = f(\widetilde{\mathcal{G}}; \theta),
\end{equation}
where $\mathbf{Z} = [\mathbf{z}_1, \dots, \mathbf{z}_N]$ and $\widetilde{\mathbf{Z}} = [\widetilde{\mathbf{z}}_1, \dots, \widetilde{\mathbf{z}}_M]$ denote the node embeddings from the original and corrupted graphs, respectively. To compute a graph-level summary vector, a permutation-invariant readout function $\mathcal{R}(\cdot)$ is applied over the original node embeddings $\mathbf{s} = \mathcal{R}(\mathbf{Z})$, where $\mathbf{s}$ serves as the global context embedding for the graph.

The contrastive objective is then defined to distinguish between positive pairs $(\mathbf{z}_i, \mathbf{s})$ sampled from the original graph and negative pairs $(\widetilde{\mathbf{z}}_j, \mathbf{s})$ from the corrupted one. A discriminator $\mathcal{D}(\cdot, \cdot)$ is trained to assign high scores to positive pairs and low scores to negatives. The training objective is:
\begin{equation}
    \mathcal{L}_{\text{DGI}} = -\frac{1}{N + M} \left(
    \sum_{i=1}^{N} \log \mathcal{D}(\mathbf{z}_i, \mathbf{s}) +
    \sum_{j=1}^{M} \log \left(1 - \mathcal{D}(\widetilde{\mathbf{z}}_j, \mathbf{s}) \right)
    \right),
    \label{eq:dgi}
\end{equation}
where $N$ and $M$ are the number of nodes in the original and corrupted graphs, respectively. This local-global objective encourages the model to learn representations that are not only locally expressive but also globally aware. Recent extensions have improved upon DGI by integrating additional structural priors, such as structural mutual information from an information bottleneck perspective~\cite{zhao2023deep}, or by tailoring the approach for low-resource settings like few-shot node classification~\cite{electronics12010239}.

\subsubsection{Discussion}

The pretraining of GFMs has emerged as a critical driver for improving generalization and transferability across diverse graph learning tasks. While various pretraining paradigms have been developed, each presents distinct benefits and trade-offs depending on the underlying assumptions, supervision signals, and application scenarios. In summary, supervised pretraining provides strong task alignment but requires labeled data; generative approaches enable flexible and scalable learning through reconstruction objectives; and contrastive methods deliver powerful, discriminative representations but rely heavily on augmentation quality and require intensive computation. No single method is universally superior—rather, each presents complementary advantages.

\begin{itemize}[leftmargin=10pt]
    \item \textbf{Supervised Pretraining} directly optimizes the model with respect to labeled graph data, often resulting in task-aligned and semantically rich representations. This approach is particularly effective when the pretraining task closely matches the downstream objectives, as it enables faster convergence and better performance. However, the main limitation lies in the reliance on large-scale, high-quality labeled datasets, which are costly and time-consuming to obtain. This constraint limits the scalability and generality of supervised pretraining.

    \item \textbf{Generative Pretraining}, including auto-regressive modeling and auto-encoding modeling, encourages models to learn the underlying structure of graph data by reconstructing input components. These methods leverage only the input data itself, enabling label-free pretraining at scale. The key strength of generative approaches is their flexible and scalable framework, which is efficient than contrastive learning. However, generative models may lack strong task specificity, and their reconstruction-based objectives do not always correlate with downstream performance.

    \item \textbf{Contrastive Pretraining} has gained popularity due to its simplicity and effectiveness in learning discriminative graph representations. Instance-instance contrastive learning excels at producing fine-grained embeddings by distinguishing between positive and negative pairs, while instance-context contrastive learning focuses on aligning local representations with global summaries. These methods often outperform generative models on graphs \cite{hou2022graphmae}. Nonetheless, contrastive methods are sensitive to the choice of augmentations, sampling strategies, and the availability of informative negatives. Additionally, computing contrastive loss can be computationally intensive, especially when modeling large or densely connected graphs.
\end{itemize}

\subsection{Adaptation} 

In this section, we provide a comprehensive overview of adaptation strategies for GFMs, highlighting how pretrained models can be effectively applied to diverse downstream tasks. It categorizes these strategies into six key paradigms—transfer learning, distillation, test-time adaptation, graph prompting, in-context learning, and prototype learning—each tailored to specific data conditions, supervision levels, and deployment constraints. 

\subsubsection{Transfer Learning}

Transfer learning is a foundational paradigm in modern machine learning, wherein a model pre-trained on a large source dataset is adapted to a new target domain with limited labeled data. In the context of GFMs, transfer learning enables the reuse of structural and semantic knowledge acquired during large-scale pretraining, thus improving performance and generalization on downstream graph-related tasks. The key idea is to initialize the model with pre-trained parameters, thereby leveraging learned representations and reducing the risk of overfitting—particularly beneficial in scenarios with limited task-specific data. Depending on the adaptation strategy, the pre-trained GFM may be directly applied to a downstream task or further fine-tuned using task-specific supervision. Formally, let $\theta$ denote the parameters of a pre-trained GFM, and let $D_{\text{target}}$ be the target dataset comprising graph-structured instances. Fine-tuning optimizes the model parameters by minimizing the task-specific loss:
\begin{equation}
    \theta^* = \arg\min_{\theta} \mathcal{L}_{\text{target}}(f_{\text{GFM}}(\mathbf{A}, \mathbf{X}; \theta), D_{\text{target}}),
\end{equation}
where $f_{\text{GFM}}$ is the GFM model operating on node features $\mathbf{X}$ and adjacency matrix $\mathbf{A}$.

\noindent\textbf{Direct Adaptation.}
In direct adaptation, the pre-trained GFM is applied to the downstream task without any parameter updates. This zero-shot setting evaluates the generalization ability of the pre-trained model in a plug-and-play fashion. For instance, OFA~\cite{liu2024one} introduces a unified framework that transforms graph data into textual prompts and leverages pre-trained LLMs for graph inference. It constructs a prompt graph $\mathcal{P} = (\mathcal{V}_p, \mathcal{E}_p, \mathcal{R}_p)$ by appending a virtual prompt node and its relations to the original graph, enabling in-context learning without task-specific fine-tuning.

\begin{figure*}[!t]
    \centering
    \includegraphics[width=\linewidth]{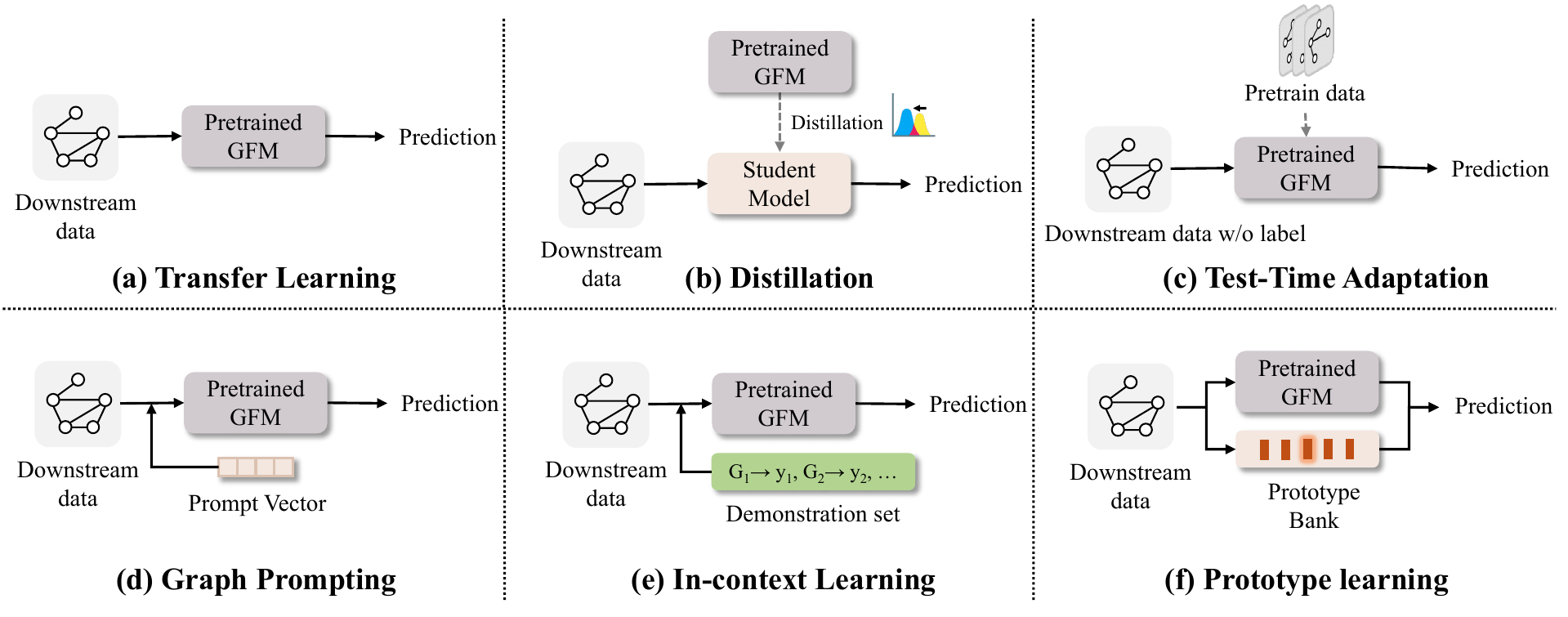}
    \caption{
        \textbf{Adaptation Strategies for Graph Foundation Models.}
        We illustrate six representative paradigms for adapting pretrained GFMs to downstream tasks:
        \textbf{(a)} \textit{Transfer Learning} via fine-tuning;
        \textbf{(b)} \textit{Distillation} to a smaller student model;
        \textbf{(c)} \textit{Test-Time Adaptation} without labeled data;
        \textbf{(d)} \textit{Graph Prompting} through learnable prompt vectors;
        \textbf{(e)} \textit{In-Context Learning} using demonstration sets; and
        \textbf{(f)} \textit{Prototype Learning} via class prototype alignment.
        These strategies support different levels of supervision, generalization, and computational efficiency.
    }
    \label{fig:adaptation}
\end{figure*}

\noindent\textbf{Full Fine-Tuning.}
Full fine-tuning involves updating all parameters of the pre-trained model on the target task. This approach offers maximal flexibility and adaptation capacity but can lead to overfitting when the target dataset is small or noisy. It is typically employed when the downstream task is sufficiently different from the pretraining objectives or when high task-specific accuracy is critical \cite{zhuang2020comprehensive}.

\noindent\textbf{Adaptive Fine-Tuning.}
Adaptive fine-tuning seeks to selectively update specific layers, modules, or parameters in the model, guided by the characteristics of the target task. This selective tuning reduces computational cost and mitigates overfitting. {AUX-TS}~\cite{han2021adaptive} exemplifies this paradigm by dynamically selecting auxiliary tasks based on their semantic similarity to the target task. The similarity scores are learned and used to weigh each auxiliary signal during the fine-tuning process, enabling task-aware adaptation.

\noindent\textbf{Parameter-Efficient Fine-Tuning (PEFT).}
PEFT techniques aim to retain the generalization capabilities of pre-trained models while significantly reducing the number of trainable parameters. These methods typically freeze the backbone model and introduce lightweight modules such as adapters or low-rank transformations. For example, {AdapterGNN}~\cite{li2024adaptergnn} integrates an adapter module into GNN layers, where the transformation is defined as:
\begin{equation}
    \mathbf{A}(\mathbf{x}) = \text{BN}(\mathbf{W}_{\text{up}} \cdot \text{ReLU}(\mathbf{W}_{\text{down}} \cdot \mathbf{x})),
\end{equation}
with $\mathbf{W}_{\text{down}}$ and $\mathbf{W}_{\text{up}}$ being trainable low-dimensional projections and $\text{BN}$ denoting batch normalization. Similarly, {GraphLoRA}~\cite{yang2024graphlora} applies low-rank adaptation to reduce parameter complexity, and \textsc{GPF}~\cite{fang2023universal} introduces a learnable task-specific vector $\mathbf{p}$, which is concatenated to node features $\mathbf{x}_i$, while freezing the backbone GNN.

\subsubsection{Distillation}

Knowledge distillation is a model compression technique aimed at transferring the knowledge encoded in a large, powerful model (the \emph{teacher}) to a smaller, more efficient model (the \emph{student}). In the context of GFMs, distillation enables the deployment of compact models that maintain competitive performance while reducing inference time and resource consumption. Let $f_{\text{teacher}}$ and $f_{\text{student}}$ denote the outputs of the teacher and student models, respectively. The distillation objective typically combines supervision from ground-truth labels with guidance from the teacher outputs. This can be formalized as:
\begin{equation}
    \mathcal{L}_{\text{distillation}} =
    \lambda \, \mathcal{L}_{\text{true}}(f_{\text{student}}(\mathbf{A}, \mathbf{X}), D_{\text{target}}) +
    (1 - \lambda) \, \mathcal{L}_{\text{match}}(f_{\text{student}}(\mathbf{A}, \mathbf{X}), f_{\text{teacher}}(\mathbf{A}, \mathbf{X})),
\end{equation}
where $\mathcal{L}_{\text{true}}$ denotes the supervised loss with respect to the ground-truth labels, $\mathcal{L}_{\text{match}}$ measures the discrepancy between the student and teacher outputs (e.g., using Kullback-Leibler divergence in general), and $\lambda \in [0, 1]$ balances the contributions of the two terms. The rationale behind distillation is that the teacher outputs—often referred to as "soft targets"—contain richer supervisory signals than one-hot labels \cite{gou2021knowledge}. These include information about inter-class similarities and decision boundaries that are difficult for a student model to infer directly from the data \cite{hinton2015distilling}. Moreover, intermediate representations within the teacher model (e.g., node embeddings or attention maps) can be used to further enhance the student's learning through feature-level alignment \cite{wang2024training}.

In the graph domain, distillation techniques have been extended to incorporate structural and relational knowledge. For instance, \textsc{G-CRD}~\cite{joshi2022representation} proposes a contrastive representation distillation framework that preserves global graph topology. Instead of only matching final predictions, G-CRD aligns student and teacher node embeddings by maximizing agreement in a shared representation space, effectively transferring topological cues. Beyond prediction-level and embedding-level supervision, additional distillation strategies include:
\begin{itemize}[leftmargin=10pt]
    \item \textbf{Graph Structure Distillation:} Preserving relational patterns such as adjacency, edge importance, or motif distributions in the student model \cite{zhang2021graph,tian2022learning,wang2024training}.
    \item \textbf{Attention-Based Distillation:} Mimicking the attention maps learned by graph attention models (e.g., GAT) to preserve neighbor importance \cite{lee2019graph,wu2022knowledge}.
    \item \textbf{Multi-View or Multi-Task Distillation:} Transferring knowledge from auxiliary tasks or diverse views of the graph to improve robustness \cite{ma2019graph,ren2021multi,liu2025learning}.
\end{itemize}

\subsubsection{Test-Time Adaptation}

Test-Time Adaptation (TTA) \cite{wang2020tent,liang2025comprehensive} refers to the process of adapting a pre-trained GFM during inference, using the test data itself to update the model in real time. Unlike fine-tuning and distillation, which are performed in a dedicated training phase prior to deployment, TTA operates entirely at inference time. This paradigm is particularly useful in scenarios involving distributional shifts \cite{chen2022contrastive} between training and test data or where access to labeled target-domain samples is limited or unavailable. TTA typically involves adjusting model parameters on-the-fly in an online fashion \cite{boudiaf2022parameter}, as each new test sample or batch is processed. Let $\mathcal{G}_{\text{test}}$ denote a test graph and $f_\theta$ the pre-trained GFM with parameters $\theta$. The model is adapted by minimizing a self-supervised loss function defined over the test data:
\begin{equation}
    \theta' = \theta - \eta \nabla_\theta \mathcal{L}_{\text{self}}(f_\theta(\mathcal{G}_{\text{test}})),
\end{equation}
where $\eta$ is the learning rate and $\mathcal{L}_{\text{self}}$ represents a self-supervised objective, such as entropy minimization, pseudo-label consistency, or structural smoothness. These objectives do not require labeled data and are often chosen to encourage confident, stable predictions under test-time perturbations.

The key assumption behind TTA is that the incoming test data itself contains useful signals that can help tailor the model to the target distribution. By leveraging this data for real-time adaptation, the model can dynamically adjust to distributional shifts, enhance robustness, and improve prediction accuracy without requiring re-training or access to source-domain data.

\noindent\textbf{Graph Transformation-Based Adaptation.}
\textsc{GTrans}~\cite{jin2022empowering} exemplifies a data-centric TTA approach by performing graph refinement at test time. Rather than updating model parameters directly, GTrans modifies the input graph to better suit the fixed, pre-trained GNN. It learns perturbations on node features and graph topology to minimize a surrogate loss, effectively generating an adapted graph $\mathcal{G}'$ that is more compatible with the pretrained model:
\begin{equation}
    \mathcal{G}' = \mathcal{G}_{\text{test}} + \delta^*, \quad \delta^* = \arg\min_\delta \mathcal{L}_{\text{surrogate}}(f_\theta(\mathcal{G}_{\text{test}} + \delta)),
\end{equation}
where $\delta$ denotes learnable perturbations and $\mathcal{L}_{\text{surrogate}}$ approximates task-specific losses without requiring ground-truth labels. This method improves performance by refining the input data rather than the model.

\noindent\textbf{Test-Time Supervision.}
Recently, \textsc{LLM-TTT}~\cite{zhang2024test} introduces a novel TTA paradigm that leverages the generative and annotative capabilities of LLMs to aid inference on text-attributed graphs. In this framework, LLMs are used to generate pseudo-labels for unlabeled test nodes, which are then used to adapt the GNN at test time. The two-stage pipeline consists of (i) annotation by the LLM using node descriptions and graph context, and (ii) refinement of the GNN using the generated pseudo-labels:
\begin{equation}
    \mathbf{\hat{y}}_v = \text{LLM}(\text{Prompt}(v)), \quad \theta' = \theta - \eta \nabla_\theta \mathcal{L}_{\text{pseudo}}(f_\theta(\mathcal{G}_{\text{test}}), \hat{\mathbf{y}}).
\end{equation}
This strategy demonstrates the synergy between language-based supervision and graph-based reasoning, improving performance under test-time constraints without requiring manual annotations.


\subsubsection{Graph Prompting}

Graph prompting~\cite{fang2023universal, sun2023all, li2025instanceaware, fu2025edge} is an emerging paradigm that adapts GFMs to downstream tasks from a data-centric perspective. Unlike traditional fine-tuning, which updates the model parameters, graph prompting keeps the model frozen and instead learns additional prompt vectors $\mathcal{P}$ that guide the model behavior. This approach draws inspiration from prompting strategies in NLP, where carefully designed inputs can elicit desired behaviors from large language models. Graph prompting methods can be broadly categorized into two genres: \emph{data-level prompting}, which modifies the input graph data, and \emph{representation-level prompting}, which adjusts internal representations within the model.

\noindent\textbf{Data-Level Prompting.}
Data-level prompting adapts the input graph by injecting learnable signals either into the feature space or the structure. Given a graph $\mathcal{G} = (\mathbf{X}, \mathbf{A})$, where $\mathbf{X}$ is the node feature matrix and $\mathbf{A}$ is the adjacency matrix, data-level prompting defines a transformation function $t_D$ that produces a prompted graph $\widetilde{\mathcal{G}} = (\widetilde{\mathbf{X}}, \widetilde{\mathbf{A}})$ using prompt vectors $\mathcal{P}$:
\begin{equation}
    (\widetilde{\mathbf{X}}, \widetilde{\mathbf{A}}) = t_D((\mathbf{X}, \mathbf{A}); \mathcal{P}).
\end{equation}

A simple yet effective strategy is to modify only the node features. For each node $v \in \mathcal{V}$, a prompt vector $\mathbf{p}_v$ is learned and added to the original feature vector $\widetilde{\mathbf{x}}_v = \mathbf{x}_v + \mathbf{p}_v$. To reduce complexity, some approaches use a shared prompt vector for all nodes, i.e., $\mathbf{p}_1 = \mathbf{p}_2 = \cdots = \mathbf{p}_N = \mathbf{p}$~\cite{fang2023universal}. However, shared prompts may lack expressiveness. To address this, recent works introduce an attention-based mechanism over a set of $J$ basis vectors $\{\mathbf{b}_1, \ldots, \mathbf{b}_J\}$~\cite{fang2023universal, lee2024subgraph, li2025instanceaware, zhu2024relief, li2025fairnessaware, jiang2024unified}. Each node-specific prompt is computed as a weighted combination $\widetilde{\mathbf{x}}_v = \mathbf{x}_v + \mathbf{p}_v = \mathbf{x}_v + \sum_{j=1}^J w_{v,j} \cdot \mathbf{b}_j$, where $w_{v,j}$ denotes the learned attention weight of node $v$ over basis vector $j$.

An alternative strategy is \emph{insertion-based prompting}, which introduces learnable prompt nodes into the graph. These prompt nodes are connected to existing graph nodes either uniformly~\cite{zhu2023sgl, ge2024psp} or based on similarity metrics~\cite{sun2023all}, resulting in an enriched structure that encourages the model to focus on task-relevant subgraphs.

\noindent\textbf{Representation-Level Prompting.}
Representation-level prompting modifies latent node representations instead of the input graph. Given a hidden representation $\mathbf{h}_v$ for node $v$, a transformation function $t_R$ applies the learned prompts to obtain a prompted embedding:
\begin{equation}
    \widetilde{\mathbf{h}}_v = t_R(\mathbf{h}_v; \mathcal{P}).
\end{equation}
A common approach is to apply an element-wise (Hadamard) product between the representation and a prompt vector $\widetilde{\mathbf{h}}_v = \mathbf{p}_v \odot \mathbf{h}_v$, where $\mathbf{p}_v$ acts as a gating vector that highlights task-relevant dimensions in $\mathbf{h}_v$. Similar to data-level prompting, $\mathbf{p}_v$ may be shared across nodes~\cite{liu2023graphprompt} or generated in a node-specific manner. For example, ProNoG~\cite{yu2025non} computes prompt vectors based on multi-hop ego-networks of each node, allowing localized and personalized adaptation.

\subsubsection{In-Context Learning}

In-Context Learning (ICL) \cite{dong2022survey} is a form of few-shot learning that enables LLMs to adapt to novel tasks without parameter updates. Instead of fine-tuning, the model is conditioned on a small set of input-label pairs—known as \emph{demonstrations}—that are provided directly in the input sequence. This paradigm capitalizes on the remarkable in-context generalization capabilities of pre-trained language models, such as GPT-3~\cite{brown2020language} and its successors. Formally, let the demonstration set be denoted as $\mathcal{C}_K = \{(q_k, y_k)\}_{k=1}^K$, where each tuple $(q_k, y_k)$ represents a query and its corresponding label. Given a new query $q_v$ for a test sample $v$, the foundation model $f$ uses the demonstration set $\mathcal{C}_K$ to generate a predicted label:
\begin{equation}
    \hat{y}_v = f(\mathcal{C}_K, q_v).
\end{equation}
The model thus performs inference by conditioning on the examples in $\mathcal{C}_K$ as contextual guidance, without modifying its internal parameters.

In the context of GFMs, ICL has gained attention as a promising strategy for handling text-attributed graphs, where each graph component (e.g., node, edge, or subgraph) is associated with descriptive textual information. LLMs can process such textual attributes directly, allowing graph-based tasks to be reformulated as natural language problems solvable via prompting. However, applying ICL to TAGs introduces several challenges, most notably the construction of high-quality demonstration sets. Unlike i.i.d. samples in typical NLP settings, graph data exhibits rich structural dependencies, such as homophily, transitivity, and higher-order relations~\cite{guo2023gpt4graph}. As a result, selecting representative and informative demonstrations requires careful consideration of the underlying graph structure. To address this, {AskGNN}~\cite{hu2024lets} introduces a GNN-based retriever that selects relevant node-label pairs from the graph based on structural and semantic similarity. These selected instances are then formatted as natural language demonstrations and fed to an LLM for prediction. Similarly, retrieval-augmented generation (RAG) techniques~\cite{li2025large} have been employed to enhance demonstration quality by dynamically retrieving the most relevant graph samples from an external memory or training set. Beyond node classification, ICL has also been explored in more complex graph tasks such as \emph{knowledge graph completion}. In this setting, each sample corresponds to a triple (subject, relation, object), and ICL can be used to infer missing entities or relations \cite{sehwag2024context,khorashadizadeh2023exploring}

\subsubsection{Prototype Learning}

Prototype learning \cite{wang2020generalizing} is a classification paradigm that represents each class by a prototype vector in the embedding space and classifies instances based on proximity to these prototypes \cite{snell2017prototypical}. Unlike traditional approaches that rely on dedicated classifiers (e.g., fully connected layers or softmax heads), prototype learning performs classification by comparing instance representations with class prototypes, offering a more interpretable and often parameter-efficient alternative. In the context of GFMs, prototype learning has gained traction due to its compatibility with both node-level and graph-level tasks. Given a learned representation for a node or subgraph, the model assigns a class label corresponding to the closest prototype in embedding space \cite{liu2024one}. Formally, the predicted label for an instance $v$ with representation $\mathbf{h}_v$ is computed as:
\begin{equation}
    \hat{y}_v = \arg\min_{c} \, \text{dist}(\mathbf{h}_v, \mathbf{h}_c),
\end{equation}
where $\mathbf{h}_c$ denotes the prototype for class $c$ and $\text{dist}(\cdot, \cdot)$ is a distance metric, typically Euclidean or cosine distance. Prototype learning methods can be broadly categorized into two classes based on how the prototypes are generated: (i) prototypes derived from node representations, and (ii) prototypes learned via additional class nodes from external resources.

\noindent\textbf{Prototypes from Node Representations.}
A common and intuitive strategy is to compute class prototypes by averaging the representations of labeled nodes in the training set \cite{wang2024gft}. These prototypes serve as centroids that capture the semantic distribution of each class in the embedding space. Specifically, for class $c$, the prototype $\widetilde{\mathbf{h}}_c$ is calculated as: $\widetilde{\mathbf{h}}_c = \text{Mean} \left( \{ \mathbf{h}_v \, | \, y_v = c, \, v \in \mathcal{V}_l \} \right)$, where $\mathcal{V}_l$ denotes the set of labeled nodes and $\mathbf{h}_v$ is the representation of node $v$ obtained from a GFM encoder. This method is entirely parameter-free and leverages the inductive bias of neighborhood aggregation in GNNs. It has been adopted in recent works for few-shot or prompt-based graph learning~\cite{liu2023graphprompt, yu2024generalized, yu2025non, yu2024hgprompt}. Despite its simplicity, it often achieves strong performance, especially when the embeddings are well-separated in the latent space.

\noindent\textbf{Prototypes from Extra Class Nodes.}
An alternative approach models class prototypes explicitly as \emph{learnable nodes} within an auxiliary graph. This method constructs a bipartite graph $\mathcal{G}_g = (\mathcal{V}_g, \mathcal{E}_g)$, where the node set $\mathcal{V}_g$ is composed of data nodes $\mathcal{V}_d$ and class nodes $\mathcal{V}_c = \{c_1, c_2, \ldots, c_C\}$. Each data node in $\mathcal{V}_d$ corresponds to a labeled instance (e.g., a node or graph), while each class node represents a distinct class label. Edges are typically constructed between every data node and every class node to facilitate cross-node interaction. The class prototype $\hat{\mathbf{h}}_c$ is then defined as the learned embedding of class node $c$ after message passing on the graph. Such prototype construction has been employed in recent works \cite{huang2023prodigy,liu2024one,wang2024gft}, where LLMs are integrated into a graph prompt framework with class nodes.

\subsubsection{Discussion}

The diverse adaptation strategies for GFMs reflect a rich design space tailored to different downstream scenarios, data regimes, and computational constraints. Each method offers distinct benefits and trade-offs in terms of adaptability, scalability, and task performance. Transfer learning and distillation are well-suited for static environments with available training labels; test-time adaptation and prompting enable flexible inference in dynamic or low-resource settings; in-context learning excels in zero-shot generalization; and prototype learning provides interpretable, few-shot-friendly classification. Here is the detailed discussion.

\begin{itemize}[leftmargin=10pt]
    \item \textbf{Transfer Learning} is arguably the most conventional yet effective strategy for adapting GFMs. By initializing from a pre-trained model, it significantly reduces the need for large labeled datasets and accelerates convergence. Full fine-tuning provides maximum flexibility, enabling the model to specialize for task-specific patterns. However, it often requires extensive computational resources and risks overfitting on small target datasets. Adaptive fine-tuning and parameter-efficient fine-tuning mitigate these issues by limiting the number of trainable parameters, though they may underperform when task distributions deviate significantly from the pre-training regime.

    \item \textbf{Distillation} offers a compelling path for compressing and deploying GFMs in resource-constrained environments. By transferring knowledge from a large teacher model to a smaller student, it balances performance and efficiency. Moreover, distillation enables deployment in latency-sensitive applications such as edge devices or mobile platforms. The main limitation lies in the quality of the distilled knowledge—student models may fail to fully capture the nuanced relational patterns encoded by the teacher, especially when the structural complexity of graph data is high.

    \item \textbf{Test-Time Adaptation} addresses the challenge of domain shift without requiring access to source data or retraining. It enables models to dynamically adapt to new distributions, making it well-suited for continual and online learning scenarios. However, TTA is inherently limited by the lack of ground-truth supervision during inference and may suffer from unstable updates if the self-supervised signals are noisy or uninformative. Additionally, TTA assumes that test data arrives in a sequential or batch-wise fashion, which may not align with all practical settings.

    \item \textbf{Graph Prompting} is an efficient alternative to parameter tuning, leveraging learned prompt vectors to guide frozen GFMs toward specific tasks. Its main advantage lies in its modularity—prompts can be reused, swapped, or composed without modifying the base model. Data-level prompting and representation-level prompting both offer flexible mechanisms to influence model behavior. However, prompting often relies on careful prompt engineering or tuning, and performance may plateau if the model lacks sufficient task alignment. Moreover, prompts may not generalize well across tasks or domains without re-optimization.

    \item \textbf{In-Context Learning} enables zero-shot or few-shot adaptation without any gradient updates, making it ideal for quick deployment in low-resource or dynamic environments. When applied to text-attributed graphs, ICL allows LLMs to infer over graph-structured data by conditioning on task demonstrations. This approach eliminates the need for fine-tuning and benefits from the generative capacity of LLMs. However, the effectiveness of ICL hinges on the quality and relevance of demonstrations. Constructing demonstration sets for graphs is particularly challenging due to inter-instance dependencies, and poor demonstration selection can lead to significant performance degradation.

    \item \textbf{Prototype Learning} introduces a parameter-efficient and interpretable classification framework. It is particularly appealing for few-shot learning, where class prototypes derived from limited labeled data provide strong generalization. Methods based on averaging node representations are simple and effective, while graph-based class node construction captures richer semantics. Nonetheless, prototype learning assumes that class clusters are well-formed in the embedding space, which may not hold in noisy or heterophilic graphs. Furthermore, its reliance on distance metrics may overlook complex decision boundaries that could be better captured by discriminative classifiers.
\end{itemize}

\newpage

\section{Universal Graph Foundation Models}
\label{sec:universal gfm}

\subsection{Design Principle}

Graph foundation models are designed to operate across diverse domains and tasks, adapting to various graph structures and distributions. This vision parallels the role of LLMs in natural language processing and VLMs in computer vision. By leveraging insights from existing foundation models, we can formulate principles that guide the development of GFMs capable of addressing cross-domain and cross-task challenges. This section delineates the core characteristics and design principles essential for constructing such models.

\begin{itemize}[leftmargin=10pt]
    \item \textbf{Handling Heterogeneous Graph Distributions}: Graphs originating from various domains—such as molecular structures, social networks, and financial transaction systems—exhibit significant variability in size, connectivity, node attributes, density, and associated tasks. A GFM must generalize across such heterogeneous distributions with minimal domain-specific retraining. Analogous to LLMs, which leverage next-token prediction to extract semantic representations from extensive text corpora, GFMs require specifically designed self-supervised pretraining objectives to facilitate learning from large-scale, multi-domain graph datasets. The goal is to embed diverse graph structures into a unified latent space that enables meaningful cross-domain knowledge transfer.

    \item \textbf{Addressing Task Conflicts}: Graph-based tasks often involve competing objectives. For instance, node classification requires an understanding of local adjacency structures (e.g., homophily and heterophily), whereas graph classification focuses on higher-order motifs and structural patterns. LLMs reconcile such conflicts by framing all language-related tasks within a unified objective, question-answering. Similarly, GFMs must establish a shared inductive bias that harmonizes task representations across graphs. Multi-task learning techniques can further aid in balancing divergent learning objectives, ensuring that the model remains effective across a broad spectrum of graph-based tasks.

    \item \textbf{Facilitating Positive Transfer}: Despite domain shifts, certain high-level graph properties—such as topology patterns and homophily—exhibit consistency across diverse datasets. A well-designed GFM should learn a shared latent space that aligns graphs from different domains while preserving task-specific nuances. The challenge lies in mitigating discrepancies between tasks that possess distinct inductive biases while promoting knowledge transfer \cite{mao2024graph, liu2024one, zhao2024all,wang2024tackling}. Achieving this requires carefully balancing pretraining objectives with downstream adaptations, employing task-aware mechanisms to align model representations. Effective pretraining strategies should not only capture comprehensive graph semantics but also maintain adaptability, ensuring seamless alignment between pretraining and downstream tasks.
\end{itemize}

In the following sections, we examine existing universal GFMs through the lens of their approaches to addressing the three core challenges outlined above. Specifically, we categorize these methods across three fundamental levels: (1) \textit{model-level}, which focuses on model unification strategies to enhance transferability across tasks and domains; (2) \textit{pretrain-level}, which explores techniques for domain alignment during pretraining to enable cross-domain generalization; and (3) \textit{adaptation-level}, which investigates downstream task adaptation mechanisms that facilitate efficient fine-tuning and transfer learning.

\subsection{Graph Model-Based Universal GFM}

\begin{table*}[!t]
    \centering
    \rowcolors{2}{shadecolor}{white}
    \caption{Summary of GNN-based universal GFMs. }
    \label{tab:universal gfm gnn}
    \resizebox{\linewidth}{!}{
        \begin{tabular}{llllp{3.5cm}p{3.5cm}p{3.5cm}l}
            \toprule
            \textbf{Method Name}                              & \textbf{Backbone} & \textbf{Pretrain} & \textbf{Adaptation}             & \textbf{Feature Align}                    & \textbf{Structure Align}          & \textbf{Task Align}                 & \textbf{Github}                                                    \\
            \midrule
            \textbf{OpenGraph} \cite{xia2024opengraph}        & GNN               & Supervised        & Finetune                        & Data - SVD                                & N/A                               & Explicit - Subgraph                 & \href{https://github.com/HKUDS/OpenGraph}{Link}                    \\ 
            \textbf{OFA} \cite{liu2024one}                    & GNN               & Supervised        & Graph Prompting, Prototype      & Data - Text Attribute                     & N/A                               & Explicit - Subgraph                 & \href{https://github.com/LechengKong/OneForAll}{Link}              \\ 
            \textbf{MetaFP} \cite{jing2023deep}               & GNN               & Supervised        & Distillation                    & Model - Projection                        & N/A                               & N/A                                 & -                                                                  \\ 
            \textbf{SCORE} \cite{wang2024towards}             & GNN               & Supervised        & Test-time Adaptation            & Data - Text Attribute, Model - Projection & Data - Augment, Data - Synthesize & Explicit - Link                     & -                                                                  \\ 
            \textbf{STAGE} \cite{shen2024zeroshot}            & GNN               & Supervised        & Test-time Adaptation            & Data - Others                             & N/A                               & N/A                                 & -                                                                  \\ \midrule
            \textbf{HoloGNN} \cite{bevilacqua2025holographic} & GNN               & Contrastive       & Finetune                        & N/A                                       & N/A                               & Implicit - Regularizer              & -                                                                  \\ 
            \textbf{FoToM} \cite{davies2023its}               & GNN               & Contrastive       & Finetune                        & N/A                                       & Loss - Pretrain                   & N/A                                 & \href{https://pypi.org/project/fotom/}{Link}                       \\ 
            \textbf{EGI} \cite{zhu2021transfer}               & GNN               & Contrastive       & Finetune                        & N/A                                       & N/A                               & Explicit - Subgraph                 & \href{https://github.com/GentleZhu/EGI}{Link}                      \\ 
            \textbf{GIT} \cite{wang2024learning}              & GNN               & Contrastive       & Finetune, Prototype             & Data - Text Attribute                     & N/A                               & Explicit - Tree                     & \href{https://github.com/Zehong-Wang/GIT}{Link}                    \\ 
            \textbf{RiemannGFM} \cite{sun2025riemanngfm}      & GNN               & Contrastive       & Finetune                        & N/A                                       & Model - Tangent Space             & Implicit - Codebook                 & \href{https://github.com/RiemannGraph/RiemannGFM}{Link}            \\ 
            \textbf{BooG} \cite{cheng2024boosting}            & GNN               & Contrastive       & Finetune, Prototype             & Data - Text Attribute                     & Data - Augment                    & Explicit - Subgraph                 & \href{https://anonymous.4open.science/r/BooG-EE42/README.md}{Link} \\ 
            \textbf{UniGLM} \cite{fang2024uniglm}             & GNN               & Contrastive       & Finetune, Test-time Adaptation  & Data - Text Attribute                     & N/A                               & N/A                                 & -                                                                  \\ 
            \textbf{AnyGraph} \cite{xia2024anygraph}          & GNN               & Contrastive       & Finetune, Graph Prompting       & Data - SVD, Model - MoE                   & Data - Augment                    & Explicit - Subgraph                 & \href{https://github.com/HKUDS/AnyGraph}{Link}                     \\ 
            \textbf{GCC} \cite{qiu2020gcc}                    & GNN               & Contrastive       & Finetune                        & N/A                                       & N/A                               & Explicit - Subgraph                 & \href{https://github.com/THUDM/GCC}{Link}                          \\ 
            \textbf{GraphAlign} \cite{hou2024graphalign}      & GNN               & Contrastive       & Finetune                        & Data - Text Attribute                     & N/A                               & Implicit - MoE                      & \href{https://github.com/THUDM/GraphAlign}{Link}                   \\ 
            \textbf{HGPROMPT} \cite{yu2024hgprompt}           & GNN               & Contrastive       & Graph Prompting                 & Model - Projection                        & N/A                               & Explicit - Subgraph                 & \href{https://github.com/Starlien95/HGPrompt}{Link}                \\ 
            \textbf{OMOG} \cite{liu2024onemodel}              & GNN               & Contrastive       & Graph Prompting                 & Data - Text Attribute                     & Model - MoE                       & N/A                                 & -                                                                  \\ 
            \textbf{SAMGPT} \cite{yu2025samgpt}               & GNN               & Contrastive       & Graph Prompting                 & Model - Projection                        & Model - Prompt Learning           & Explicit - Subgraph                 & -                                                                  \\ 
            \textbf{Prodigy} \cite{huang2023prodigy}          & GNN               & Contrastive       & Graph Prompting                 & N/A                                       & Loss - Multi-task                 & Explicit - Subgraph                 & \href{https://github.com/snap-stanford/prodigy}{Link}              \\ \midrule
            \textbf{PGT} \cite{he2024generalizing}            & GNN               & Generative        & Finetune                        & N/A                                       & N/A                               & N/A                                 & -                                                                  \\ 
            \textbf{UniGraph} \cite{he2024unigraph}           & GNN               & Generative        & Finetune, In-context, Prototype & Data - Text Attribute                     & Loss - Pretrain                   & Explicit - Subgraph                 & \href{https://github.com/yf-he/UniGraph}{Link}                     \\ 
            \textbf{UniGraph 2} \cite{he2025unigraph2}        & GNN               & Generative        & Finetune                        & Data - Text Attribute                     & Loss - Pretrain                   & Explicit - Subgraph, Implicit - MoE & \href{https://github.com/yf-he/UniGraph2}{Link}                    \\ 
            \textbf{UniAug} \cite{tang2024cross}              & GNN               & Generative        & Test-time Adaptation            & Data - Node Property                      & Data - Augment                    & N/A                                 & \href{https://github.com/WenzhuoTang/UniAug}{Link}                 \\ \midrule
            \textbf{All in One} \cite{sun2023all}             & GNN               & Hybrid            & Finetune, Prototype             & Data - SVD                                & Model - Meta Learning             & Explicit - Subgraph                 & \href{https://github.com/sheldonresearch/ProG}{Link}               \\ 
            \textbf{PatchNet} \cite{sun2025handling}          & GNN               & Hybrid            & Finetune                        & Model - Projection                        & N/A                               & Implicit - Augment                  & \href{https://github.com/zjunet/PatchNet}{Link}                    \\ 
            \textbf{AUX-TS} \cite{han2021adaptive}            & GNN               & Hybrid            & Finetune                        & N/A                                       & N/A                               & Implicit - Aux. Loss                & \href{https://github.com/Sean-Huang65/AUX-TS}{Link}                \\ 
            \textbf{GFT} \cite{wang2024gft}                   & GNN               & Hybrid            & Finetune, Prototype             & Data - Text Attribute                     & Loss - Pretrain, Model - Codebook & Explicit - Tree                     & \href{https://github.com/Zehong-Wang/GFT}{Link}                    \\ 
            \textbf{ProNoG} \cite{yu2025non}                  & GNN               & Hybrid            & Graph Prompting                 & N/A                                       & Model - Codebook                  & Explicit - Subgraph                 & -                                                                  \\ 
            \textbf{MultiGPrompt} \cite{yu2023multigprompt}   & GNN               & Hybrid            & Graph Prompting                 & Model - Projection                        & Loss - Pretrain                   & Explicit - Subgraph                 & \href{https://github.com/Nashchou/MultiGPrompt}{Link}              \\ 
            \textbf{GraphPrompt} \cite{liu2023graphprompt}    & GNN               & Hybrid            & Graph Prompting                 & N/A                                       & N/A                               & Explicit - Subgraph                 & \href{https://github.com/Starlien95/GraphPrompt}{Link}             \\ 
            \textbf{RAGraph} \cite{jiang2024ragraph}          & GNN               & Hybrid            & Graph Prompting, In-context     & N/A                                       & Model - Codebook                  & N/A                                 & \href{https://github.com/Artessay/RAGraph/}{Link}                  \\ 
            \textbf{IGAP} \cite{yan2024inductive}             & GNN               & Hybrid            & Graph Prompting                 & Model - Projection                        & Reg                               & Explicit - Link                     & -                                                                  \\
            \bottomrule
        \end{tabular}
    }
\end{table*}

Graph model-based universal GFMs primarily follow a \textit{pretrain-then-adapt} paradigm. Given a large-scale, diverse graph database spanning multiple domains and tasks, the goal is to pretrain a graph encoder that captures transferable structural and semantic patterns, which can then be adapted to unseen graphs for downstream tasks. We summarize existing works based on the backbones, pretraining methods, adaptation strategies, as well as the techniques used in handling feature, structure, and task heterogeneity in Table \ref{tab:universal gfm gnn}.

\subsubsection{Model Unification}

A core challenge in building universal GFMs lies in designing GNN encoders capable of generalizing across tasks, domains, and graph topologies. Existing approaches toward model unification can be broadly categorized into two paradigms: (1) \textit{explicit unification} through task and input reformulation, and (2) \textit{implicit unification} via architectural generalization and invariance enforcement.



\noindent\textbf{Explicit Unification.}
Explicit unification approaches seek to transform diverse graph-based tasks into a unified prediction format, facilitating a shared pretraining and inference pipeline. Based on task granularity, we identify three major categories: link-level, subgraph-level, and tree-level unification.

\begin{itemize}[leftmargin=10pt]
    \item \textbf{Link-Level Unification.}
          One strategy frames all graph-based tasks as link prediction problems by introducing \textit{class nodes} into the graph and predicting links between these nodes and task-relevant nodes. Unlike traditional architectures that append a classifier atop node embeddings, these models cast classification as a link prediction task:
          \begin{equation}
              \hat{y}_{v_i} = \sigma \left( \text{sim}(\mathbf{h}_{v_i}, \mathbf{h}_{c_k}) \right), \quad \forall v_i \in \mathcal{V},\; c_k \in \mathcal{V}_c,
          \end{equation}
          where $\mathbf{h}_{v_i}$ and $\mathbf{h}_{c_k}$ denote the embeddings of the target node and class node, respectively, and $\sigma$ is a scoring function such as softmax or sigmoid. GPPT~\cite{sun2022gppt} pioneered this approach by combining masked edge prediction and prompting techniques. Subsequent methods~\cite{yan2024inductive,yu2025non} extend this framework with prompt tokens for enhanced task alignment. While elegant and general, this formulation may overlook fine-grained substructures essential for certain prediction tasks.

    \item \textbf{Subgraph-Level Unification.}
          To incorporate local structural context, a second category formulates graph tasks at the \textit{subgraph level}~\cite{qiu2020gcc,liu2023graphprompt,sun2023all,huang2023prodigy,liu2024one,he2024unigraph}. This two-stage process first extracts ego-graphs centered around target nodes and then applies a GNN encoder to each subgraph. For instance, in node classification, an ego-graph is constructed around each node, where the label of the induced subgraph corresponds to the label of the central node. The same principle extends to edge- and graph-level tasks, thereby unifying diverse graph tasks at the subgraph level. This is formulated as:
          \begin{equation}
              \hat{y}_{v_i} = f_{\text{Classifier}} \left( f_{\text{GNN}}(\mathcal{G}_{v_i}) \right), \quad \mathcal{G}_{v_i} = \operatorname{Ego}(\mathcal{G}, v_i, r),
          \end{equation}
          where $\mathcal{G}_{v_i}$ is the ego-subgraph around node $v_i$ within radius $r$. The resulting subgraph embedding encoded via GNN $f_{\text{GNN}}$ is passed to a classifier $f_{\text{Classifier}}$, which can be a linear head~\cite{qiu2020gcc} or a class-node-based scoring function~\cite{liu2023graphprompt,sun2023all}.

          Early works, such as GCC \cite{qiu2020gcc}, adopt contrastive pretraining to capture structural patterns across multi-domain graphs. This paradigm is further extended by GraphPrompt \cite{liu2023graphprompt}, Prodigy \cite{huang2023prodigy}, and All in One \cite{sun2023all}, which introduce prompt learning techniques to enhance alignment between pretraining and downstream tasks. OFA \cite{liu2024one} and UniGraph \cite{he2024unigraph} generalize this approach to cross-domain scenarios. This paradigm supports a unified framework for node, edge, and graph-level tasks and has shown strong empirical performance in domain adaptation and transfer learning. However, subgraph-level unification suffers from two major limitations: (1) subgraph extraction introduces substantial computational overhead, increasing both time and memory costs, and (2) message-passing GNNs may struggle to capture essential substructures, leading to suboptimal representation learning \cite{garg2020generalization,chen2020can,morris2023wl,zhang2024beyond}.

    \item \textbf{Tree-Level Unification.}
          A more recent and efficient alternative introduces \textit{virtual nodes} that connect to task-relevant nodes, eliminating the need for subgraph extraction. These virtual nodes serve as surrogates for tree-structured representations, with their embeddings used for downstream predictions. For example, in node classification, virtual nodes are linked to all original nodes, and their embeddings, learned via message passing, serve as the final representations:
          \begin{equation}
              \hat{y} = f_{\text{Classifier}} \left( f_{\text{GNN}}(\mathcal{G}^+)_{\text{[v]}} \right), \quad \mathcal{G}^+ = \mathcal{G} \cup \{v\} \cup \{(v, v_i) \mid v_i \in \mathcal{V}_{\text{task}} \},
          \end{equation}
          where $\mathcal{G}^+$ augments the original graph with a virtual node $v$ connected to all task-relevant nodes, $f_{\text{GNN}}(\mathcal{G}^+)_{\text{[v]}}$ indicates the embedding of the virtual nodes, and $\mathcal{V}_{\text{task}}$ is the set of task-relevant nodes (e.g., all nodes for node classification). The GNN encoder computes the embedding of $v$, which reflects aggregated information from the graph. This design significantly improves efficiency while preserving representation capacity. GFT \cite{wang2024gft} pioneered this approach, demonstrating both empirical and theoretical evidence that tree similarity correlates with improved task transferability. GIT \cite{wang2024learning} further formalized the stability, transferability, and generalization of tree-based representations from a theoretical perspective.
\end{itemize}

\noindent\textbf{Implicit Unification.}
While explicit task reformulation provides a principled way to unify graph tasks, an alternative line of research focuses on \textit{implicit architectural unification}—modifying the underlying model design to enable generalization across diverse tasks and domains without altering the task definitions themselves. One representative example is HoloGNN~\cite{bevilacqua2025holographic}, which addresses the limitation that conventional GNN architectures are often hardwired for specific task types, such as node-level or link-level classification. These models may fail to generalize across tasks that involve different permutation symmetries or structural granularity. To overcome this, HoloGNN introduces \textit{expansion} and \textit{reduction} maps that explicitly model node-permutation symmetries. By decomposing the input graph into permutation-invariant components and then reconstructing task-specific views, HoloGNN enables a single architecture to adapt flexibly across varied learning tasks. SCORE~\cite{wang2024towards} proposes \textit{relation graph} framework: Rather than operating directly on input graphs, SCORE constructs a semantic interaction graph that encodes cross-domain entity relationships. By integrating \textit{semantic-conditioned message passing}, the model dynamically adapts to domain-specific patterns while preserving shared structural and semantic invariants. AnyGraph~\cite{xia2024anygraph} focuses on enhancing model expressiveness through a mixture-of-experts architecture combined with high-order structural injection. By introducing high-order connectivity patterns and gating mechanisms, AnyGraph captures both local and non-local interactions while maintaining modularity and scalability. OpenGraph~\cite{xia2024opengraph} tackles graph heterogeneity from a data representation perspective. It introduces a \textit{topology-aware tokenizer} that converts variable-sized graph structures (e.g., adjacency matrices) into fixed-length sequences suitable for Transformer-based encoders. This tokenizer preserves key topological properties while allowing foundation models to operate uniformly across graphs of different sizes and shapes. Together, these implicit unification strategies demonstrate that architecture design, independent of task reformulation, plays a pivotal role in building transferable and robust GFMs. As graph data continues to grow in complexity and diversity, the interplay between explicit and implicit unification mechanisms will be essential for the development of scalable, general-purpose graph models.

\subsubsection{Domain Alignment in Pretraining}

\noindent\textbf{Feature Alignment.} Ensuring feature consistency across diverse graph structures is fundamental to designing GNN-based GFMs, as a single GNN inherently struggles to generalize across heterogeneous graph features \cite{wang2025can}. Existing approaches can be broadly categorized into two paradigms: (1) textual and multimodal feature alignment and (2) model-based feature alignment.

\begin{itemize}[leftmargin=10pt]
    \item \textbf{Textual and Multimodal Feature Alignment.}
          To unify heterogeneous graph signals, recent works project textual or multimodal attributes into a shared representation space. Given a graph $\mathcal{G} = (\mathbf{X}, \mathbf{A}, \mathbf{D})$, where $\mathbf{D} = \{ \mathbf{d}_i \}_{i=1}^N$ denotes the set of node-level textual or multimodal descriptions, a pretrained encoder $f_{\text{enc}}$ (e.g., SentenceBERT \cite{reimers2019sentence}, CLIP \cite{radford2021learning}) is used to map each description to a latent embedding:
          \begin{equation}
              \mathbf{h}_i = f_{\text{enc}}(\mathbf{d}_i), \quad \forall v_i \in \mathcal{V},
          \end{equation}
          where $\mathbf{h}_i \in \mathbb{R}^d$ represents the aligned feature vector of node $v_i$ in the shared semantic space. In this framework, $\mathbf{d}_i$ may correspond to natural language descriptions, visual content, or a combination thereof, depending on the graph modality.

          Models such as OFA~\cite{liu2024one} utilize fixed templates and SentenceBERT to encode textual descriptions, while UniGraph~\cite{he2024unigraph} fine-tunes the encoder $f_{\text{enc}}$ during pretraining to improve alignment across graph domains. UniGLM~\cite{fang2024uniglm} further adopts contrastive pretraining to train $f_{\text{enc}}$ for stronger discriminative alignment. To extend beyond single-modal representations, UniGraph2~\cite{he2025unigraph2} employs CLIP as a unified encoder for both text and image modalities, effectively aligning multimodal node features into a coherent embedding space. In more advanced designs such as GraphAlign~\cite{hou2024graphalign}, a mixture-of-experts model dynamically selects among several encoders using a learnable routing mechanism.

          Despite their effectiveness, one practical bottleneck of multimodal feature alignment lies in the limited availability of rich textual or image-labeled graphs. To address this, TANS~\cite{wang2025can} proposes synthetic graph augmentation by leveraging LLMs to automatically generate textual descriptions from raw graph structures, enabling broader adoption of alignment-based GFMs even in non-textural settings.

    \item \textbf{Model-Based Feature Alignment.}
          Beyond preprocessing with textual or multimodal encoders, a parallel line of research focuses on \textit{architectural innovations} that directly align heterogeneous node features through model design. These approaches aim to internally reconcile feature discrepancies across graphs from different domains or modalities, without relying solely on external encoders. One common strategy involves introducing a domain-specific projection function $f_{\text{proj}}$ that transforms raw node features $\mathbf{x}_i$ into a unified latent space $\mathbf{h}_i = f_{\text{proj}}(\mathbf{x}_i), \forall v_i \in \mathcal{V}$. For instance, DARE~\cite{jing2023deep} adopts \textit{model reprogramming}, where $f_{\text{proj}}$ consists of lightweight input and output adapters appended to a frozen pretrained GNN. This enables effective adaptation across downstream tasks without modifying core model parameters. Such reprogramming reduces training costs and enhances parameter reuse across domains.

          Another technique employs \textit{singular value decomposition (SVD)} to orthogonalize feature spaces before further alignment. Let $\mathbf{X} = \mathbf{U} \boldsymbol{\Sigma} \mathbf{V}^\top$ be the SVD decomposition of node features. The transformed features $\mathbf{U} \boldsymbol{\Sigma}$ serve as normalized input embeddings:
          \begin{equation}
              \mathbf{h}_i = f_{\text{align}}(\mathbf{U} \boldsymbol{\Sigma}_i),
          \end{equation}
          where $f_{\text{align}}$ can incorporate learnable tokens~\cite{yu2025samgpt,yan2024inductive}, LLM-based augmentation modules~\cite{xia2024opengraph}, or mixture-of-expert routers~\cite{xia2024anygraph} to further adapt feature semantics.

          PatchNet~\cite{sun2025handling} proposes a more compositional approach by constructing \textit{graph patches}—small, learnable semantic units representing transferable substructures. Let $\mathcal{P}_i = \{ \mathcal{G}_{i}^{(1)}, \dots, \mathcal{G}_{i}^{(K)} \}$ denote the set of $K$ patches extracted for node $v_i$. Each patch is encoded independently and then aggregated to form the final representation $\mathbf{h}_i = \text{Aggregate} \left( \left\{ f_{\text{patch}}(\mathcal{G}_{i}^{(k)}) \right\}_{k=1}^K \right)$, where $f_{\text{patch}}$ is a GNN-based encoder over patches, and \texttt{Aggregate} denotes a pooling or attention mechanism.

          Despite their effectiveness in aligning heterogeneous node features, model-based alignment approaches exhibit several notable limitations. A primary drawback lies in their limited generalizability to unseen graphs. Many of these methods—such as domain-specific projection layers, SVD-based transformations, or learnable tokens—are tightly coupled to the feature distribution and structural patterns of the pretraining graphs. As a result, when deployed on graphs from entirely new domains or with previously unseen feature spaces, the alignment mechanisms may fail to preserve semantic consistency or yield meaningful representations.
\end{itemize}

\noindent\textbf{Structure Alignment.}
Graphs originating from different domains often exhibit fundamentally distinct structural patterns. For instance, social networks are characterized by high clustering coefficients and frequent triangular motifs, while molecular graphs are dominated by small cycles and functional substructures. Such structural heterogeneity presents a significant challenge for building graph foundation models that generalize across domains. To address this, a range of strategies have been proposed to align structural information during pretraining and inference.

\begin{itemize}[leftmargin=10pt]
    \item \textbf{Domain-Invariant Pretraining Objectives.}
          One class of methods seeks to learn domain-invariant structural representations by optimizing pretraining objectives that generalize across diverse graph topologies. FoToM~\cite{davies2023its} adopts adversarial contrastive learning to disentangle domain-specific features while preserving task-relevant structure. By learning representations that are indistinguishable across domains from the perspective of an adversary, the model acquires a form of structure-aware invariance.

    \item \textbf{Structural Vocabulary Construction.}
          Another approach involves defining a shared vocabulary \cite{yang2023vqgraph} of structural motifs that capture recurring patterns across graphs. GFT~\cite{wang2024gft} constructs a tree-based codebook during pretraining that encodes canonical structural features. This codebook remains fixed during inference, providing a consistent basis for interpreting new graphs. Similarly, RiemannGFM~\cite{sun2025riemanngfm} introduces a geometric extension, modeling both tree and cycle motifs in Riemannian space to better capture curved and hierarchical graph structures. These vocabulary-based approaches promote structural alignment by grounding graph representations in discrete, reusable structural primitives.

    \item \textbf{Graph Prompting for Structural Adaptation.}
          Graph prompt learning has also emerged as an effective mechanism for structural alignment. IGAP~\cite{yan2024inductive} analyzes spectral discrepancies between graphs and uses learnable prompt tokens to align representations in the spectral domain. Empirical results suggest that graph signals are more transferable in the low-frequency space, which can be amplified through spectral prompt injection. BooG~\cite{cheng2024boosting} introduces virtual nodes during pretraining to harmonize structural contexts across graph domains. ProNoG~\cite{yu2025non} extends this further by employing a control network that generates node-specific prompts adaptively, enabling fine-grained structural calibration without manual intervention.

    \item \textbf{Mixture-of-Experts for Structural Diversity.}
          An alternative perspective on structure alignment emphasizes architectural diversity. OMOG~\cite{liu2024onemodel} proposes a mixture-of-experts (MoE) framework, motivated by the observation that a single GNN often fails to capture the inductive biases necessary for structurally diverse graphs. OMOG pretrains multiple specialized GNNs, each tuned to a distinct structural domain, and stores them in a model bank. At inference time, a gating function dynamically selects the most relevant expert based on the similarity between the input and pretrained graphs. This expert routing mechanism reduces negative transfer and provides modular flexibility in adapting to novel structures.
\end{itemize}

\noindent\textbf{Multi-Objective Learning.} To effectively balance the inductive biases inherent in different downstream tasks, it is essential to employ diverse pretraining objectives tailored to distinct learning paradigms. A prevalent strategy involves leveraging multi-task learning frameworks, enabling models to jointly optimize multiple objectives \cite{wang2024gft,zhang2022chasing, he2024unigraph,he2024generalizing,tian2024ugmae,ju2023multi,liu2023fair}.

For instance, to comprehensively capture knowledge embedded in graphs, GFT \cite{wang2024gft} integrates node-level, link-level, and semantic-level reconstruction objectives during pretraining, allowing the model to extract structural and semantic information from multiple perspectives. Similarly, UniGraph \cite{he2024unigraph} co-trains a GNN and an LLM in a unified framework, employing contrastive learning on the GNN side while leveraging masked token prediction on the LLM side. This synergistic approach enhances the model's capability to learn both graph topology and rich semantic representations. PGF \cite{he2024generalizing} simultaneously optimizes feature reconstruction and local graph structure reconstruction under a graph transformer framework, demonstrating its effectiveness in complex industry-scale applications such as game data modeling.

To achieve a better trade-off among competing objectives, recent works have explored advanced optimization techniques, including Pareto optimization, learnable task tokens, and meta-learning. Specifically, MultiGPrompt \cite{yu2023multigprompt} introduces multiple learnable pretext tokens to bridge the gap between diverse task objectives, such as local-global contrastive learning, global-global contrastive learning, and link prediction. ParetoGNN \cite{ju2023multi} employs Pareto optimization to balance five distinct pretraining tasks, achieving an optimal trade-off across objectives. Meanwhile, All in One \cite{sun2023all} applies meta-learning to optimize multi-task prompt initialization, improving generalization and adaptability across various tasks.

\subsubsection{Downstream Task Adaptation}

\noindent\textbf{Fine-Tuning.}
Fine-tuning the pretrained model or employing linear probing remains the most widely adopted approach for adapting GFMs to downstream tasks, a paradigm that has been extensively utilized in graph self-supervised learning \cite{qiu2020gcc}. Traditional graph self-supervised learning methods typically follow a two-step procedure: (1) pretraining the model on a source graph, and (2) appending a linear classifier for downstream classification, either through full fine-tuning or linear probing, where only the final layer is updated. Extending this approach, GFMs aim to generalize to unseen graphs in an inductive setting—a key objective for achieving universal graph learning. However, positive transfer across domains remains challenging. To bridge this gap, researchers have proposed various strategies to facilitate rapid adaptation of pretrained models to downstream tasks.

\noindent\textbf{Transfer Learning.}
A primary strategy for downstream adaptation is transfer learning, wherein a pretrained model is fine-tuned on new graphs to improve task-specific performance. While effective, this approach is computationally expensive, motivating the development of faster transfer learning techniques. EGI \cite{zhu2021transfer} enhances transferability by capturing essential graph structures through ego-graph distribution modeling, linking transferability to local graph Laplacians of the source and target domains. Additionally, AUX-TS \cite{han2021adaptive} identifies discrepancies between the optimization objectives of self-supervised pretraining and downstream tasks, introducing auxiliary tasks with adaptive weighting to bridge this gap and improve adaptation efficiency.

\noindent\textbf{Prompt Learning.}
Inspired by prompt learning in LLMs, researchers have explored graph prompt learning as an efficient alternative to traditional fine-tuning. Instead of updating the pretrained model, this approach fine-tunes learnable graph prompt tokens, facilitating rapid adaptation to downstream tasks. GPPT \cite{sun2022gppt} pioneers this direction by introducing three key components: (1) \textit{prompt addition}, which reformulates input nodes as token pairs; (2) \textit{prompt answering}, where pretrained models estimate linking probabilities between tokens; and (3) \textit{prompt tuning}, which optimizes pretext task losses with orthogonal initialization and regularization, enabling adaptation without modifying the backbone model. Subsequent works have further refined this paradigm by (1) unifying pretraining and downstream tasks at the subgraph level \cite{liu2023graphprompt,huang2023prodigy,sun2023all,liu2024one}, (2) leveraging multi-scale graph prompt tokens \cite{yan2024inductive,yu2025samgpt}, (3) designing advanced prompt templates \cite{sun2023all, huang2023prodigy, liu2024one, yu2024hgprompt}, and (4) incorporating external knowledge sources \cite{jiang2024ragraph,wang2024towards}. These advancements collectively enhance the efficiency and generalizability of graph prompt learning.

\noindent\textbf{Prototype Learning.}
While graph prompt learning minimizes the number of tunable parameters, it still requires additional tuning efforts. Prototype learning offers an alternative approach by constructing class-specific prototype embeddings and using them for downstream task predictions, a technique widely employed in few-shot learning. Given $k$ classes, each with $n$ samples, class prototypes are formed by averaging the available samples. A new instance is then assigned to the closest prototype. Several works adopt this paradigm to facilitate efficient adaptation. OFA \cite{liu2024one}, UniGraph \cite{he2024unigraph}, and UniGraph2 \cite{he2025unigraph2} introduce a query-support framework, where support graphs are used to construct class prototypes. This framework can be further extended to zero-shot learning by leveraging external resources, such as textual class embeddings \cite{liu2024one}. GFT \cite{wang2024gft} follows a similar paradigm but incorporates limited fine-tuning on a minimal set of examples, demonstrating that even a small number of labeled samples can significantly enhance downstream performance.

\noindent\textbf{Structural Augmentation.}
Beyond model-centric adaptation approaches, structural augmentation has emerged as a complementary strategy for improving downstream task performance. UniAug \cite{tang2024cross} introduces a universal structural augmentor based on a discrete diffusion model, pretrained exclusively on graph structures from over 1,000 datasets. This augmentor can enhance the structures of downstream graphs through guided generation, benefiting node-, link-, and graph-level tasks. Empirical results suggest that increasing the volume of pretraining data improves downstream generalization, as it enriches the model's understanding of diverse structural patterns across graphs.

\subsection{Language Model-Based Universal GFM}

\begin{table*}[!t]
    \centering
    \rowcolors{2}{shadecolor}{white}
    \caption{Summary of LLM-based universal GFMs. }
    \label{tab:universal gfm llm}
    \resizebox{\linewidth}{!}{
        \begin{tabular}{llllp{3.5cm}p{3.5cm}p{3.5cm}l}
            \toprule
            \textbf{Method Name}                                & \textbf{Backbone} & \textbf{Pretrain} & \textbf{Adaptation}  & \textbf{Feature Align} & \textbf{Structure Align} & \textbf{Task Align} & \textbf{Github}                                             \\
            \midrule
            \textbf{LangGFM} \cite{lin2024langgfm}                 & LLM               & Supervised        & Finetune, In-context & Data - Text Attribute  & Loss - Pretrain          & Explicit - QA       & -                                                           \\
            \textbf{Meta-Transformer} \cite{zhang2023meta}      & LLM               & Generative        & Finetune             & N/A                    & N/A                      & N/A                 & \href{https://github.com/invictus717/MetaTransformer}{Link} \\ 
            \textbf{QueryRAG} \cite{li2025large}                & LLM               & Generative        & In-context           & Data - Text Attribute  & Data - Augment           & Explicit - QA       & -                                                           \\ 
            \textbf{GraphAgent} \cite{yu2024graphagent}         & LLM               & Generative        & In-context           & Data - Text Attribute  & Data - Augment           & Explicit - QA       & -                                                           \\ 
            \textbf{Graph-ToolFormer} \cite{zhang2023graph}     & LLM               & Generative        & In-context           & Data - Text Attribute  & Data - Augment           & Explicit - QA       & \href{https://github.com/jwzhanggy/Graph_Toolformer}{Link}  \\ 
            \textbf{InstructGraph} \cite{wang2024instructgraph} & LLM               & Generative        & Finetune, In-context & Data - Text Attribute  & Data - Augment           & Explicit - QA       & \href{https://github.com/wjn1996/InstructGraph}{Link}       \\ 
            \textbf{Beyond Text} \cite{hu2023beyond}            & LLM               & Generative        & In-context           & Data - Text Attribute  & Data - Augment           & Explicit - QA       & -                                                           \\ 
            \textbf{Graph Agent} \cite{wang2023graph}           & LLM               & Generative        & In-context           & Data - Text Attribute  & Data - Augment           & Explicit - QA       & -                                                           \\ 
            \textbf{InstructGLM} \cite{ye2024language}          & LLM               & Generative        & Finetune, In-context & Data - Text Attribute  & Data - Augment           & Explicit - QA       & \href{https://github.com/agiresearch/InstructGLM}{Link}     \\ 
            \textbf{GraphICL} \cite{sun2025graphicl}            & LLM               & Generative        & In-context           & Data - Text Attribute  & Data - Augment           & Explicit - QA       & -                                                           \\
            \bottomrule
        \end{tabular}

    }
\end{table*}

This section explores approaches that leverage large language models as predictors for graph-based tasks. LLM-based GFMs have design objective centered around three key components: (1) developing an effective tokenization strategy to transform graph data into text-based representations, (2) post-training LLMs to incorporate graph-specific knowledge, and (3) designing advanced adaptation techniques to enhance alignment with downstream tasks. We summarize the LLM-based universal GFMs in Table \ref{tab:universal gfm llm}. This formulation consists of three core components:

\begin{enumerate}[leftmargin=15pt]
    \item \textbf{Graph Tokenization:} Define a tokenization function $\mathcal{T}$ that maps graph $\mathcal{G}$ into a text sequence:
          \begin{equation}
              \mathbf{s} = \mathcal{T}(\mathcal{G}),
          \end{equation}
          where $\mathbf{s} = [s_1, s_2, \dots, s_L]$ is a sequence of $L$ tokens representing the nodes, edges, and attributes of the input graph.

    \item \textbf{Post-Training on Graph Data:} Fine-tune a pretrained LLM with graph-specific objectives (e.g., masked token prediction, topology autoencoding) using the tokenized input $\mathbf{s}$:
          \begin{equation}
              \theta^* = \arg\min_{\theta} \mathcal{L}_{\text{graph}}(\operatorname{LLM}(\mathbf{s}; \theta), y),
          \end{equation}
          where $\mathcal{L}_{\text{graph}}$ denotes the graph-aware training loss and $y$ is the supervision signal.

    \item \textbf{Downstream Adaptation:} For downstream task inference, apply an adaptation function $\mathcal{A}$ (e.g., in-context learning, prompting, instruction tuning) to structure the input prompt and decode predictions:
          \begin{equation}
              \hat{y} = \mathcal{A}(\operatorname{LLM}(\mathbf{s})),
          \end{equation}
          where $\hat{y}$ is the model prediction adapted to the downstream task format.
\end{enumerate}

\subsubsection{Model Unification}

Large language models are inherently designed as task-agnostic architectures capable of solving a wide range of text-based problems within a unified framework. To extend their capabilities to graph-structured data, a critical challenge lies in developing effective conversion schemes that translate graphs into representations amenable to language modeling. Specifically, the goal is to align the inductive biases of graph structures with the sequential and semantic processing capabilities of LLMs. Existing approaches to this problem fall broadly into two categories: (1) \textit{natural language conversion}, which reformulates graph structures as textual narratives, and (2) \textit{structured format conversion}, which encodes graphs into organized, code-like representations such as JSON or nested lists.

\noindent\textbf{Natural Language Conversion.} Natural language conversion methods~\cite{ye2024language,lin2024langgfm,sun2025graphicl,hu2023beyond,yu2024graphagent,li2025large} represent graph elements (nodes, edges, and attributes) using human-readable descriptions. These techniques leverage the inherent linguistic capabilities of LLMs to perform reasoning over graphs by narratively expressing relational patterns and structural dependencies. For example, in citation graphs, nodes representing academic papers can be described using titles, abstracts, and citation relationships, thereby embedding graph semantics into natural language text~\cite{hu2023beyond}. While sharing the core objective of converting graphs into text, existing methods vary significantly in terms of their prompt design, context modeling, and reasoning strategies:

\begin{itemize}[leftmargin=10pt]
    \item \textbf{Template Construction:} Most approaches begin with handcrafted or automated templates that standardize the textual description of graph elements. These templates capture key graph components—such as node labels, edge types, degrees, or graph-level statistics—and embed them in coherent textual forms. QueryRAG~\cite{li2025large}, for instance, incorporates detailed structural summaries including adjacency properties and graph size indicators.

    \item \textbf{Hierarchical Context Inclusion:} To improve graph comprehension, several methods incorporate multi-hop neighborhood information~\cite{ye2024language,sun2025graphicl}. This allows LLMs to access not only direct relationships but also broader structural context. Such descriptions are often layered hierarchically to reflect structural depth.

    \item \textbf{Task-Specific Reasoning and Prompting:} Some models further introduce dynamic reasoning mechanisms, using multi-step prompts, memory updates, and agent-style decision-making~\cite{yu2024graphagent,sun2025graphicl}. These guided prompts help the model iteratively refine its understanding and decision process, improving performance on complex graph tasks such as traversal, path-finding, or subgraph reasoning.
\end{itemize}

\noindent\textbf{Structured Format Conversion.}
Structured conversion approaches~\cite{wang2023graph,zhang2023graph,wang2024instructgraph} provide an alternative to free-form text by encoding graphs into well-defined data structures such as JSON, code blocks, or nested lists. This design preserves graph topology while enabling LLMs to process the information in a structured and interpretable manner. The core idea is to treat graphs as serialized data that LLMs can parse, analyze, and manipulate. These methods typically exhibit the following key characteristics:

\begin{itemize}[leftmargin=10pt]
    \item \textbf{Predefined Templates and Tokenization:} Structured methods use domain-specific templates to define graph elements. For instance, nodes may be represented as objects with fields for attributes and neighbors, while edges are encoded as relations within nested arrays~\cite{zhang2023graph}. Automated tools (e.g., ChatGPT) can populate these templates during preprocessing.

    \item \textbf{API-Augmented Inference:} Several approaches extend the inference pipeline by integrating APIs that enable dynamic graph traversal or data retrieval during the reasoning process. This augmentation facilitates tasks such as knowledge graph completion or personalized recommendation~\cite{zhang2023graph}.

    \item \textbf{Hierarchical Context Encoding:} Similar to natural language methods, structured formats may also include hierarchical representations of node neighborhoods. GraphAgent~\cite{wang2023graph} demonstrates this by incorporating recursive context trees, improving the LLM's ability to model nested relational dependencies.
\end{itemize}

Both paradigms—natural language and structured conversion—offer viable paths to unify graph reasoning under LLM frameworks. Natural language prompts prioritize interpretability and ease of human understanding, while structured formats provide clarity and consistency, particularly for complex or large-scale graph tasks. The choice between these methods often reflects trade-offs in generalization, computational efficiency, and compatibility with downstream applications.

\subsubsection{Domain Alignment in Model Training}

Once graph templates have been defined and translated into structured textual representations, LLMs can naturally interpret and reason over graph data using their native text-processing capabilities. A straightforward approach involves directly leveraging pretrained LLMs as zero-shot or few-shot predictors. In this setting, graph-formatted prompts are fed into the model, enabling \textit{in-context learning} without any additional parameter updates~\cite{sun2025graphicl,hu2023beyond,wang2023graph,yu2024graphagent,li2025large}. This paradigm capitalizes on the strong generalization and compositional reasoning abilities of LLMs, allowing them to perform tasks through purely prompt-based supervision.

While zero-shot performance is often competitive, empirical evidence suggests that post-training adaptation—particularly techniques such as supervised fine-tuning (SFT) and preference alignment (e.g., PPO, DPO)—can substantially improve alignment between LLMs and graph-structured tasks. Accordingly, recent research has explored various strategies to better adapt LLMs to graph reasoning through domain-specific training objectives and multi-task instruction tuning.

\noindent\textbf{Instruction-Tuned Fine-Tuning.} InstructGraph~\cite{wang2024instructgraph} introduces a supervised fine-tuning approach that incorporates structured graph reasoning tasks into the LLM training process. By pairing graph-encoded inputs with curated instructions and target outputs, the model learns to follow graph-specific reasoning patterns. Preference alignment techniques such as reinforcement learning with human feedback (RLHF) or direct preference optimization (DPO) are further employed to enhance output faithfulness and task adherence. InstructGLM~\cite{ye2024language} generalizes this concept through multi-prompt training, where a diverse set of graph tasks—including classification, generation, and summarization—are cast into instruction-based prompts. The LLM is trained across this multi-task corpus to promote generalization and cross-task transfer. This unified training paradigm enables LLMs to handle graph problems with varying input-output formats and structural complexities.

\noindent\textbf{Self-Supervised Graph Alignment.} LangGFM~\cite{lin2024langgfm} proposes an alternative fine-tuning strategy grounded in self-supervised learning (SSL). Inspired by traditional graph pretraining techniques, LangGFM introduces two novel SSL objectives tailored for LLM-based graph understanding: \textit{Topology Autoencoding} and \textit{Feature Masked Autoencoding}. The topology autoencoder encourages the model to reconstruct structural information (e.g., edge connections, adjacency statistics) from textual graph descriptions, while the feature masked autoencoder masks node attributes and predicts them from context—analogous to masked language modeling, but applied to graph-encoded text. These self-supervised objectives promote alignment between LLM representations and graph topologies without requiring manually labeled data. As a result, LangGFM achieves strong performance on downstream tasks while maintaining label efficiency and robustness to domain shifts.

\subsubsection{Downstream Task Adaptation}

\noindent\textbf{Zero-Shot Reasoning.} Large language models exhibit strong zero-shot generalization capabilities, allowing them to perform a wide range of tasks without explicit fine-tuning. This property extends naturally to LLM-based GFMs, enabling them to solve graph-related tasks directly from textual representations of graph data. By designing appropriate prompts that align graph structures with LLMs' pretraining distributions, researchers have demonstrated that LLMs can perform node classification, link prediction, and reasoning tasks without updating any model parameters~\cite{ye2024language,hu2023beyond,wang2023graph,wang2024instructgraph,zhang2023meta}. The effectiveness of zero-shot adaptation hinges on high-quality template engineering, where graphs are serialized into descriptive formats that LLMs can interpret semantically.

\noindent\textbf{In-Context Learning.} Beyond zero-shot reasoning, LLMs can leverage in-context learning to improve task performance by utilizing demonstrations within input prompts. This approach involves prepending labeled graph instances or structured explanations before queries, injecting task-specific knowledge without requiring model parameter updates. Several works explore this paradigm by employing basic in-context learning strategies using labeled graph instances, effectively guiding LLMs in downstream tasks \cite{hu2023beyond,wang2023graph,wang2024instructgraph}.

To enhance in-context learning, researchers have developed advanced prompting techniques that incorporate structured representations of graph data. GraphICL \cite{sun2025graphicl} introduces a structured prompting framework that enables general-purpose LLMs to outperform specialized graph models in resource-constrained and out-of-domain tasks. The designed prompts consist of four key components: (1) \textit{task descriptions} to define objectives, (2) \textit{anchor node text} to contextualize relevant entities, (3) \textit{structure-aware information} to provide topological insights, and (4) \textit{labeled demonstrations} to facilitate few-shot learning setting. These elements collectively enable adaptation to node classification and link prediction tasks. GraphAgent \cite{yu2024graphagent} reformulates downstream graph tasks as an agent planning problem, where LLMs take structured actions based on graph topology and node content. This framework incorporates a hierarchical memory mechanism that integrates both long-term and short-term memory modules, enabling efficient handling of large-scale graph data. By storing high-quality examples dynamically, GraphAgent significantly improves LLMs' ability to generalize across graph domains. QueryRAG \cite{li2025large} enhances in-context learning by explicitly incorporating query nodes, contextual information from graph neighbors, and corresponding labels into structured prompts. This design ensures that graph structure serves as an inherent contextual feature, allowing LLMs to better capture relational dependencies and improve reasoning over graph data.

\subsection{Graph-Language Co-Training Universal GFM}

\begin{table*}[!t]
    \centering
    \rowcolors{2}{shadecolor}{white}
    \caption{Summary of graph-language co-training universal GFMs. }
    \label{tab:universal gfm gnn llm}
    \resizebox{\linewidth}{!}{
        \begin{tabular}{llllp{3.5cm}p{3.5cm}p{3.5cm}l}
            \toprule
            \textbf{Method Name}                                     & \textbf{Backbone} & \textbf{Pretrain} & \textbf{Adaptation}  & \textbf{Feature Align} & \textbf{Structure Align}          & \textbf{Task Align}                & \textbf{Github}                                            \\
            \midrule
            \textbf{GOFA} \cite{kong2025gofa}                        & GNN + LLM         & Supervised        & Finetune, In-context & Data - Text Attribute  & Data - Augment                    & Explicit - QA, Explicit - Subgraph & \href{https://github.com/JiaruiFeng/GOFA}{Link}            \\
            \textbf{GraphGPT} \cite{tang2024graphgpt}                & GNN + LLM         & Supervised        & Finetune, In-context & Data - Text Attribute  & Data - Augment                    & Explicit - QA                      & \href{https://github.com/HKUDS/GraphGPT}{Link}             \\
            \textbf{GALLM} \cite{luo2024enhance}                     & GNN + LLM         & Supervised        & Finetune, In-context & Data - Text Attribute  & Data - Augment                    & Explicit - QA                      & -                                                          \\
            \textbf{LLaGA} \cite{chenllaga}                          & GNN + LLM         & Supervised        & In-context           & Data - Text Attribute  & Data - Augment                    & Explicit - QA                      & \href{https://github.com/VITA-Group/LLaGA}{Link}           \\ \midrule
            \textbf{GraphCLIP} \cite{zhu2024graphclip}               & GNN + LLM         & Contrastive       & Finetune             & Data - Text Attribute  & Loss - Pretrain                   & Explicit - QA                      & \href{https://github.com/ZhuYun97/GraphCLIP}{Link}         \\
            \textbf{GraphTranslator} \cite{zhang2024graphtranslator} & GNN + LLM         & Generative        & Finetune, In-context & Data - Text Attribute  & Data - Augment                    & Explicit - QA                      & \href{https://github.com/alibaba/GraphTranslator}{Link}    \\
            \textbf{PromptGFM} \cite{zhu2024llm}                     & GNN + LLM         & Generative        & Finetune, In-context & Data - Text Attribute  & Data - Augment, Model - Codebook  & Explicit - QA                      & \href{https://github.com/agiresearch/PromptGFM}{Link}      \\
            \textbf{GraphPrompter} \cite{liu2024can}                 & GNN + LLM         & Generative        & Finetune             & Data - Text Attribute  & Data - Augment                    & Explicit - QA                      & \href{https://github.com/franciscoliu/graphprompter}{Link} \\
            \textbf{TEA-GLM} \cite{wang2025llms}                     & GNN + LLM         & Generative        & Finetune, In-context & Data - Text Attribute  & Data - Augment                    & Explicit - QA                      & \href{https://github.com/W-rudder/TEA-GLM}{Link}           \\
            \textbf{NT-LLM} \cite{ji2024nt}                          & GNN + LLM         & Generative        & Finetune             & Data - Text Attribute  & Data - Augment                    & Explicit - QA                      & -                                                          \\
            \textbf{AskGNN} \cite{hu2024lets}                        & GNN + LLM         & Generative        & Finetune, In-context & Data - Text Attribute  & Data - Augment, Model - Retriever & Explicit - QA                      & -                                                          \\
            \textbf{GraphAgent} \cite{yu2024graphagent}              & GNN + LLM         & Generative        & Finetune, In-context & Data - Text Attribute  & Data - Augment                    & Explicit - QA                      & \href{https://github.com/HKUDS/GraphAgent}{Link}           \\
            \bottomrule
        \end{tabular}
    }
\end{table*}

Large language models have demonstrated remarkable capabilities in processing unstructured data such as natural language, exhibiting strong generalization in zero-shot and few-shot settings. Their ability to perform compositional reasoning, follow instructions, and adapt to diverse downstream tasks has made them a cornerstone of modern AGI. However, LLMs are inherently limited in their capacity to process structured data—such as graphs—which encode high-order dependencies, long-range interactions, and complex relational structures that are not naturally captured by sequential token representations. Although one strategy involves tokenizing graph-structured data into textual sequences to leverage LLMs directly, this transformation often introduces substantial inductive bias mismatch. Critical structural information—such as neighborhood topologies, connectivity patterns, and graph invariants—is frequently lost or obfuscated during linearization, leading to suboptimal reasoning performance. This challenge motivates the need for hybrid architectures that preserve graph inductive biases while enabling semantic generalization through LLMs.

Drawing inspiration from VLMs, which integrate visual embeddings into LLMs to support multimodal reasoning, recent efforts have explored hybrid Graph-Language Models that combine the representational power of GNNs with the reasoning capabilities of LLMs. These approaches aim to unify structural and semantic signals by bridging the modality gap between graphs and language. The development of such models typically involves three key steps: (1) training a GNN to capture structural information, (2) projecting graph embeddings into the token space of an LLM, and (3) leveraging LLMs for downstream inference tasks. Formally, this formulation consists of the following core components:

\begin{enumerate}[leftmargin=15pt]
    \item \textbf{Graph Encoding:} A GNN encoder is trained to extract node- or graph-level embeddings that capture both local and global structural dependencies:
          \begin{equation}
              \mathbf{H} = \operatorname{GNN}(\mathbf{X}, \mathbf{A}),
          \end{equation}
          where $\mathbf{X}$ and $\mathbf{A}$ are the node features and adjacency matrix, respectively, and $\mathbf{H} = \{\mathbf{h}_1, \dots, \mathbf{h}_N\}$ denotes the learned node embeddings.

    \item \textbf{Cross-Modality Projection:} The graph embeddings are mapped into the token space of the LLM via a projection function $\rho$, aligning structural information with the language model's input representation:
          \begin{equation}
              \mathbf{Z} = \rho(\mathbf{H}), \quad \mathbf{Z} \in \mathbb{R}^{N \times d},
          \end{equation}
          where $\mathbf{Z}$ serves as the token-aligned graph representation and $d$ is the dimensionality of the LLM token space.

    \item \textbf{Language-Based Inference:} The LLM $f_{\text{LLM}}$ consumes a prompt $\mathcal{P}$ that incorporates the projected graph embeddings $\mathbf{Z}$ alongside optional task instructions and demonstration tokens:
          \begin{equation}
              \hat{y} = f_{\text{LLM}}([\mathcal{P} \ \| \ \mathbf{Z}]),
          \end{equation}
          where $\|$ denotes the concatenation operation and $\hat{y}$ is the model prediction for the downstream task.
\end{enumerate}

\subsubsection{Model Unification}

Recent approaches integrating GNNs with LLMs generally adopt a framework akin to visual-language models, wherein a graph encoder extracts structural features, and an LLM performs reasoning over the extracted representations. Various techniques have been proposed to bridge the gap between graph embeddings and token-based language models, improving alignment, interpretability, and reasoning over graph-structured data.

Several methods focus on encoding graph structures into tokenized representations that align with LLM processing. LLaGA \cite{chenllaga} restructures nodes into sequential representations using two distinct templates: \textit{Neighborhood Detail}, which captures local connectivity, and \textit{Hop-Field Overview}, which encodes broader structural context. A learned projection module aligns these structured representations with LLM token embeddings. GraphAgent \cite{yang2024graphagent} introduces \textit{Graph-Token Grounding}, mapping nodes and edges into structured Python objects, while \textit{Graph Tokenization} embeds discrete graph entities for generative and predictive tasks. The model employs an intent recognition module to dynamically adjust system prompts based on user queries. GraphTranslator \cite{zhang2024graphtranslator} facilitates interaction between GNNs and LLMs through a translation module that converts graph embeddings into textual representations, with a producer module generating alignment data to ensure consistency between graph structures and natural language descriptions. NT-LLM \cite{ji2024nt} employs \textit{Node Tokenization}, selecting anchor nodes to efficiently encode graph structures while preserving topological integrity. This method enhances relational representation and reasoning capabilities within LLMs. TEA-GLM \cite{wang2025llms} refines prompt construction to generalize across diverse graph tasks, structuring prompts into three components: (1) graph information, (2) task descriptions, and (3) multiple-choice answers for cross-dataset reasoning. A dedicated projection module maps graph tokens into instruction-tuned LLMs, ensuring alignment across different tasks.

\subsubsection{Domain Alignment in Model Training}

Once unified graph-text representations are obtained, domain alignment during model training plays a crucial role in enhancing reasoning capabilities and generalization across diverse graph tasks.

\noindent\textbf{Supervised Alignment.}
Supervised fine-tuning remains the most prevalent strategy for aligning GNN and LLM components. Typically, graph encoders are appended to LLMs, and the combined architecture is trained end-to-end on labeled datasets~\cite{chenllaga,luo2024enhance,ji2024nt,liu2024can,zhang2024graphtranslator}. This allows the model to learn tight coupling between structural features and textual reasoning, adapting to specific downstream applications.

\noindent\textbf{Self-Supervised Alignment.}
To reduce reliance on labeled data, several methods adopt self-supervised learning paradigms. GALLM~\cite{luo2024enhance} introduces a text matching objective using both manual and learnable \textit{soft category prompts}, optimized via backpropagation. The model follows a two-stage pipeline consisting of SSL-based pretraining followed by supervised fine-tuning. GOFA~\cite{kong2025gofa} formulates graph understanding as sentence completion and reasoning, unifying diverse tasks such as structural analysis and question answering under a shared prompt space. LangGFM~\cite{lin2024langgfm} adapts traditional GNN pretraining ideas into LLM-friendly formats, proposing \textit{Topology Autoencoding} and \textit{Feature Masked Autoencoding} to enhance structural reasoning via textualized inputs.

\noindent\textbf{CLIP-like Alignment.}
Inspired by CLIP~\cite{radford2021learning}, GraphCLIP~\cite{zhu2024graphclip} applies contrastive learning to align graph representations with textual summaries. Large-scale graph-summary pairs are generated with the help of LLMs and used to train a contrastive encoder-decoder system. This pretraining strategy facilitates zero-shot and few-shot transfer, enabling robust generalization across domains and graph types.

\subsubsection{Downstream Task Adaptation}

\noindent\textbf{Zero-Shot Reasoning.}
LLMs inherently support zero-shot reasoning by leveraging their pretrained knowledge and prompt-based adaptation. When coupled with graph representations, this capability enables effective inference on graph tasks such as node classification and link prediction without any additional fine-tuning~\cite{chenllaga,kong2025gofa,yang2024graphagent,liu2024can}. Zero-shot adaptation is particularly valuable in low-resource settings, where labeled graph data is scarce or unavailable.

\noindent\textbf{In-Context Learning (ICL).}
Several methods leverage ICL to improve graph reasoning by embedding task demonstrations directly into prompts. AskGNN~\cite{hu2024lets} enhances classification accuracy through \textit{ICL Example Purification}, a two-stage method that selects informative support examples via LLM-based scoring and filters class-imbalanced samples. This improves both representativeness and balance of in-context examples. RAG-based methods such as RAGraph~\cite{jiang2024ragraph} retrieve graph-relevant contexts during inference, dynamically enriching prompts and improving generalization to previously unseen graph structures.

\noindent\textbf{Interpretability.}
One of the major advantages of using LLMs for graph reasoning is their ability to produce human-readable explanations. GraphTranslator~\cite{zhang2024graphtranslator} showcases this through multi-turn dialog reasoning over user activity graphs, supporting interpretability across behavioral analytics, social network dynamics, and recommendation scenarios. By combining natural language outputs with structured representations, such models enhance transparency and trustworthiness in graph-based AI systems.

\subsection{Discussion}

Universal GFMs aim to generalize across graph domains and tasks by leveraging large-scale pretraining and adaptable architectures. Each type exhibits unique strengths and limitations, reflecting inherent trade-offs in model expressiveness, scalability, generalization, and interpretability. In summary, GNN-based models are efficient and structure-aware but semantically limited; LLM-based models are flexible and language-driven but struggle with graph topology; hybrid models aim to unify the best of both worlds but introduce significant system complexity.

\noindent\textbf{Graph Model-Based GFMs.} 
GNN-based universal models are structurally aligned with graph data and excel at capturing topological information through message passing and local neighborhood aggregation. These models are highly effective for tasks where structure is critical. \textbf{Advantages:} (1) Strong inductive biases tailored to graphs, supporting generalization across graph domains. (2) Computationally efficient for large-scale graphs due to localized computations. (3) Well-suited for structural tasks involving graph motifs. \textbf{Limitations:} (1) Limited ability to encode rich semantic information (e.g., textual or multimodal node attributes). (2) Difficulty scaling to out-of-domain tasks without significant architecture or prompt redesign. (2) Struggles with long-range dependencies due to local message passing constraints.

\noindent\textbf{Language Model-Based GFMs.} 
LLM-based approaches \cite{zhao2023graphtext} leverage pretrained language models to reason over graph-structured data via textual or structured prompts. These methods translate graphs into sequences, enabling zero-shot or in-context reasoning across heterogeneous tasks. \textbf{Advantages:} (1) Exceptional generalization across tasks and domains via instruction following and prompt-based adaptation. (2) Flexibility to support multimodal data (e.g., text and images). (3) High interpretability due to natural language outputs and reasoning chains. \textbf{Limitations:} (1) Loss of structural fidelity when graph topology is linearized into token sequences. (2) Heavy computational cost due to large model sizes and input serialization overhead. (3) Lack of built-in graph inductive biases, requiring extensive template engineering for structural tasks.

\noindent\textbf{Graph-Language Co-Training GFMs.} 
Hybrid models integrate GNNs and LLMs to unify the structural reasoning of GNNs with the semantic and general-purpose reasoning capabilities of LLMs. Examples include LLaGA~\cite{chenllaga}, GraphTranslator~\cite{zhang2024graphtranslator}, and GraphCLIP~\cite{zhu2024graphclip}. These models project graph embeddings into the LLM token space, enabling multimodal and cross-domain inference. \textbf{Advantages:} (1) Combines structural awareness and semantic richness, improving task transferability and expressiveness. (2) Supports both structured and unstructured inputs, making it ideal for real-world applications (e.g., recommendation, knowledge graphs). (3) Enables flexible adaptation through prompting, fine-tuning, or contrastive alignment techniques. \textbf{Limitations:} (1) Complex model architecture with high computational and memory demands. (2) Requires careful alignment between graph embeddings and token representations. (3) Integration of supervision signals across modalities remains a challenging research problem.

\newpage

\section{Task-Specific Graph Foundation Models}
\label{sec:task gfm}

\subsection{Design Principle}

Task-specific GFMs are designed to operate across multiple domains while focusing on solving a single task (e.g., node classification, link prediction, graph generation). Although graphs from different domains may share inductive biases relevant to a particular downstream task, their structural properties can vary significantly. Therefore, an effective task-specific GFM must not only capture task-aware invariances across domains but also align disparate graph distributions to ensure robust generalization. This section outlines the core characteristics and design principles essential for developing task-specific GFMs.

\begin{itemize}[leftmargin=10pt]
    \item \noindent\textbf{Task-Specific Inductive Bias.} Different graph tasks impose distinct inductive biases on the learning process. For instance, node classification often relies on homophily and heterophily principles, link prediction emphasizes local and global connectivity patterns, while graph classification focuses on recognizing meaningful substructures. Despite structural variations across graphs, task-related representations tend to exhibit shared patterns. By leveraging this observation, a well-designed GFM can learn domain-invariant representations while preserving essential task-specific knowledge, ensuring adaptability across diverse datasets.

    \item \noindent\textbf{Cross-Domain Alignment.} Graphs from different domains differ in their structural properties, node types, feature distributions, and connectivity patterns. To ensure generalization to unseen graph distributions with minimal retraining, the model must learn representations that remain robust across domains. Achieving this requires techniques such as domain-adaptive architectures, where GNN layers dynamically adjust based on input domain characteristics, or domain-specific feature modulation through attention mechanisms and parameter-efficient adaptation. Such strategies help mitigate performance degradation when transitioning between domains.

    \item \noindent\textbf{Balancing Domain Generalization and Task-Specific Adaptation.} A fundamental challenge in task-specific GFMs is balancing domain invariance with task-specific adaptation. Overemphasizing domain understanding may introduce inductive biases that favor specific domains at the expense of task-specific generalizations, whereas fully generalized models risk overlooking critical domain-specific patterns, thereby reducing performance. To strike this balance, domain generalization techniques such as domain regularization, normalization, and adversarial training can be employed. These approaches help prevent domain overfitting and catastrophic forgetting, enabling models to maintain both cross-domain adaptability and high task-specific performance.

\end{itemize}

In the following subsections, we systematically explore the design principles of GFMs across six fundamental graph-related tasks: node classification, link prediction, graph classification, question answering, anomaly detection, and recommendation. For each task, we examine the underlying design philosophies, key challenges, state-of-the-art methodologies, and promising directions for future research.

\subsection{Node-level Task}

\begin{table*}[!t]
    \centering
    \rowcolors{2}{shadecolor}{white}
    \caption{Summary of task-specific GFMs on node-level tasks. }
    \label{tab:task gfm node}
    \resizebox{\linewidth}{!}{
        \begin{tabular}{llllp{3.5cm}p{3.5cm}p{3.5cm}p{3.5cm}l}
            \toprule
            \textbf{Method Name}                                          & \textbf{Task} & \textbf{Backbone} & \textbf{Pretrain} & \textbf{Adaptation}            & \textbf{Feature Align} & \textbf{Structure Align}                & \textbf{Task Align}    & \textbf{Github}                                            \\
            \midrule
            \textbf{ENGINE} \cite{zhu2024efficient}                       & Node-Level          & GNN               & Supervised        & Finetune                       & Data - Text Attribute  & Loss - Pretrain                         & N/A                    & \href{https://github.com/ZhuYun97/ENGINE}{Link}            \\ 
            \textbf{LLM-GNN} \cite{chen2023label}                         & Node-Level          & GNN               & Supervised        & Finetune                       & N/A                    & Data - Augment                          & N/A                    & \href{https://github.com/CurryTang/LLMGNN}{Link}           \\ 
            \textbf{GCNMF} \cite{taguchi2021graph}                        & Node-Level          & GNN               & Supervised        & Finetune                       & Data - Others          & N/A                                     & N/A                    & -                                                          \\ 
            \textbf{PCFI} \cite{um2023confidence}                         & Node-Level          & GNN               & Supervised        & Finetune                       & Data - Others          & N/A                                     & N/A                    & \href{https://github.com/daehoum1/pcfi}{Link}              \\ 
            \textbf{GRAFENNE} \cite{gupta2023grafenne}                    & Node-Level          & GNN               & Supervised        & Finetune                       & Data - Others          & N/A                                     & N/A                    & -                                                          \\ 
            \textbf{GraphAny} \cite{zhao2025fully}                     & Node-Level          & GNN               & Supervised        & N/A                       & Model - Projection     & N/A                                     & Implicit - Regularizer & \href{https://github.com/DeepGraphLearning/GraphAny}{Link} \\ 
            \textbf{E-LLaGNN} \cite{jaiswal2024all}                       & Node-Level          & GNN               & Supervised        & Finetune, Test-time Adaptation & Data - Text Attribute  & Model - Retriever                       & N/A                    & -                                                          \\ 
            \textbf{GraphFM} \cite{lachi_graphfm_2024}                           & Node-Level          & GNN               & Supervised        & Finetune                       & Model - Projection     & Model - Structure Learning              & N/A                    & -                                                          \\ 
            \textbf{TPP} \cite{niu2024replay}                             & Node-Level          & GNN               & Supervised        & Test-time Adaptation           & N/A                    & Model - Prompt Learning                 & N/A                    & \href{https://github.com/mala-lab/TPP}{Link}               \\ 
            \textbf{SimMLP} \cite{wang2024training}                       & Node-Level          & GNN               & Contrastive       & Finetune                       & N/A                    & Loss - Pretrain                         & N/A                    & \href{https://github.com/Zehong-Wang/SimMLP}{Link}         \\ 
            \textbf{FUG}                                                  & Node-Level          & GNN               & Contrastive       & Finetune                       & Data - SVD             & N/A                                     & N/A                    & \href{https://github.com/hedongxiao-tju/FUG}{Link}         \\ 
            \textbf{GraphControl} \cite{zhu2024graphcontrol}              & Node-Level          & GNN               & Contrastive       & Finetune                       & Model - Projection     & Model - Prompt Learning                 & N/A                    & \href{https://github.com/wykk00/GraphControl}{Link}        \\ 
            \textbf{ZeroG} \cite{li2024zerog}                             & Node-Level          & GNN               & Contrastive       & Finetune, Prototype            & Data - Text Attribute  & Loss - Pretrain                         & Explicit - Subgraph    & \href{https://github.com/NineAbyss/ZeroG}{Link}            \\ 
            \textbf{GraphLoRA} \cite{yang2024graphlora}                   & Node-Level          & GNN               & Contrastive       & Finetune                       & Model - Projection     & Model - MoE, Model - Structure Learning & Implicit - Regularizer & \href{https://github.com/AllminerLab/GraphLoRA}{Link}      \\ 
            \textbf{MDGFM} \cite{wang2025multi}                           & Node-Level          & GNN               & Contrastive       & Graph Prompting                & Model - Projection     & Model - Prompt Learning                 & Explicit - Link        & -                                                          \\ 
            \textbf{GPT-GNN} \cite{hu2020gpt}                             & Node-Level          & GNN               & Generative        & Finetune                       & N/A                    & Loss - Pretrain                         & N/A                    & \href{https://github.com/acbull/GPT-GNN}{Link}             \\ 
            \textbf{GSPT} \cite{song2024pure}                             & Node-Level          & GNN               & Generative        & Finetune, Prototype            & Data - Text Attribute  & Loss - Multi-task                       & N/A                    & -                                                          \\ 
            \textbf{GPPT} \cite{sun2022gppt}                              & Node-Level          & GNN               & Generative        & Graph Prompting                & N/A                    & Loss - Pretrain                         & Explicit - Link        & \href{https://github.com/MingChen-Sun/GPPT}{Link}          \\ 
            \textbf{GCOPE} \cite{zhao2024all}                             & Node-Level          & GNN               & Hybrid            & Graph Prompting                & Model - Projection     & Loss - Pretrain                         & N/A                    & \href{https://github.com/cshhzhao/GCOPE}{Link}             \\ \midrule
            \textbf{GDL4LLM} \cite{zhou2025each}                          & Node-Level          & LLM               & Generative        & Finetune                       & Data - Node Property   & N/A                                     & N/A                    & -                                                          \\ 
            \textbf{GraphText} \cite{zhao2023graphtext}                   & Node-Level          & LLM               & Generative        & Test-time Adaptation           & Data - Text Attribute  & Data - Augment                          & Explicit - QA          & \href{https://github.com/AndyJZhao/GraphText}{Link}        \\ \midrule
            \textbf{LOGIN} \cite{qiao2024login}                           & Node-Level          & GNN + LLM         & Supervised        & Finetune                       & Data - Text Attribute  & Data - Augment                          & Explicit - QA          & \href{https://github.com/QiaoYRan/LOGIN}{Link}             \\ 
            \textbf{LangGSL} \cite{su2024bridging}                        & Node-Level          & GNN + LLM         & Supervised        & Finetune                       & Data - Text Attribute  & Model - Structure Learning              & N/A                    & -                                                          \\ 
            \textbf{Cella} \cite{zhang2024cost}                           & Node-Level          & GNN + LLM         & Supervised        & Test-time Adaptation           & Data - Text Attribute  & Model - Structure Learning              & N/A                    & -                                                          \\ 
            \textbf{G2P2} \cite{wen2023augmenting}                        & Node-Level          & GNN + LLM         & Contrastive       & Graph Prompting                & Data - Text Attribute  & Loss - Pretrain                         & N/A                    & -                                                          \\ 
            \textbf{LangTopo} \cite{guan_langtopo_2024}                   & Node-Level          & GNN + LLM         & Generative        & Finetune                       & Data - Text Attribute  & Data - Augment                          & N/A                    & -                                                          \\ 
            \textbf{Dr.E} \cite{liu2024multiviewempoweredstructuralgraph} & Node-Level          & GNN + LLM         & Generative        & Finetune                       & Data - Text Attribute  & Model - Codebook                        & Explicit - QA          & \href{https://github.com/Timothy914/Dr.E}{Link}            \\
            \bottomrule
        \end{tabular}
    }
\end{table*}

Node-level tasks focus on predicting the properties or roles of nodes within a graph. Node classification is one of the fundamental node-level downstream tasks for graph learning that involves prediction on individual nodes within graphs, which gains significant attention in GFMs~\cite{xiao2022graph, mao2024position}. The node classification task is one of the most common node-level downstream tasks, which focuses on inferring the node labels~\cite{wang2024gft, liu2024one, pei2020geom, ma2025adaptive}. Formally, given a graph $\gG = \{\gV, \gE\}$ with node features $\gX$, the node classification task aims to learn a function $f:\gV \rightarrow \gY$ that maps each node to label $\hat{y_i} \in\gY$ to minimize the discrepancy between $\hat{y_i}$ and the ground truth $y_i$. Conventional GNNs employ message-passing frameworks that aggregate feature information from neighboring nodes to the center node, and further utilize a classifier, such as MLP, to predict node labels based on learned representations~\cite{qian2022co}. However, GFMs for node-level tasks still face the following challenges: (i) graph heterogeneity, (ii) cross-domain alignment, and (iii) balancing between domain generalization and task-specific adaptation. In this section, we will discuss existing methods for dealing with these challenges.

\subsubsection{Handling Graph Heterogeneity}

\textbf{Heterogeneous Feature Space.}
To handle the heterogeneous feature space challenge, existing works ~\cite{wang2023heterogeneous, qian2022co, ma2025llm} widely adopt MLPs to map features into a shared latent space, while these approaches may become ineffective in general cases, such as missing attribute features, dynamic attribute feature space, etc.
In light of this, earlier works~\cite{taguchi2021graph, um2023confidence} focus on missing feature imputation problems in graph representation learning, where several node features are absent or the distribution is different from others.
GCNmf~\cite{taguchi2021graph} utilizes the Gaussian Mixture Model (GMM)~\cite{smieja2018processing} to model missing features.
Particularly, it computes the expected activation of neurons in the first layer of GNN to handle the missing feature issue and learning graph representation simultaneously.
Another work, PCPI~\cite{um2023confidence} advances GCNmf in more challenging scenarios, i.e., high rates of missing features, by introducing pseudo-confidence, a channel-wise shortest path distance between missing feature nodes and the nearest known feature nodes.
In detail, PCPI proposes a feature imputation scheme that performs channel-wise inter-node diffusion to recover the missing feature and node-wise inter-channel propagation to refine the node features.
Besides the missing feature imputation problems, several studies~\cite{gupta2023grafenne, zhao2025fully, zhao2024fug} have explored the dynamic feature sets issue in graphs.
For example, GRAFENNE~\cite{gupta2023grafenne} implements an allotropic transformation on graphs, decoupling nodes, and attribute features via bipartite encoding.
GraphAny~\cite{zhao2025fully} computes fully-inductive features based on interactions between graph kernels and achieves inductive generalization across heterogeneous feature spaces. Additionally, GRAFENNE introduces a bipartite message-passing framework for these allotropically transformed graphs, allowing the model parameters size to remain independent of feature dimensions.
This approach alleviates the heterogeneous and diverse feature dimensions issues in graphs, making the model adaptable to unseen nodes and features.
Moreover, FUG~\cite{zhao2024fug} introduces a feature-universal contrastive pre-training strategy that avoids the need for model rebuilding and data reshaping to handle the feature heterogeneity issue.
Specifically, it designs an encoder with contrastive constraints to emulate the Principle Component Analysis~\cite{abdi2010principal} generation of the basis transformation matrix, which is utilized to adapt features in various spaces.

\noindent\textbf{Structure Heterogeneity.} Besides the heterogeneous feature space issue, structure heterogeneity is another challenge in building GFMs for downstream node classification tasks~\cite{wang2022survey, ma2025adaptive}.
GPT-GNN~\cite{hu2020gpt}, one of the earlier works, proposes a generative pre-train GNNs strategy, which can be factorized into node attribute and edge generation steps, to capture the inherent dependency between node attributes and graph structure over unlabeled graph data.
Building on the pre-training concept, GPPT~\cite{sun2022gppt} introduces a prompt mechanism in the fine-tuning state, which modifies the input nodes to token pairs that are directly applied to pre-trained GNNs.
In considering the deployment of GNNs in latency-sensitive applications, simMLP~\cite{wang2024training} introduces a simple yet efficient self-supervised framework that learns Multiple-Layer Perceptrons (MLPs) on graphs that align the representations encoded by graph context-aware GNNs and neighborhood dependency-free MLPs, thereby fully integrating the structural information into MLPs.

\subsubsection{Cross-domain Alignment}
\noindent\textbf{GNN-based Models.} The aforementioned methods primarily pre-train and fine-tune GNNs over an identical graph dataset, which is trivial to adapt cross-domain scenarios where graphs with various node features and structure types.
To handle this, one line of approaches~\cite{hu2020gpt, sun2022gppt} employs pure GNNs for GFMs development.
GCOPE~\cite{zhao2024all} designs an "All in One and One for All" framework that trains a single model on diverse datasets to handle versatile downstream tasks across a variety of domains.
Meanwhile, TPP~\cite{niu2024replay} introduces Graph Class-Incremental Learning (GCIL), which employs the Laplacian smoothing approach to generate task-specific prototypical embeddings for node classification tasks in various graphs.
TPP theoretically analyzes the task-specific prototypes of the same graph task are nearly the same with a large smoothing step, while those of different tasks are distinct due to differences in graph structure and node attributes.
Moreover, TPP adopts a graph prompting approach for GCIL that learns a small discriminative graph prompt for each task, essentially resulting in a separate classification mode for each graph task, thereby ensuring the trained GCIL model is both reply-free and forget-free.
GraphControl~\cite{zhu2024graphcontrol} leverages universal structural pre-trained models to align the input space across various graphs and incorporate unique characteristics of target data as conditional inputs.
Afterward, these conditions are integrated into the model during fine-tuning or prompt turning through ControlNet~\cite{zhang2023adding}, facilitating various downstream node classifications.
GraphLoRA~\cite{yang2024graphlora} is another "pre-train, fine-tune" framework that utilizes Structure-aware Maximum Mean Discrepancy~\cite{borgwardt2006integrating} to align divergent node feature distributions across source and target graphs.
Besides the pre-trained GNN, GraphLoRA injects another small GNN in the fine-tuning stage to effectively bridge structural distribution gaps while mitigating catastrophic forgetting.
Similarly, MDGFM~\cite{wang2025multi} applies token operators for semantic alignment between graphs, and further refines each source domain using graph structure learning (GSL), integrating both feature and topology information in the pre-train stage.
Moreover, GraphAny~\cite{zhao2025fully} attempts a full-inductive setup with respect to new structures, features, and label spaces, which includes two components: LinearGNNs and an inductive attention module.
The LinearGNNs enable efficient inductive inference on unseen graphs, while the inductive attention module learns to adaptively aggregate predictions from multiple LinearGNNs.

\noindent\textbf{Graph Transformers.} Another line of approaches directly employs transformers for node representation learning in various domains.
GSPT~\cite{song2024pure} is a feature-centric pre-training framework for text-attributed graphs (TAGs) that leverages a standard Transformer~\cite{vaswani2017attention} to learn a unified model for node representations.
GraphFM~\cite{lachi_graphfm_2024} extends GSPT to multi-domain scenarios via Perceiver-based encoder~\cite{jaegle2107perceiver} to compress domain-specific features into a common latent space.
GDL4LLM~\cite{zhou2025each} treats graphs as a new language, which translates graphs into a graph language corpus and pre-trains transformer-based LLMs on this corpus to enable understanding of graph structures.
The LLM is further fine-tuned as the next token prediction task for downstream node classification tasks.

\noindent\textbf{Hybrid Methods.}
Taking advantage of the exceptional natural language understanding capabilities of LLMs~\cite{wu2025comprehensive, ma2025llm}, several works~\cite{wen2023augmenting} also concentrate on hybrid methods that leverage both GNNs and LLMs to develop GFMs
Hybrid methods often explore Text-Attributed Graphs (TAGs) datasets~\cite{ma2023hypergraph, qian2025adaptive} and can be further divided into two groups based on the types of the classifier mechanisms for downstream tasks, i.e., GNNs or LLMs.
G2P2~\cite{wen2023augmenting} utilizes GNN as a predictor for text classification that jointly pre-trains text and graph encoders via three graph interaction-based contrastive strategies: (i) text-node interactions, (ii) text-summary interactions, and (iii) node-summary interactions.
Another work, ENGINE~\cite{zhu2024efficient} provides a parameter and memory-efficient fine-tuning method for textual graphs that incorporates GNN with LLM via side structure.
Moreover, LLM-GNN~\cite{chen2023label} introduces a label-free node classification pipeline, i.e., "LLMs-as-annotator", which merely trains GNNs over a small ratio of annotated nodes by LLMs for node classification tasks on remaining nodes.
Following the "LLMs-as-annotator" concept, several approaches~\cite{qiao2024login, zhang2024cost} have been introduced to advance LLM-GNN to some extent.
LOGIN~\cite{qiao2024login} develops an "LLMs-as-Consultants" paradigm that merely consults LLMs over low-confident prediction nodes and augments the original graphs based on the LLM feedback.
Similarly, Cella~\cite{zhang2024cost} employs an active node selection process to sift out representative nodes based on label inharmonicity and entropy.
Cella then annotates these representative nodes via LLMs and applies a Dirichlet Energy-based graph rewiring strategy~\cite{zhou2021dirichlet} to minimize the adverse effects of noisy or missing links in the original graphs.
In addition to annotation, some studies leverage LLMs to enhance the node representation in GFMs.
EhLLaGNN~\cite{jaiswal2024all} samples high-quality neighborhoods through LLMs, followed by on-demand neighborhood feature enhancement using diverse prompts from its prompt catalog.
These enhanced neighborhood features are further aggregated with the central node to generate representative node embeddings for downstream node-level tasks.

\noindent\textbf{Hybrid Approaches.}
Another group of works directly uses LLMs as classifiers for downstream node-level tasks. GraphText~\cite{zhao2023graphtext} advocates a framework that enables training-free graph reasoning in text space by translating graphs into natural language. It incorporates the inductive bias of GNNs by constructing a graph-syntax tree and then processing natural language sequences derived from the traversal of the graph-syntax tree to perform node prediction and reasoning through LLMs. Meanwhile, zeroG~\cite{li2024zerog} proposes a zero-shot learning framework that first leverages LLMs to encode text graphs into uniform feature space and further employs the LoRA strategy to train a small language model over sampled subgraph data based on prompts.

\subsubsection{Domain Generalization v.s. Task-specific Adapation}

As demonstrated in LLMNodeBed~\cite{wu2025comprehensive}, LLMs have superior domain generalization ability, while GNNs demonstrate better task-specific adaptation by modeling structure information.
To mitigate the capability of LLMs and GNNs for GFMs, several works have attempted to align the GNNs with LLMs in node representations.
Inspired by vector quantization~\cite{gray1984vector},
LangTopo~\cite{guan_langtopo_2024} proposes a framework that aligns language descriptions of graphs with tokenized topological modeling, enabling LLMs to learn graph structures. Another work, i.e., Dr.E~\cite{liu2024multiviewempoweredstructuralgraph}, introduces a dual-residual vector quantized-variational autoEncoder that aligns LLMs with graph data in natural language. GSLM~\cite{su2024bridging} designs a co-training pipeline that trains language models and GNNs iteratively. After filtering out noisy information from raw node texts via LLMs, it iteratively optimizes an LLM and a GNN, where the LLM generates graph structures, embeddings, and pseudo labels based on cleaned text attributes. In turn, the GNN refines the graph structure and provides updated pseudo labels back to the LLM.

\subsubsection{Future Direction}

Although the node-level task has gained significant attention in the literature, and numerous studies have been proposed to study the node-level task, several directions remain under-explored and may be regarded as further directions.
Key areas include exploring graph pattern modeling beyond message-passing frameworks~\cite{wang2025gpm}, robustness~\cite{zhu2024robust}, and explainability of GNNs~\cite{li2022survey, longa2025explaining}.

\subsection{Link-Level Task}

\begin{table*}[!t]
    \centering
    \rowcolors{2}{shadecolor}{white}
    \caption{Summary of task-specific GFMs on link-level tasks. }
    \label{tab:task gfm link}
    \resizebox{\linewidth}{!}{
        \begin{tabular}{llllp{3.5cm}p{3.5cm}p{3.5cm}p{3.5cm}l}
            \toprule
            \textbf{Method Name}                              & \textbf{Tasks} & \textbf{Backbone} & \textbf{Pretrain} & \textbf{Adaptation}  & \textbf{Feature Align} & \textbf{Structure Align} & \textbf{Task Align}    & \textbf{Github}                                             \\
            \midrule
            \textbf{MOTIF} \cite{huang2025expressive}         & Link-Level         & GNN               & Supervised        & Finetune             & Data - Text Attribute  & Model - Codebook         & Explicit - Link        & -                                                           \\  
            \textbf{ULTRA} \cite{galkin2023towards}           & Link-Level         & GNN               & Supervised        & Finetune             & Data - Text Attribute  & Model - Codebook         & Explicit - Link        & \href{https://github.com/DeepGraphLearning/ULTRA}{Link}     \\  
            \textbf{KG-ICL} \cite{cui2025prompt}              & Link-Level         & GNN               & Supervised        & Finetune             & Data - Text Attribute  & Model - Codebook         & Explicit - Link        & \href{https://github.com/nju-websoft/KG-ICL}{Link}          \\  
            \textbf{UltraQuery} \cite{galkin2024foundation}   & Link-Level         & GNN               & Supervised        & Finetune             & Data - Text Attribute  & Model - Codebook         & Explicit - Link        & \href{https://github.com/DeepGraphLearning/ULTRA}{Link}     \\  
            \textbf{Cross-GCN} \cite{jiang2021cross}          & Link-Level         & GNN               & Supervised        & Finetune             & N/A                    & Loss - Auxiliary         & N/A                    & -                                                           \\  
            \textbf{UniLP} \cite{dong2024universal}           & Link-Level         & GNN               & Supervised        & Finetune             & N/A                    & N/A                      & Explicit - Subgraph    & -                                                           \\  
            \textbf{GITL} \cite{zheng2023you}                 & Link-Level         & GNN               & Supervised        & Distillation         & N/A                    & Loss - Auxiliary         & N/A                    & -                                                           \\  
            \textbf{CNHHEC} \cite{10246298}                   & Link-Level         & GNN               & Supervised        & Test-time Adaptation & N/A                    & N/A                      & Implicit - Regularizer & -                                                           \\  
            \textbf{GraphFormers} \cite{yang2021graphformers} & Link-Level         & GNN               & Contrastive       & Finetune, Prototype  & Data - Text Attribute  & N/A                      & N/A                    & \href{https://github.com/microsoft/GraphFormers}{Link}      \\  
            \textbf{ISDEA+} \cite{gao2023double}              & Link-Level         & GNN               & Contrastive       & Finetune             & Data - Text Attribute  & Model - Codebook         & Explicit - Link        & -                                                           \\  
            \textbf{MTDEA} \cite{zhou2023multi}               & Link-Level         & GNN               & Contrastive       & Finetune             & N/A                    & Model - Codebook         & Explicit - Link        & -                                                           \\  
            \textbf{Edgeformers} \cite{jinedgeformers}        & Link-Level         & GNN               & Hybrid            & Finetune             & Data - Text Attribute  & N/A                      & N/A                    & \href{https://github.com/PeterGriffinJin/Edgeformers}{Link} \\
            \bottomrule
        \end{tabular}
    }
\end{table*}

In graph learning, link-level tasks focus on understanding and predicting the relationships (or links) between nodes in a graph. These tasks are crucial for analyzing the structure and dynamics of networks, such as social networks, biological networks, or recommendation systems. These tasks include link prediction, edge classification, network completion and so on. GFMs focusing on link-level tasks need the ability to generalize and transfer knowledge across different domains and datasets. Thus, this presents several challenges. For instance, heterogeneous relationships pose a significant issue, as links can represent a variety of semantics—such as "friendship" versus "transaction" in social networks. During inference, GFMs have to learn transferable representations that enable inference on any graph with arbitrary entity and relation vocabularies. Additionally, temporal dynamics must be considered, as links in dynamic graphs, like financial networks, necessitate the modeling of time-sensitive patterns.

\subsubsection{Inductive Reasoning Approaches}
A large number of graph foundation models designed for link-specific tasks focus on the problem where novel nodes and relation types occur at test time.

\noindent \textbf{Knowledge Graph-based Methods.} Ultra, \cite{galkin2023towards}, as an approach for learning universal and transferable graph representations, leverages a transformation to convert the original graph to relation graph $\mathcal{G}_r=\text{LIFT}(\mathcal{G})$ to build a graph of relations $\mathcal{G}_r=(\mathcal{V}_l,\mathcal{R}_l,\mathcal{E}_l)$ given a graph $\mathcal{G}=(\mathcal{V,R,E})$. Then we can obtain the $d$-dimensional node representations $\mathbf{X} \in \mathbb{R}^{|\mathcal{R}| \times d}$ of $\mathcal{G}_r$ given a query $(h,q,?)$. Ultra utilizes the invariance of the relational structure and employs relative relation representations for parameterizing any unseen relation instead of relative entity representations.
Furthermore, UltraQuery \cite{galkin2024foundation}, another graph foundation model, was proposed focusing on the inductive zero-shot complex logical query answering (CLQA) problem. The novel inductive relation projection design and a differentiable but non-parametric fuzzy logical operator enable UltraQuery to be vocabulary-indepedent and generalizable to new entities and relations.

\noindent \textbf{Double Equivariance-based Methods.} Through the perspective of double equivariance (for nodes and relation types), \cite{gao2023double, zhou2023multi} tried to solve the same problem. Specifically, given a training graph $\mathbf{A}^{tr}$, with node set $\mathcal{V}^{tr}$ and relation set $\mathcal{R}^{tr}$, they aim to learn a model capable of accurately predicting missing triplets in a test graph $\mathbf{A}^{te}$, with node set $\mathcal{V}^{te}$ and relation set $\mathcal{R}^{te}$, involving both new nodes and new relations types: $\mathcal{V}^{tr} \not\subseteq \mathcal{V}^{te}$ and $\mathcal{R}^{tr} \not\subseteq \mathcal{R}^{te}$. MTDEA, \cite{zhou2023multi}, learns to partition the set of relations into distinct clusters, where each cluster exclusively contains relation types that are exchangeable among themselves seen as a multi-task setting. Accordingly, they train multiple graph models, one for each cluster (relation types) so that at test time, they can employ an adaptation procedure to assign new relation types to the most appropriate cluster, thus ensuring generalization to previously unseen relation types.

\noindent \textbf{Theoretical Analysis.} Recently, authors further demonstrated the generality of double equivariant structural representations in a theoretical framework \cite{gao2023double}. They found that Ultra \cite{galkin2023towards} and InGram \cite{lee2023ingram} conform to this framework despite their diverse architectural designs. Beyond that, they proposed a more robust and stable variant of InGram \cite{lee2023ingram}, named DEq-InGram and a modeling framework called ISDEA+, which could transform any GNNs designed for homogeneous graphs into double equivariant models suitable for knowledge graphs. Concurrently, \cite{huang2025expressive} also conducted a rigorous study of the expressive power of knowledge graph foundation models and found that the expressive power depends on the motifs that are used to learn the relation representations. Given a KG $\mathcal{G}=(\mathcal{V,R,E})$, they summarize the frameworks into three steps-use a set $\mathcal{F}$ of motifs to compute a relational hypergraph $\text{LIFT}_{\mathcal{F}}(G)$, apply a relation encoder on the hypergraph $\text{LIFT}_{\mathcal{F}}(G)$ to obtain relation representations and use the relation representations and apply an entity encoder on the KG $\mathcal{G}$ to obtain final link encodings. Additionally, they designed richer motifs than binary motifs in existing works to increase the foundation model's performance.

\subsubsection{In-Context Learning Approaches}
KG-ICL \cite{cui2025prompt}, on the other hand, used graph prompt learning and in-context learning techniques. They extract a prompt graph $\mathcal{P}_c=(\mathcal{E}_{pmt}\in \mathcal{E},\mathcal{R}_{pmt}\in \mathcal{R},\mathcal{T}_{pmt}\in \mathcal{T})$ with an example fact about the query $(u,q,v)$ and the KG $\mathcal{G}=(\mathcal{E},\mathcal{R},\mathcal{T})$ and employ a unifier tokenizer to map entities and relations in prompt graphs into predefined tokens for further training the foundation knowledge graph model. Meanwhile, UniLP \cite{dong2024universal} used in-context learning to combine the generalizability of heuristic approaches with the pattern-learning capabilities of parametric models. The built universal link predictor can autonomously identify connectivity patterns across diverse graphs, ready for immediate application to any unseen graph dataset without targeted training.

\subsubsection{Transformer-based Approaches}
With the development of attention mechanism \cite{vaswani2017attention}, some works combined GNNs and Transformers to utilize the language model's generalization ability. For instance, GraphFormers \cite{yang2021graphformers} fuse the text encoding and the graph aggregation into an iterative workflow by nesting the layerwise GNN components alongside the transformer blocks of language models. Edgeformers \cite{jinedgeformers} further incorporates text semantics on edges in a contextualized way on textual-edge networks $\mathcal{G}=(\mathcal{V},\mathcal{E},\mathcal{D})$ where each edge $e_{ij}\in \mathcal{E}$ is associated with a document $d_{ij}\in \mathcal{D}$. The inherent generalization capability of language models endows these models with the potential for transfer learning and serves as foundation models for link prediction.

\subsubsection{Hybrid Methods}

Meanwhile, there are also other works from different views. For example, DGASN \cite{10246298} attempted to study the problem of cross-network homophilous and heterophilous edge classification where we have two graphs $\mathcal{G}^A$ and $\mathcal{G}^B$ and the aligned adjacent matrix $\mathbf{A}^{A,B}\in \{0,1\}^{n_A\times n_B}$. It employs adversarial domain adaptation to mitigate domain divergence when transferring knowledge from source domains to target domains. In \cite{zheng2023you}, authors design a new setting, Graph Intersection-induced Transfer Learning (GITL), meaning that a denser graph may share nodes with a sparse original graph, which offers a natural bridge for transferring selective, meaningful knowledge. They also propose a framework to tackle this setting from two angles: training instance optimization and prediction broadcast.

\subsubsection{Future Directions}
Future graph foundation models for link-level tasks may explore in-context learning concepts within more challenging scenarios, such as dynamic and heterogeneous graphs. Another promising avenue for future research is to enhance the computational efficiency of Graph Foundation Models (GFMs), enabling scalability to real-world large-scale graphs. This is necessary because current algorithms typically demand a high pre-processing cost and extensive training time.

\subsection{Graph-Level Task}

\begin{table*}[!t]
    \centering
    \rowcolors{2}{shadecolor}{white}
    \caption{Summary of task-specific GFMs on graph-level tasks. }
    \label{tab:task gfm graph}
    \resizebox{\linewidth}{!}{
        \begin{tabular}{llllp{3.5cm}p{3.5cm}p{3.5cm}p{3.5cm}l}
            \toprule
            \textbf{Method Name}                             & \textbf{Tasks-Specific} & \textbf{Backbone} & \textbf{Pretrain} & \textbf{Adaptation}  & \textbf{Feature Align}                      & \textbf{Structure Align}            & \textbf{Task Align}    & \textbf{Github}                                             \\
            \midrule
            \textbf{AAGOD} \cite{10.1145/3580305.3599244}    & Graph-Level             & GNN               & Supervised        & Finetune             & N/A                                         & Data - Augment                      & N/A                    & \href{https://github.com/BUPT-GAMMA/AAGOD}{Link}            \\ 
            \textbf{G-Adapter} \cite{gui2024g}               & Graph-Level             & GNN               & Supervised        & Finetune             & N/A                                         & Loss - Auxiliary                    & N/A                    & -                                                           \\ 
            \textbf{GTOT-Tuning} \cite{ijcai2022p518}        & Graph-Level             & GNN               & Supervised        & Finetune             & N/A                                         & Loss - Auxiliary                    & N/A                    & -                                                           \\ 
            \textbf{CDFSGC} \cite{hassani2022cross}          & Graph-Level             & GNN               & Contrastive       & Finetune, Prototype  & Data - Node Property                        & Loss - Multi-task                   & N/A                    & -                                                           \\ 
            \textbf{L2P-GNN} \cite{Lu_Jiang_Fang_Shi_2021}   & Graph-Level             & GNN               & Contrastive       & Finetune             & N/A                                         & Model - Meta Learning               & N/A                    & \href{https://github.com/rootlu/L2P-GNN}{Link}              \\ 
            \textbf{GROVER} \cite{rong2020self}              & Graph-Level             & GNN               & Generative        & Finetune             & N/A                                         & Loss - Pretrain                     & N/A                    & -                                                           \\ 
            \textbf{G-TUNING} \cite{sun2024fine}             & Graph-Level             & GNN               & Generative        & Finetune             & N/A                                         & Model - Structure Learning          & Implicit - Codebook    & -                                                           \\ 
            \textbf{Mole-BERT} \cite{xiamole}                & Graph-Level             & GNN               & Hybrid            & Finetune             & N/A                                         & Loss - Pretrain, Model - Codebook   & N/A                    & \href{https://github.com/junxia97/Mole-BERT}{Link}          \\ 
            \textbf{AdapterGNN} \cite{li2024adaptergnn}      & Graph-Level             & GNN               & Hybrid            & Finetune             & N/A                                         & Loss - Auxiliary                    & N/A                    & \href{https://github.com/Lucius-lsr/AdapterGNN}{Link}       \\ 
            \textbf{Feature-Struct} \cite{frasca2024towards} & Graph-Level             & GNN               & Hybrid            & Finetune             & Model - Projection                          & N/A                                 & N/A                    & -                                                           \\ 
            \textbf{GPF} \cite{fang2023universal}            & Graph-Level             & GNN               & Hybrid            & Graph Prompting      & N/A                                         & Model - Prompt Learning             & N/A                    & \href{https://github.com/zjunet/GPF}{Link}                  \\ 
            \textbf{PLM-SGT} \cite{wenkel2023pretrained}     & Graph-Level             & LLM               & Supervised        & Finetune             & N/A                                         & Data - Augment                      & Explicit - QA          & -                                                           \\ 
            \textbf{GraphsGPT} \cite{gao2024graph}           & Graph-Level             & LLM               & Generative        & Finetune, Prototype  & N/A                                         & Data - Augment                      & Explicit - QA          & \href{https://github.com/A4Bio/GraphsGPT}{Link}             \\ 
            \textbf{LLM4GraphGen} \cite{yao2024exploring}    & Graph-Level             & LLM               & Generative        & In-context           & Data - Node Property, Data - Text Attribute & Data - Augment                      & Explicit - QA          & -                                                           \\ 
            \textbf{GIMLET} \cite{zhao2023gimlet}            & Graph-Level             & GNN + LLM         & Supervised        & In-context           & Data - Text Attribute                       & N/A                                 & Explicit - QA          & \href{https://github.com/zhao-ht/GIMLET}{Link}              \\ 
            \textbf{GALLON} \cite{xu2024llm}                 & Graph-Level             & GNN + LLM         & Supervised        & Distillation         & Data - Node Property                        & Loss - Auxiliary                    & N/A                    & -                                                           \\ 
            \textbf{UniMoT} \cite{zhang2024unimot}           & Graph-Level             & GNN + LLM         & Generative        & Finetune             & Data - Text Attribute                       & Model - Codebook, Model - Retriever & N/A                    & -                                                           \\ \midrule
            \textbf{DiGress} \cite{vignac2022digress}        & Graph Generation        & GNN               & Supervised        & Test-time Adaptation & N/A                                         & Data - Augment                      & N/A                    & -                                                           \\ 
            \textbf{GraphVAE} \cite{simonovsky2018graphvae}  & Graph Generation        & GNN               & Supervised        & Test-time Adaptation & N/A                                         & Data - Augment                      & N/A                    & -                                                           \\ 
            \textbf{GDSS} \cite{jo2022score}                 & Graph Generation        & GNN               & Supervised        & Test-time Adaptation & N/A                                         & Data - Augment                      & N/A                    & \href{https://github.com/harryjo97/GDSS}{Link}              \\ 
            \textbf{UniAug} \cite{tang2024cross}             & Graph Generation        & GNN               & Generative        & Test-time Adaptation & Data - Node Property                        & Data - Augment                      & N/A                    & \href{https://github.com/WenzhuoTang/UniAug}{Link}          \\ 
            \textbf{LLM4GraphGen} \cite{yao2024exploring}    & Graph Generation        & LLM               & Generative        & In-context           & Data - Node Property, Data - Text Attribute & Data - Augment                      & Explicit - QA          & -                                                           \\ 
            \textbf{InstructG2I} \cite{jin2024instructg2i}   & Graph Generation        & GNN + LLM         & Supervised        & Test-time Adaptation & N/A                                         & Data - Augment                      & N/A                    & \href{https://github.com/PeterGriffinJin/InstructG2I}{Link} \\ 
            \textbf{LGGM} \cite{wang2024large}               & Graph Generation        & GNN + LLM         & Generative        & Test-time Adaptation & N/A                                         & Data - Augment                      & Implicit - Regularizer & \href{https://lggm-lg.github.io/}{Link}                     \\
            \bottomrule
        \end{tabular}
    }
\end{table*}

Graph-level tasks for graph foundation models encompass graph classification, graph regression, graph generation, and more. Numerous researchers are engaged in pre-training and fine-tuning efforts aimed at developing a foundation model that is well-suited for cross-domain datasets and tasks. However, the designing algorithm faces challenges because graphs vary drastically across domains. For example, molecules follow strict chemical rules (e.g., valency), while social networks exhibit scale-free or community structures. They also range from small (e.g., proteins with tens of nodes) to massive (e.g., citation networks with millions of nodes), requiring flexible architectures. From the task perspective, graph-level objectives (e.g., predicting drug efficacy vs. classifying social networks) require different inductive biases, complicating multi-task learning.

\subsubsection{Pre-Training Stages}
Early works mainly focused on the pre-training strategies of graph models as they tried to build a powerful base model that could be used for various downstream tasks. In L2P-GNN \cite{Lu_Jiang_Fang_Shi_2021}, authors pre-trained the GNN model to simulate the fine-tuning process on downstream tasks, so as to directly optimize the pre-trained model’s quick adaptability to downstream tasks. During pre-training, L2P-GNN constructed a parent task $\mathcal{T}_G$ consisting of $k$ child tasks $\{\mathcal{T}_G^1,...,\mathcal{T}_G^k\}$ for a graph $\mathcal{G}$. It also designed a dual adaptation mechanism at both node and graph levels to utilize the intrinsic structures of label-free graph data as self-supervision to learn local and global representations simultaneously. Mole-BERT \cite{xiamole} proposed triplet masked contrastive learning (TMCL) using triplet $(\mathcal{G},\mathcal{G}^{M1},\mathcal{G}^{M2})$ for graph-level pre-training to model the heterogeneous semantic similarity between molecules for effective molecule retrieval. Along with Masked Atoms Modeling (MAM), their pre-training method can achieve superior performance while not requiring any domain knowledge. GROVER \cite{rong2020self} combines node-, edge-, and graph-level self-supervised tasks in their GNN-Transformer style framework. After pre-training on large-scale unlabeled molecular datasets, GROVER can learn rich implicit knowledge and transfer it to downstream graph-level tasks easily. GraphsGPT \cite{gao2024graph} designed an end-to-end pure transformer-based encoder to learn graph words $\mathcal{W}=\text{Graph2Seq}([GP]_1,[GP]_2,...,[GP]_k,FTSeq)$ and decoder (GraphGPT) to restore the graph structure ${h,p}=\text{GraphGPT}([\mathcal{W},[BOS])$. Pretrained on 100M molecules, Graph2Seq excels in graph-level tasks including graph classification and regression. Also at the same time, GraphGPT can serve as a strong graph generator. A recent work \cite{frasca2024towards} analyzed the extent to which pre-trained GNNs can be transferred across datasets by measuring the impact of pretraining datasets on downstream generalization and the inclusion of feature information via structuralization.

\subsubsection{Fine-Tuning Stages}
Another line of work, on the other hand, centered on the fine-tuning process of graph models.

\noindent \textbf{Full Fine-tuning.} G-Tuning \cite{sun2024fine} identifies the structural divergence between pre-training and downstream graphs and proposes to preserve the generative patterns of the downstream tasks, the graphon, which is a continuous and symmetric function $W:[0,1]^2 \to [0,1]$ indicating the probability of two points $u_i,u_j \in [0,1]$ forming an edge. The specific design makes G-Tuning suitable for cross-domain tasks. GTOT-Tuning \cite{ijcai2022p518} formulate graph local knowledge transfer as an Optimal Transport (OT) problem with a structural prior and construct the GTOT regularizer to constrain the fine-tuned model behaviors. By preserving the local feature invariances between fine-tuned and pre-trained models, GTOT-Tuning has great generalization ability.

\noindent \textbf{Parameter-Efficient Fine-tuning.} There are also many parameter-efficient fine-tuning (PEFT) approaches. For example, AdapterGNN \cite{li2024adaptergnn} and G-adapter \cite{gui2024g} both use the adapter module ($\mathbf{A}(x)=\text{BN}(\mathbf{W}_{up}(\text{ReLU}(\mathbf{W}_{down}(\mathbf{x}))))$) in the GNN area. By introducing a small amount of tunable parameters, they can outperform the traditional full fine-tuning method while improving the base model's generalization. GPF \cite{fang2023universal}, as a prompt learning technique, injects learnable prompt vectors into the feature space of the original datasets. Specifically, given a learnable prompt vector $\boldsymbol{p}_i$, node $v_i$ will have a prompted feature vector $\tilde{\boldsymbol{x}}_i = \boldsymbol{x}_i + \boldsymbol{p}_i$ and GPF simply sets a prompt vector $\boldsymbol{p}$ shared by all the nodes, i.e., $\boldsymbol{p}_1=\boldsymbol{p}_2=\cdots=\boldsymbol{p}_n=\boldsymbol{p}$.
Authors also provide rigorous derivations to demonstrate the universality of GPF and make a guarantee of its effectiveness. The effectiveness and efficacy position these PEFT methods as a compelling alternative to full fine-tuning for downstream adaptations, such as molecular graph classification and regression.

\subsubsection{LLM-incorporated Approaches}
Recently with the prosperity of LLMs, some works are utilizing the inherited knowledge from LLMs to help solve the graph-level problems.

\noindent \textbf{LLM for Understanding.} In an early work \cite{wenkel2023pretrained}, authors first transformed the graphs into pure text and then fine-tuned GPT-2 and GPT-3 using natural language. Results on molecular classification demonstrate the promising of this direction. GALLON \cite{xu2024llm} utilized multimodal molecular data to learn representations and extract prior knowledge from powerful and pre-trained large language models (e.g., GPT-4V) using prompt by leveraging their multimodality capabilities, i.e. $\mathcal{R_i}=\text{LLM}(\mathcal{P_i},\mathcal{E_i},\mathcal{S_i},\mathcal{I_i})$ for molecule $\mathcal{G_i}$. It further distills the advantages of GNN and LLM into an MLP, aiming to capture the most effective representations for molecular structures. GIMLET \cite{zhao2023gimlet} also extends language models to handle graph and text data by applying the transformer mechanism with generalized position embedding and decoupled attention. It incorporates molecule graphs $\mathcal{G}$ and task instructions $T$ into the graph-text language model GIMLET and decodes the output as text uniformly for different tasks, i.e. $\hat{y}_{str}=\text{GIMLET}(\mathcal{G},T)$, where $\hat{y}_{str}$ is the label string.
Instruction-based pretraining expressed in natural language enables GIMLET to transfer to a broad range of zero-shot graph-level tasks. UniMoT \cite{zhang2024unimot} introduced a molecule tokenizer specifically designed for LLMs, enabling the tokenization of molecules into short sequences of causal-dependent molecule tokens. It can unify the modalities of molecule $\{s_i\}_{i=1}^{M}$ and text $\{t_i\}_{i=1}^{M}$ under a shared token representation and an autoregressive training paradigm. With the help of LLMs and multi-stage training strategies, UniMoT excels at both graph comprehension and generation tasks.

\noindent \textbf{LLM for Generation.} LLM4GraphGen \cite{yao2024exploring}, on the other hand, explores the ability of LLMs for graph generation with systematical task designs and extensive experiments. They designed comprehensive experiments to evaluate the graph generation ability of LLMs by proposing tasks with varying difficulty, including rule-based graph generation, distribution-based graph generation, and property-based graph generation. Diverse LLMs and prompts revealed insightful observations.

\subsubsection{Graph Generation}

Graph generation, a crucial subdomain of graph-level tasks, seeks to generate graphs that adhere to specific rules, distributions, or domain-based properties. Designing GFMs for such tasks requires handling a wide spectrum of structural and semantic complexities across domains. Traditional graph generative models such as GraphVAE \cite{simonovsky2018graphvae}, GDSS \cite{jo2022score}, and DiGress \cite{vignac2022digress} are typically limited to single domains and struggle with generalization. Recent works like UniAug \cite{tang2024cross} address this by incorporating diffusion models that scale across diverse graph distributions, aiming for universality. In particular, it employs a structure-only discrete diffusion model to pre-train on thousands of graphs and augment downstream datasets with guided structure synthesis, improving generalization without relying on feature similarity. These approaches signify a shift toward cross-domain scalability in graph generation, aligning GFMs with the multi-task and multi-domain capabilities observed in language and vision foundation models.

Inspired by the success of large generative models like Stable Diffusion and GPT, recent graph foundation models embrace large-scale pre-training and cross-modal prompting to enhance graph generation. LGGM \cite{wang2024large} pre-trains on over 5,000 graphs from 13 domains, encoding diverse structural patterns that enable superior zero-shot and fine-tuned generation. LGGM also introduces Text-to-Graph generation, wherein users provide textual prompts—such as graph domains or structural statistics (e.g., clustering coefficient)—to guide generation, leveraging the world knowledge embedded in language models. In parallel, InstructG2I \cite{jin2024instructg2i} proposes a multimodal approach where graph-structured data enriches with image and text attributes to guide diffusion-based image generation. Its graph-conditioned generation is both expressive and controllable, offering smooth interpolation across styles or domains. These pre-training and prompting strategies reflect the growing synergy between LLMs and GNNs, expanding the frontier of GFMs toward open-ended, user-controllable graph synthesis.

\subsubsection{Hybrid Approaches}

Meanwhile, there are also other relevant works regarding graph models.

\noindent \textbf{Few-shot Graph Classification.} For instance, CDFSGC \cite{hassani2022cross} and CDTC \cite{article} studied the problem of few-shot graph classification across domains. In this setting, we aim to learn a model with good generalization ability that can predict the label of graphs in the target domain $\mathcal{D}^T$ given few-shot annotated examples from the target domain and the source domain data $\mathcal{D}^S$ where $\mathcal{D}^T$ and $\mathcal{D}^S$ have different marginal distributions $\mathcal{P}_{\mathcal{D}^T}$ and $\mathcal{P}_{\mathcal{D}^S}$, and the label space $\mathcal{Y}^S$ and $\mathcal{Y}^T$ are disjoint. CDFSGC \cite{hassani2022cross} proposed a graph encoder that learns to attend to three congruent views of graphs, one contextual and two topological views, to learn representations of task-specific information for fast adaptation and task-agnostic information for knowledge transfer. Coupled with metric-based meta-learning frameworks, their method achieved great performance across three graph classification tasks in different domains. Furthermore, to tackle the domain shift issue, CDTC \cite{article} designed a novel Cross-domain Task Coordinator (CDTC) to leverage a small set of labeled target domain data as prompt tasks $\{\mathbf{p}^t\}_{t=1}^T$, then model the association and discover the relevance between meta-tasks from the source domain and the prompt tasks. After being integrated with the optimization-based meta-learning process and trained with reinforcement learning in an end-to-end manner, CDTC excels at multiple cross-domain few-shot graph classification tasks.

\noindent \textbf{Out-of-Distribution Detection.} AAGOD \cite{10.1145/3580305.3599244} endowed a well-trained GNN with the OOD detection ability across domains where we have $\mathcal{P}_{in}$ and $\mathcal{P}_{out}$, two distinct distributions defined in graph space. In the training phase, a graph dataset $\mathcal{D}_{id}=\{G^1,...,G^n \}$ sampled from the in-distribution $\mathcal{P}_{in}$ is available for model learning and the general purpose of is to distinguish whether a graph belongs to in-distribution $\mathcal{P}_{in}$  or not at the test phase. AAGOD does not require modifying its parameters by designing an effective framework with the Learnable Amplifier Generator (LAG) and Regularize Learning Strategy (RLS).

\subsubsection{Future Directions}
The future of graph foundation models (GFMs) for graph-level tasks lies in developing unified and flexible architectures, such as Graph Transformers with structural encoding, to capture domain-agnostic properties while enabling hierarchical pooling for multi-scale representations. Scalability challenges can be mitigated through subgraph sampling, linearized attention, and distributed training, while meta-learning and prompt-based fine-tuning will enable rapid adaptation to new domains. Robustness to noise and sparsity can be improved via graph denoising autoencoders and data augmentation. Integrating multi-modal data (e.g., text, images) and creating cross-domain benchmarks with unified metrics will further advance evaluation and standardization. Finally, prioritizing explainability through interpretable pooling and fairness via adversarial debiasing will ensure trustworthy and ethical AI applications, paving the way for GFMs to revolutionize domains like drug discovery, fraud detection, and climate modeling.

\begin{table*}[!t]
    \centering
    \rowcolors{2}{shadecolor}{white}
    \caption{Summary of task-specific GFMs on question answering, recommendation, anomaly detection. }
    \label{tab:task gfm others}
    \resizebox{\linewidth}{!}{
        \begin{tabular}{llllp{3.5cm}p{3.5cm}p{3.5cm}p{3.5cm}l}
            \toprule
            \textbf{Method Name}                              & \textbf{Tasks}     & \textbf{Backbone} & \textbf{Pretrain} & \textbf{Adaptation}  & \textbf{Feature Align} & \textbf{Structure Align} & \textbf{Task Align} & \textbf{Github}                                       \\
            \midrule
            \textbf{MCDGRAPH} \cite{zhu2024benchmarking}      & Question Answering & GNN               & Generative        & In-context           & Data - Text Attribute  & N/A                      & Explicit - QA       & \href{https://github.com/AAAndy-Zhu/VGCure}{Link}     \\
            \textbf{GT2Vec} \cite{lin_gt2vec_2025}            & Question Answering & LLM               & Contrastive       & Finetune             & Data - Text Attribute  & Data - Augment           & Explicit - QA       & -                                                     \\
            \textbf{GPT-4V-GSR} \cite{ai2023graph}            & Question Answering & LLM               & Generative        & In-context           & Data - Text Attribute  & N/A                      & Explicit - QA       & -                                                     \\
            \textbf{G-Retriever} \cite{he_g-retriever_nodate} & Question Answering & GNN + LLM         & Supervised        & In-context           & Data - Text Attribute  & N/A                      & Explicit - QA       & \href{https://github.com/XiaoxinHe/G-Retriever}{Link} \\
            \textbf{GITA} \cite{wei_gita_2024}                & Question Answering & GNN + LLM         & Generative        & Finetune             & Data - Text Attribute  & N/A                      & Explicit - QA       & -                                                     \\
            \textbf{GFM-RAG} \cite{luo2025gfm}                & Question Answering & GNN + LLM         & Hybrid            & Finetune             & Data - Text Attribute  & N/A                      & N/A                 & \href{https://github.com/RManLuo/gfm-rag}{Link}       \\ \midrule
            \textbf{SR-MDFM} \cite{gong2023unified}           & Recommendation     & GNN               & Supervised        & Test-time Adaptation & Data - Text Attribute  & Data - Augment           & N/A                 & -                                                     \\
            \textbf{PCRec} \cite{wang2021pre}                 & Recommendation     & GNN               & Contrastive       & Finetune             & N/A                    & Loss - Pretrain          & Explicit - Subgraph & -                                                     \\
            \textbf{LLMRec} \cite{wei2024llmrec}              & Recommendation     & GNN               & Hybrid            & In-context           & Data - Text Attribute  & Data - Augment           & Explicit - QA       & \href{https://github.com/HKUDS/LLMRec}{Link}          \\
            \textbf{VIP5} \cite{geng2023vip5}                 & Recommendation     & LLM               & Generative        & In-context           & Data - Text Attribute  & Data - Augment           & Explicit - QA       & \href{https://github.com/jeykigung/VIP5}{Link}        \\
            \textbf{2T} \cite{damianou2024towards}            & Recommendation     & GNN + LLM         & Supervised        & Finetune             & Data - Text Attribute  & N/A                      & N/A                 & -                                                     \\
            \textbf{RLMRec} \cite{ren2024representation}      & Recommendation     & GNN + LLM         & Hybrid            & Finetune             & Data - Text Attribute  & Loss - Pretrain          & N/A                 & \href{https://github.com/HKUDS/RLMRec}{Link}          \\ \midrule
            \textbf{AnomalyGFM} \cite{qiao_anomalygfm_2025}   & Anomaly Detection  & GNN               & Supervised        & Prototype            & Model - Projection     & Model - Prompt Learning  & Explicit - Link     & \href{https://github.com/mala-lab/AnomalyGFM}{Link}   \\
            \textbf{CDFS-GAD} \cite{chen2024towards}          & Anomaly Detection  & GNN               & Contrastive       & Graph Prompting      & Model - Projection     & Model - Prompt Learning  & N/A                 & -                                                     \\
            \textbf{ACT} \cite{wang2023cross}                 & Anomaly Detection  & GNN               & Contrastive       & Prototype            & N/A                    & Loss - Pretrain          & N/A                 & \href{https://github.com/QZ-WANG/ACT}{Link}           \\
            \textbf{UNPrompt} \cite{niu2024zero}              & Anomaly Detection  & GNN               & Contrastive       & Test-time Adaptation & Model - Projection     & Data - Augment           & N/A                 & \href{https://github.com/WenzhuoTang/UniAug}{Link}    \\
            \textbf{ARC} \cite{liu2025arc}                    & Anomaly Detection  & GNN               & Generative        & Prototype            & Model - Projection     & N/A                      & N/A                 & \href{https://github.com/yixinliu233/ARC}{Link}       \\
            \textbf{Commander} \cite{ding2021cross}           & Anomaly Detection  & GNN               & Hybrid            & Test-time Adaptation & N/A                    & Loss - Pretrain          & N/A                 & -                                                     \\
            \bottomrule
        \end{tabular}
    }
\end{table*}

\subsection{Question Answering}

Question Answering (QA) is another type of downstream task in GFMs.
Existing studies for QA tasks commonly employ GNNs as enhancer and further leverage LLMs to generate answers.
GFM-RAG~\cite{luo2025gfm} designs a graph foundation model with retrieval augmented generation for QA tasks.
It first builds a knowledge graph index (KG-index) from the documents to capture the relationships between entities.
Then, GFM-RAG feeds the query and the constructed KG-index into the pre-trained GFM retriever to obtain relevant documents for LLM generation.
The GFM retriever experiences large-scale training and can be directly applied to unseen datasets without fine-tuning.
Similarly, G-Retriever~\cite{he_g-retriever_nodate} introduces additional subgraph construction step before LLM generation for the QA tasks.
Moreover, G-Retriever also provides a GraphQA benchmark, that contains three datasets, i.e., ExplaGraphs~\cite{saha2021explagraphs}, SceneGraphs~\cite{hudson2019gqa}, and WebQSP~\cite{yih2016value, luo2023reasoning}.
GITA~\cite{wei_gita_2024} introduces a graph visualizer component to obtain graph visualizations.
These constructed visual graphs along with textual descriptions for graph structures are fed into visual language models (VLMs) to obtain QA tasks.
Another work~\cite{ai2023graph} proposes a paradigm for understanding and reasoning over graph image data by integrating image encoding and multimodal technologies, i.e., OCR.
VGCURE~\cite{zhu2024benchmarking} introduces a comprehensive benchmark that covers 22 tasks to examine the fundamental understanding and reasoning capabilities of VLMs.
GT2VEC~\cite{lin_gt2vec_2025} is a framework that learns joint embeddings of text and graph data using Large Language Models (LLMs) by projecting graph embeddings into the text embedding space and employing contrastive learning for alignment.
This approach enhances semantic coherence between the modalities for QA tasks.

\noindent\textbf{Future Directions.}
Future directions for developing GFMs for QA tasks focus on improving adaptability, scalability, and reasoning capabilities.
One promising direction is dynamic graph construction from unstructured data during retrieval, which enables models to create context-specific graphs on the fly~\cite {tian2024graph, li2024graph}.
Multimodal integration is another key direction, which allows GFMs to handle textual, visual, and other modalities for richer reasoning capability~\cite{choimultimodal, xu2024llm}.
Additionally, improving explainability and transparency in GFMs will be critical for applications requiring trust and accountability, such as medical or legal QA systems~\cite{choimultimodal, tian2024kg}.

\subsection{Graph Anomaly Detection}
Graph anomaly detection (GAD) aims to identify anomaly samples in terms of node and structure levels that deviate from the majority of samples.
Early studies~\cite{wang2023cross, ding2021cross} focus on cross-domain graph anomaly detection (CD-GAD), i.e., detecting anomalous nodes in an unlabeled target graph by training models over auxiliary, related source graphs with labeled abnormal and normal nodes.
These studies commonly employ domain adaptation approaches to tackle this problem.
Commander~\cite{ding2021cross} introduces three components, i.e., a domain discriminator for domain alignment, an anomaly classifier to detect anomalies, and an attribute decoder to provide additional signals for assessing node abnormality, to study the cross-domain graph anomaly detection tasks.
Another study, i.e. ACT~\cite{wang2023cross}, also proposes a domain adaptation approach that jointly optimizes: (i) unsupervised contrastive learning over normal representations of nodes in the target graph and (ii) anomaly-aware one-class alignment that aligns contrastive node representations and representations of labeled normal nodes in the source graph.
Moreover, in the contrastive learning stage, CD-GAD enforces deviation of normal node representations from labeled anomalous nodes in the source graph.
Besides CD-GAD, CDFS-GAD~\cite{chen2024towards} proposes a more prevalent and complex scenario of cross-domain few-shot graph anomaly detection.
The goal is to identify anomalies within sparsely labeled target graphs using auxiliary graphs from a related yet distinct domain.
To address this problem, CDFS-GAD introduces a prompt tuning module to extract domain-specific features tailored to each domain, further design an adaptive hypersphere classification loss to enhance the discrimination between normal and abnormal instances via domain-sensitive norms.

More recently, several studies have started to focus on GFMs for GAD.
ARC~\cite{liu2025arc} designs a "one-for-all" GAD model to detect anomalies across various graph datasets on-the-fly.
It is equipped with the in-context learning to extract dataset specific patterns from the target datasets with few-shot normal samples at the inference stage, without the need for fine-tune on the target dataset.
Other studies focus on zero-shot GAD tasks, i.e., no label information in the target graphs are provided.
UNPrompt~\cite{niu2024zero}, introduces a simple prompt tuning module that captures the generalized patterns among latent attributes of normal nodes while minimizing that of abnormal nodes.
Another study, AnomalyGFM~\cite{qiao_anomalygfm_2025}, proposes a GFM for graph anomaly detection that leverages graph-agnostic representations to achieve strong zero-shot generalization.
Besides, if the labels of samples s are available, it supports few-shot prompt tuning across diverse graph datasets.
By aligning learnable normal and abnormal class prototypes with node representation residuals, AnomalyGFM distills discriminative features, enabling effective anomaly measurement in a unified feature space.

\noindent\textbf{Future Directions.}
GAD methods for dynamic and heterogeneous graphs represent a promising research direction, as real-world graphs are typically dynamic—with nodes and edges continuously evolving—and heterogeneous in both node attributes and structures,  e.g., social networks and financial transaction networks~\cite{mir2023graph, kim2022graph}.
Enhancing the robustness of GNNs against adversarial attacks and noisy data is another critical research focus~\cite{bei2024guarding}.
Furthermore, improving the interpretability of GNN-based anomaly detection models is essential for fostering trust, particularly in high-stakes applications, e.g., cybersecurity and fraud detection, etc~\cite{mir2023graph}.

\subsection{Recommendation}
Recommendation systems, an important branch of AI, are trained to understand the preferences, previous decisions, and characteristics of people and products using data gathered about their interactions. Graph-based recommenders have demonstrated impressive capabilities in capturing complex user-item relationships, making them state-of-the-art approaches. With the development of GFMs and LLMs, they provide a new perspective for modern recommendation systems. The challenges are mainly due to the complexity of scalability and real-time requirements, domain-specific semantics, and dynamic user behavior. Specifically, platforms like Amazon involve graphs with billions of nodes and edges, straining computational resources. Different domains involve distinct interaction semantics (e.g., "purchase" in e-commerce vs. "follow" in social networks) and user interests may shift over time, requiring models to adapt to temporal patterns.

\noindent \textbf{GNN-based Approaches.} An early work, PCRec \cite{wang2021pre}, developed a pre-training Graph Neural Network (GNN) model for the cross-domain recommendation which adopts a contrastive self-supervised pre-training strategy. Then, the pre-trained GNN encoder can be initialized to generate node embeddings on the target domain and fine-tuned by a bipartite recommendation system using a BPR loss.

\noindent \textbf{LLM-based Approaches.} Targeting the scarcity of implicit feedback signals in recommendations, LLMRec \cite{wei2024llmrec} enhances recommender systems by incorporating LLMs to augment user-item interaction edges, item node attributes, and user node profiles by $\mathcal{P}_u = \text{LLM}(S_u, Q_u), \quad \mathcal{P}_v = \text{LLM}(S_v, Q_v)$. Training GNN backbones with a denoised data robustification mechanism enabled LLMRec to achieve great performance on various benchmarks. RLMRec \cite{ren2024representation}, as a model-agnostic framework, is another method aiming to enhance existing GNN-based recommenders with LLM-empowered representation learning. RLMRec also utilizes contrastive and generative alignment techniques to align Collaborative Filtering (CF)-side relational embeddings with LLM-side semantic representations, effectively reducing feature noise. In \cite{damianou2024towards}, authors presented a graph-based foundation modeling approach tailored to personalization for the first time. They combined the advantages of LLMs and heterogeneous GNNs (HGNNs) and designed a two-tower (2T) architecture so that while the HGNN produces general-purpose embeddings, the 2T component models the sheer size of user-item interaction data in a continuous space. The benefit of such an approach is that it unifies representation learning across various tasks and enables information sharing.

\noindent \textbf{Future Directions.} There are several promising future directions for the advancement of graph foundation models in recommendation systems. For instance, we could design lightweight GFM architectures and distributed training frameworks to enhance scalability and effectively manage large-scale graphs. Additionally, integrating graph structures with multi-modal data, such as text and images, could be explored. Adopting prompt-based learning for task-specific recommendations or introducing dynamic GFMs that capture real-time interactions and user interest shift are also viable strategies to consider.

\newpage

\section{Domain-Specific Graph Foundation Models}
\label{sec:domain gfm}

\subsection{Design Principle}

Despite the general-purpose nature of foundation models, there is growing interest in designing \textit{domain-specific GFMs}. In this setting, a single model learns shared representations that generalize across related tasks within a specific domain. Constructing such models is non-trivial, as they must effectively capture the underlying principles and key properties of the target domain. For instance, in molecular graphs, the model must recognize and preserve key motifs and functional groups, while in knowledge graphs, it must infer relationships and correlations between triplets. This section outlines the core characteristics and design principles essential for developing domain-specific GFMs.

\begin{itemize}[leftmargin=10pt]
    \item \textbf{Domain-Specific Expertise.} Different domains exhibit distinct structural properties and encode unique forms of knowledge. There is no universal inductive bias \cite{mao2024graph} that can simultaneously capture the diverse characteristics required for every domain. Therefore, domain-specific GFMs necessitate customized model architectures, pretraining paradigms, and adaptation strategies. For example, in molecular graphs, subgraph-level augmentation is crucial, as essential semantic components (e.g., aromatic rings and functional groups) are preserved at the subgraph level. Conversely, knowledge graphs often require node-level or edge-level augmentation to generate additional triplets and enhance relational reasoning.

    \item \textbf{Learning Task-Agnostic Representations.} Tasks and graphs within the same domain typically exhibit strong correlations, making it possible to learn a common graph representation that benefits multiple downstream tasks. Techniques such as multi-task learning \cite{shoghi2023molecules,zhang2024dpa}, adversarial learning, augmentation strategies, and domain regularization can be employed to achieve robust and generalizable representations. However, despite these benefits, task interference remains a challenge, potentially leading to negative transfer effects. To mitigate this, incorporating task-aware output heads \cite{batatia2023foundation} or advanced task alignment strategies \cite{zhao2023gimlet,taylor2022galactica} can improve task-specific adaptation while maintaining shared domain knowledge.

    \item \textbf{Enhancing Interpretability and Trustworthiness.} In many domain-specific applications (e.g., drug discovery, mathematical reasoning, academic research), it is not only important for GFMs to achieve high performance on downstream tasks but also to generate interpretable insights that contribute to domain advancement. For instance, a GFM pretrained on large-scale molecular datasets may aid in the discovery of novel functional groups, reaction mechanisms, or physicochemical properties. To enable such discoveries, domain-specific GFMs must prioritize interpretability and trustworthiness. This may involve incorporating explainable AI techniques, uncertainty quantification, and domain-specific validation mechanisms.

\end{itemize}

In the following subsections, we systematically explore the design principles of GFMs across eight distinct domains: molecular graphs, heterogeneous graphs, knowledge graphs, temporal graphs, academia, graph-based mathematical reasoning, causal graphs, and semantic document graphs. For each domain, we discuss the underlying design philosophies, key challenges, state-of-the-art methodologies, and promising directions for future research.

\subsection{Biology \& Molecule Graph}

\begin{table*}[!t]
    \centering
    \rowcolors{2}{shadecolor}{white}
    \caption{Summary of domain-specific GFMs on biology and molecule graphs. }
    \label{tab:domain gfm molecule}
    \resizebox{\linewidth}{!}{
        \begin{tabular}{llllp{3.5cm}p{3.5cm}p{3.5cm}p{3.5cm}l}
            \toprule
            \textbf{Method Name}                              & \textbf{Domain}           & \textbf{Backbone} & \textbf{Pretrain} & \textbf{Adaptation}    & \textbf{Feature Align} & \textbf{Structure Align}            & \textbf{Task Align}  & \textbf{Github}                                                    \\
            \midrule
            \textbf{MiniMol} \cite{klaser2024minimol}         & Biology \& Molecule Graph & GNN               & Supervised        & Finetune               & N/A                    & Data - Augment                      & Implicit - Aux. Loss & -                                                                  \\
            \textbf{DPA-2} \cite{zhang2024dpa}                & Biology \& Molecule Graph & GNN               & Supervised        & Distillation, Finetune & N/A                    & Loss - Multi-task                   & N/A                  & \href{https://zenodo.org/records/10428497}{Link}                   \\
            \textbf{JMP} \cite{shoghi2023molecules}           & Biology \& Molecule Graph & GNN               & Supervised        & Finetune               & Data - Node Property   & Loss - Multi-task                   & N/A                  & \href{https://nima.sh/jmp/}{Link}                                  \\
            \textbf{MACE} \cite{batatia2023foundation}        & Biology \& Molecule Graph & GNN               & Supervised        & Finetune               & Data - Node Property   & N/A                                 & N/A                  & \href{https://github.com/ACEsuit/mace/}{Link}                      \\
            \textbf{MolGPS} \cite{sypetkowski2025scalability} & Biology \& Molecule Graph & GNN               & Supervised        & Test-time Adaptation   & N/A                    & N/A                                 & N/A                  & \href{https://github.com/datamol-io/graphium}{Link}                \\
            \textbf{DiG} \cite{zheng2023towards}              & Biology \& Molecule Graph & GNN               & Generative        & Finetune               & Data - Text Attribute  & Data - Augment                      & N/A                  & \href{https://distributionalgraphormer.github.io/}{Link}           \\
            \textbf{GROVER} \cite{rong2020self}               & Biology \& Molecule Graph & GNN               & Generative        & Finetune               & N/A                    & Loss - Pretrain                     & N/A                  & -                                                                  \\
            \textbf{GTFM} \cite{mizera2024graph}              & Biology \& Molecule Graph & GNN               & Generative        & Finetune               & N/A                    & Loss - Multi-task                   & N/A                  & -                                                                  \\
            \textbf{Mole-BERT} \cite{xiamole}                 & Biology \& Molecule Graph & GNN               & Hybrid            & Finetune               & N/A                    & Loss - Pretrain, Model - Codebook   & N/A                  & \href{https://github.com/junxia97/Mole-BERT}{Link}                 \\ \midrule
            \textbf{MolecularGPT} \cite{liu2024moleculargpt}  & Biology \& Molecule Graph & LLM               & Supervised        & Finetune, In-context   & Data - Text Attribute  & N/A                                 & Explicit - QA        & \href{https://github.com/NYUSHCS/MolecularGPT}{Link}               \\
            \textbf{GP-GPT} \cite{lyu2024gp}                  & Biology \& Molecule Graph & LLM               & Supervised        & Finetune               & Data - Text Attribute  & Data - Augment                      & Explicit - QA        & -                                                                  \\
            \textbf{BioBRIDGE} \cite{wang2023biobridge}       & Biology \& Molecule Graph & LLM               & Contrastive       & In-context             & Data - Text Attribute  & Loss - Pretrain                     & Explicit - QA        & \href{https://github.com/RyanWangZf/BioBridge}{Link}               \\
            \textbf{GraphsGPT} \cite{gao2024graph}            & Biology \& Molecule Graph & LLM               & Generative        & Finetune, Prototype    & N/A                    & Data - Augment                      & Explicit - QA        & \href{https://github.com/A4Bio/GraphsGPT}{Link}                    \\
            \textbf{ESMFold} \cite{lin2022language}           & Biology \& Molecule Graph & LLM               & Generative        & Finetune               & Data - Text Attribute  & N/A                                 & Explicit - QA        & -                                                                  \\
            \textbf{CaR} \cite{qian2023can}                   & Biology \& Molecule Graph & LLM               & Generative        & Finetune, In-context   & Data - Text Attribute  & Data - Augment                      & Explicit - QA        & \href{https://github.com/ChnQ/LLM4Mol}{Link}                       \\ \midrule
            \textbf{InstructMol} \cite{cao2023instructmol}    & Biology \& Molecule Graph & GNN + LLM         & Supervised        & Finetune               & Data - Text Attribute  & Loss - Auxiliary, Model - Retriever & Explicit - QA        & \href{https://github.com/IDEA-XL/InstructMol}{Link}                \\
            \textbf{ALIGNN} \cite{li2025hybrid}               & Biology \& Molecule Graph & GNN + LLM         & Supervised        & Finetune               & Data - Text Attribute  & N/A                                 & N/A                  & \href{https://github.com/Jonathanlyj/ALIGNN-BERT-TL-crystal}{Link} \\
            \textbf{GIMLET} \cite{zhao2023gimlet}             & Biology \& Molecule Graph & GNN + LLM         & Supervised        & In-context             & Data - Text Attribute  & N/A                                 & Explicit - QA        & \href{https://github.com/zhao-ht/GIMLET}{Link}                     \\
            \textbf{GALLON} \cite{xu2024llm}                  & Biology \& Molecule Graph & GNN + LLM         & Supervised        & Distillation           & Data - Node Property   & Loss - Auxiliary                    & N/A                  & -                                                                  \\
            \textbf{Text2Mol} \cite{edwards2021text2mol}      & Biology \& Molecule Graph & GNN + LLM         & Contrastive       & Finetune               & Data - Text Attribute  & Loss - Multi-task                   & N/A                  & \href{https://github.com/cnedwards/text2mol}{Link}                 \\
            \textbf{MoleculeSTM} \cite{liu2023multi}          & Biology \& Molecule Graph & GNN + LLM         & Contrastive       & Finetune               & Data - Text Attribute  & Loss - Pretrain                     & Explicit - QA        & \href{https://github.com/chao1224/MoleculeSTM/tree/main}{Link}     \\
            \textbf{CLAMP} \cite{seidl2023enhancing}          & Biology \& Molecule Graph & GNN + LLM         & Contrastive       & Finetune               & Data - Text Attribute  & Loss - Pretrain                     & Explicit - QA        & \href{https://github.com/ml-jku/clamp}{Link}                       \\
            \textbf{GIT-Mol} \cite{liu2024git}                & Biology \& Molecule Graph & GNN + LLM         & Contrastive       & Finetune               & Data - Text Attribute  & Loss - Multi-task                   & Explicit - QA        & \href{https://github.com/AI-HPC-Research-Team/GIT-Mol}{Link}       \\
            \textbf{MolFM} \cite{luo2023molfm}                & Biology \& Molecule Graph & GNN + LLM         & Contrastive       & Finetune               & Data - Text Attribute  & Loss - Pretrain                     & Explicit - QA        & \href{https://github.com/PharMolix/OpenBioMed}{Link}               \\
            \textbf{MoMu} \cite{su2022molecular}              & Biology \& Molecule Graph & GNN + LLM         & Contrastive       & Finetune, In-context   & Data - Text Attribute  & Loss - Pretrain                     & Explicit - QA        & \href{https://github.com/BingSu12/MoMu}{Link}                      \\
            \textbf{UniMoT} \cite{zhang2024unimot}            & Biology \& Molecule Graph & GNN + LLM         & Generative        & Finetune               & Data - Text Attribute  & Model - Codebook, Model - Retriever & N/A                  & \href{https://uni-mot.github.io/}{Link}                            \\
            \textbf{MolCA} \cite{liu2023molca}                & Biology \& Molecule Graph & GNN + LLM         & Generative        & Finetune               & Data - Text Attribute  & Model - Retriever                   & Explicit - QA        & \href{https://github.com/acharkq/MolCA}{Link}                      \\
            \textbf{ReLM} \cite{shi2023relm}                  & Biology \& Molecule Graph & GNN + LLM         & Generative        & In-context             & Data - Node Property   & Model - Retriever                   & Explicit - QA        & \href{https://github.com/syr-cn/ReLM}{Link}                        \\
            \bottomrule
        \end{tabular}
    }
\end{table*}

Molecular graphs present unique challenges distinct from other graph domains due to their rich atomic and bond-level structural complexity and chemical feature diversity. Unlike homogeneous graphs, molecular representations must capture chemical symmetries, such as invariance under rotations, translations, and atom permutations \cite{musil2021physics}. Additionally, molecular properties arise from intricate interactions spanning multiple scales—from local functional groups to global molecular topology—and are influenced by quantum effects and flexible 3D conformations not captured by purely 2D connectivity \cite{stark20223d}. Further challenges include scalability to an astronomically large molecular space and limited labeled data due to costly experiments \cite{wang2022molecular}. These issues necessitate specialized graph foundation models integrating symmetry-aware, self-supervised, and physics-informed approaches for reliable molecular prediction and generalization.

\subsubsection{Graph Model-based Approaches}

Graph data format is a natural fit for molecules because each molecule can be represented as a graph of atoms and bonds. Graph-based methods such as message-passing operators in Graph Neural Networks, permit explicit modeling of local and long-range interactions within the graph, providing a structure-aware representation often crucial for accurate property predictions.

\noindent\textbf{3D Graph.} One line of research focuses on developing \emph{equivariant} graph models that incorporate three-dimensional (3D) coordinates and symmetries. The Equivariant Foundation Model for Atomistic Materials Chemistry \cite{batatia2023foundation} preserves rotational and translational symmetries, thereby enabling accurate classification and link prediction in materials chemistry. Similarly, Joint Multi-domain Pre-training (JMP) \cite{shoghi2023molecules} employs GemNet-OC to bridge small molecules, catalysts, and bulk materials for atomic property prediction, highlighting that an equivariant 3D message-passing strategy can generalize across diverse chemical domains.

\noindent\textbf{2D Graph.} Other works concentrate on purely 2D connectivity and multi-task learning. Mole-BERT \cite{xiamole} combines a Graph Isomorphism Network (GIN) with masked-atom modeling and contrastive tasks, advancing property prediction by leveraging learned atom-level representations. Graphium \cite{beaini2023towards} similarly supports multi-task learning across quantum-mechanical and bioassay datasets, employing GCNs or GINE variants. These frameworks underscore how large-scale or multi-task pre-training on 2D molecular graphs can yield strong predictive performance on tasks such as toxicity classification or ADMET endpoints.

\noindent\textbf{Efficiency-Focused Methods.} While many of those graph models aim for maximal accuracy, some recent efforts prioritize parameter efficiency. MiniMol \cite{klaser2024minimol} adopts a compact GNN-based design and multi-task pre-training to handle quantum and biological assays simultaneously. By using fewer parameters while retaining robust predictive power, MiniMol exemplifies a trend toward more deployable, foundation-like GNN architectures. Altogether, these approaches show that well-designed message passing—either in 2D or with explicit 3D coordinates—remains a reliable backbone for molecular property prediction, and serves as a springboard for subsequent multimodal or language-driven models.

\subsubsection{Language Model-based Approaches}

Language-based methods are motivated by the remarkable success of Transformers in capturing complex dependencies in sequences, including natural language. Because molecules can be linearized (for instance, via SMILES) or otherwise tokenized, researchers have explored purely Transformer architectures to model chemical or biological sequences, effectively treating them like language data.

\noindent\textbf{Transformer-based Models.} A prominent branch of LM-based research applies Transformers directly to molecular graphs. GROVER \cite{rong2020self} merges GNN-like message passing with global self-attention for large-scale property prediction. DiG \cite{zheng2024predicting} (built on a Graphormer-style design) models entire equilibrium distributions of molecular conformations, providing thermodynamic insights beyond simple endpoint predictions. GraphGPT \cite{gao2024graph} views each node and edge as a “token,” training on 100 million molecules in a purely Transformer-based manner, whereas the Graph Transformer Foundation Model (GTFM) \cite{mizera2024graph} specializes in ADMET tasks. BioBridge \cite{wang2023biobridge}, meMeanwhile, employs a Transformer to align knowledge-graph triplets $(v_i, e_{ij}, v_j)$ across multiple biomedical modalities without relying on GNNs.

\noindent\textbf{Large Language Models.} Another direction bypasses graph encoding altogether by treating chemical or biological strings as input to large language models. Formally, given textual descriptions $\mathbf{d}_{\mathcal{G}}$, one example \cite{qian2023can} investigates whether an LLM alone, represented as $\text{LLM}(\mathbf{d}_{\mathcal{G}})$, can handle molecular property prediction in zero- or few-shot settings. MolecularGPT \cite{liu2024moleculargpt} enriches SMILES prompts $\mathbf{d}_v$ with structural “neighbor” demonstrations $\mathbf{d}_{\mathcal{N}_v}$ to guide predictions. Beyond small molecules, ESMFold \cite{lin2022language} leverages an LLM (ESM-2) trained on protein sequences $\mathbf{d}_{\text{protein}}$ to predict 3D protein structures with atomic-level resolution. Likewise, GP-GPT \cite{lyu2024gp} employs a Llama-based model $\text{LLM}(\mathbf{d}_{\text{gene}})$ to map genomic sequences to phenotypes, converting genomic knowledge into textual prompts. For cross-modal retrieval scenarios, Text2Mol \cite{edwards2021text2mol} embeds textual queries $\mathbf{d}_{\text{query}}$ and chemical representations (e.g., fingerprints $\mathbf{x}_{\text{fp}}$) into a shared embedding space. Collectively, these sequence-based Transformers and LLMs illustrate the versatility of self-attention mechanisms for capturing chemical and biological patterns, even without explicit graph message passing.

\noindent\textbf{GNNs as Auxiliary Models.} Some works pass GNN-derived embeddings $\mathbf{Z}=\text{GNN}(\mathbf{X}, \mathbf{A})$ to a Q-Former or similar projection module, producing discrete “molecule tokens” $\mathbf{z}_{\text{token}}$ processed jointly with textual data $\mathbf{d}_{\mathcal{G}}$. MolCA \cite{liu2023molca} and UniMoT \cite{zhang2024unimot} illustrate this principle by injecting graph-encoded structural information into frozen LLMs for tasks such as molecule-to-text generation, retrieval, or captioning. GIT-Mol \cite{liu2024git} expands this concept to a three-way multimodal setting $(\mathbf{Z}, \mathbf{d}_{\mathcal{G}}, \mathbf{x}_{\text{img}})$ involving graphs, images, and text. Similarly, GIMLET \cite{zhao2023gimlet} and the Molecular Multimodal Foundation Model \cite{su2022molecular} incorporate GNN-derived features $\mathbf{Z}$ into instruction-based or attention-based frameworks alongside textual prompts. InstructMol \cite{cao2023instructmol} leverages molecule–text contrastive pre-training to align graph encoders $\text{GNN}(\cdot)$ with language models, while a multi-modal structure–text approach \cite{liu2023multi} employs contrastive objectives for text-based molecule retrieval and editing.

\subsubsection{Hybrid Approaches}

Although graph models excel at capturing local connectivity and structural nuances, and language models capture global or semantic nuances (especially when text is involved), neither alone may suffice for complex, real-world applications. Hybrid models integrate structural inductive biases from graph models, denoted as $\text{GNN}(\mathbf{X}, \mathbf{A})$, and language-based reasoning from Transformers or LLMs, represented as $\text{LLM}(\mathbf{d}_{\mathcal{G}})$, enabling more comprehensive representations of molecules for diverse tasks.

\noindent\textbf{Task Specialization.} Certain hybrid models address broader or more specialized tasks explicitly. ReLM \cite{shi2023relm} formulates reaction prediction as first generating candidate products using GNN outputs $\mathbf{Z}_{\text{candidates}}$, which are subsequently ranked by an LLM conditioned on reaction conditions $\mathbf{d}_{\text{reaction}}$. CLAMP \cite{seidl2023enhancing} aligns molecular encoders $\mathbf{Z}_{\text{mol}}$ with text-based bioassay descriptions $\mathbf{d}_{\text{bioassay}}$ for zero-shot discovery tasks, and MolGPS \cite{sypetkowski2025scalability} merges message-passing neural networks (MPNN) with Transformer modules to enhance scalability on large supervised datasets. DPA-2 \cite{zhang2024dpa} integrates symmetry-preserving GNN layers and Transformer-based attention for multi-task molecular simulations, whereas MolFM \cite{luo2023molfm} unifies graph embeddings, textual Transformer embeddings, and knowledge-graph features $\mathbf{Z}_{\text{KG}}$ for cross-modal retrieval. Additional specialized hybrid designs include GALLON \cite{xu2024llm}, which distills representations from both GNN and LLM into a single multilayer perceptron (MLP) $\hat{y} = \text{MLP}(\mathbf{z}_{\text{GNN}} \| \mathbf{z}_{\text{LLM}})$ for molecular property prediction, and Hybrid-LLM-GNN \cite{li2025hybrid}, which integrates crystallographic embeddings $\mathbf{Z}_{\text{crystal}}$ with language-based features for materials property inference. Together, these hybrid approaches demonstrate the substantial advantages gained from combining graph-level inductive biases with language-based models, resulting in robust performance in tasks such as molecular property prediction, reaction modeling, and text-guided molecular manipulation.

\subsubsection{Future Directions}
Future directions for Graph Foundation Models (GFMs) in molecular graphs center around several promising research trends. Firstly, developing equivariant GNN architectures that explicitly encode symmetries such as rotation ($R$) and translation ($T$) is crucial for accurately modeling molecular properties. Leveraging large-scale self-supervised pre-training on massive unlabeled molecular datasets could significantly enhance generalization and predictive robustness, particularly when handling novel molecular scaffolds. Physics-informed neural networks that integrate quantum-chemical principles and physical laws represent another crucial avenue, enabling more faithful modeling of chemical interactions. Finally, multi-modal retrieval-augmented strategies that combine molecular graph data with textual and structural contexts are essential for improving interpretability and reliability in high-stakes applications, such as drug discovery and materials design.

\subsection{Algorithmic Graphs}

\begin{table*}[!t]
    \centering
    \rowcolors{2}{shadecolor}{white}
    \caption{Summary of domain-specific GFMs on computational graphs.}
    \label{tab:domain gfm computational}
    \resizebox{\linewidth}{!}{
        \begin{tabular}{llllp{3.5cm}p{3.5cm}p{3.5cm}p{3.5cm}l}
            \toprule
            \textbf{Method Name}                                      & \textbf{Domain}     & \textbf{Backbone} & \textbf{Pretrain} & \textbf{Adaptation}  & \textbf{Feature Align} & \textbf{Structure Align} & \textbf{Task Align} & \textbf{Github}                                                                \\
            \midrule
            \textbf{Triplet-GMPNN} \cite{ibarz2022generalist}         & Computational Graph & GNN               & Supervised        & Finetune             & Data - Node Property   & Loss - Multi-task        & N/A                 & -                                                                              \\
            \textbf{GraphForge} \cite{wang2024graphtool}              & Computational Graph & LLM               & Generative        & Finetune             & Data - Node Property   & Model - Retriever        & Explicit - QA       & \href{https://anonymous.4open.science/r/GraphTool-Instruction/README.md}{Link} \\
            \textbf{HLM-G} \cite{khurana2024hierarchical}             & Computational Graph & LLM               & Generative        & Finetune, In-context & Data - Node Property   & Data - Augment           & N/A                 & -                                                                              \\
            \textbf{PathCompare} \cite{agrawal2024can}                & Computational Graph & LLM               & Generative        & In-context           & Data - Node Property   & N/A                      & Explicit - QA       & \href{https://github.com/PalaashAgrawal/LLMGraphTraversal}{Link}               \\
            \textbf{Hyper-BAG and Hyper-COT} \cite{feng2024beyond}    & Computational Graph & LLM               & Generative        & In-context           & Data - Node Property   & N/A                      & Explicit - QA       & \href{https://github.com/iMoonLab/LLM4Hypergraph}{Link}                        \\
            \textbf{Graph Linearization} \cite{xypolopoulos2024graph} & Computational Graph & LLM               & Generative        & In-context           & Data - Node Property   & Model - Retriever        & Explicit - QA       & -                                                                              \\
            \textbf{TP} \cite{yu2023thought}                          & Computational Graph & LLM               & Generative        & In-context           & Data - Text Attribute  & N/A                      & Explicit - QA       & \href{https://github.com/Samyu0304/thought-propagation}{Link}                  \\
            \textbf{GPT4Graph} \cite{guo2023gpt4graph}                & Computational Graph & LLM               & Generative        & In-context           & Data - Node Property   & Data - Augment           & Explicit - QA       & -                                                                              \\
            \textbf{GraphAgent-Reasoner} \cite{hu2024scalable}        & Computational Graph & LLM               & Generative        & In-context           & Data - Node Property   & N/A                      & Explicit - QA       & -                                                                              \\
            \textbf{GraphTeam} \cite{li2024graphteam}                 & Computational Graph & LLM               & Generative        & In-context           & Data - Node Property   & Model - Retriever        & Explicit - QA       & -                                                                              \\
            \textbf{GUNDAM} \cite{ouyang2024gundam}                   & Computational Graph & LLM               & Generative        & Finetune, In-context & Data - Node Property   & Model - Prompt Learning  & Explicit - QA       & -                                                                              \\
            \textbf{PIE} \cite{gong2025pseudocode}                    & Computational Graph & LLM               & Generative        & In-context           & Data - Node Property   & N/A                      & Explicit - QA       & -                                                                              \\
            \textbf{GraphInstruct} \cite{luo2024graphinstruct}        & Computational Graph & LLM               & Generative        & In-context           & Data - Node Property   & Data - Augment           & Explicit - QA       & \href{https://github.com/CGCL-codes/GraphInstruct}{Link}                       \\
            \textbf{GCoder} \cite{zhang2024gcoder}                    & Computational Graph & LLM               & Generative        & Finetune, In-context & Data - Node Property   & Loss - Auxiliary         & Explicit - QA       & \href{https://github.com/Bklight999/GCoder}{Link}                              \\
            \textbf{GraphWiz} \cite{chen2024graphwiz}                 & Computational Graph & LLM               & Generative        & Distillation         & Data - Node Property   & Data - Augment           & Explicit - QA       & \href{https://github.com/nuochenpku/Graph-Reasoning-LLM}{Link}                 \\
            \textbf{GraphLLM} \cite{chai2023graphllm}                 & Computational Graph & GNN + LLM         & Generative        & Finetune, In-context & Data - Node Property   & Model - Retriever        & Explicit - QA       & \href{https://github.com/mistyreed63849/Graph-LLM}{Link}                       \\
            \textbf{GraphToken} \cite{perozzi2024let}                 & Computational Graph & GNN + LLM         & Generative        & In-context           & Data - Node Property   & Model - Retriever        & Explicit - QA       & -                                                                              \\
            \bottomrule
        \end{tabular}
    }
\end{table*}

Graphs are fundamental data structures for representing algorithmic processes, combinatorial structures, and mathematical relationships. In this context, we use the term \textit{algorithmic graphs} to refer to graph-structured inputs that encode procedural tasks—such as shortest path finding, satisfiability checking, sorting, and symbolic mathematics—that require reasoning over discrete structures.

Unlike conventional graph learning tasks focused on semantic inference (e.g., classification or recommendation), algorithmic graph problems demand models that can emulate or generalize classical algorithmic behaviors. In this section, we review GFMs designed for algorithmic reasoning, highlighting their ability to capture structural invariants, support multi-step reasoning, and generalize across instances of varying size and complexity. We further summarize representative benchmarks, learning paradigms, and model architectures that drive progress in this emerging direction.

\subsubsection{Structured Graph Reasoning}
\noindent \textbf{Task-Oriented Approaches.} Early works in graph math reasoning focused on building generalist neural solvers capable of processing multiple algorithmic tasks within a unified framework. Triplet-GMPNN~\cite{ibarz2022generalist} introduced a GNN-based GFM designed to solve problems like shortest paths, sorting, and dynamic programming, formalized as learning a mapping \(\Phi: (\mathcal{G}, t) \mapsto y\), where \(t\) denotes the task type and \(y\) the target solution. This multi-task formulation highlights the importance of task-conditioned graph message passing over adjacency matrix \(\mathbf{A}\) and features \(\mathbf{X}\). GCoder~\cite{zhang2024gcoder} frames graph reasoning as structured program generation, learning to synthesize code \(c = \Psi(\mathcal{G})\) directly from input graphs. By incorporating reinforcement learning with compiler feedback, GCoder improves program correctness and execution efficiency. This highlights how graph math GFMs can bridge declarative graph representations with executable procedural knowledge. GraphPatternBench~\cite{dai2024large} evaluates pattern recognition tasks over graph \(\mathcal{G}\), where models predict structural motifs (cliques, cycles) by learning a binary classification function \(f: \mathcal{V} \mapsto \{0,1\}\).

\noindent \textbf{Instruction-Tuned Approaches.} Instruction-tuned reasoning is introduced by GraphInstruct~\cite{luo2024graphinstruct}, where graph tasks are specified via instruction \(I\), and the GFM infers \(y = \Theta(\mathcal{G}, I)\). GraphWiz~\cite{chen2024graphwiz} enhances this process by requiring the GFM to generate interpretable step-by-step reasoning traces \([y_1, y_2, \dots, y_T]\), aligning intermediate solutions to the graph's evolving state. MAGMA~\cite{taylor2024large} further evaluates classical algorithms (e.g. BFS, DFS, Dijkstra) using similar intermediate reasoning traces across \(\{\mathcal{G}_t\}\). Structured graph representation is tackled by GraphLLM~\cite{chai2023graphllm}, which integrates learned graph embeddings \(\mathbf{z}_i\) into pre-trained language models via graph-to-text serialization. GraphToken~\cite{perozzi2024let} instead learns structural embeddings directly \(\mathbf{z}_i = \phi(v_i, \mathcal{N}(v_i), \mathbf{A})\), enabling the GFM to leverage topological context natively during reasoning.

\subsubsection{Benchmarking and Multi-Agent Collaboration}
Comprehensive benchmarking highlights strengths and limitations of current graph math GFMs. GraphArena~\cite{tang2025grapharena} evaluates both polynomial-time and NP-complete tasks, tracking correctness and hallucination:
\(\ell = \mathrm{Eval}(y_{\text{pred}}, y_{\text{true}})\). Additionally,
NLGraph~\cite{wang2023can} expands this to natural language queries over graphs, requiring models to parse task text into formal graph queries \(q = \xi(d)\) before solving them. GPT4Graph~\cite{guo2023gpt4graph} focuses on semantic and structural reasoning in graphs, including centrality estimation and graph classification. This benchmark highlights the gap between pre-trained LLMs and structured graph-aware GFMs. Multi-agent collaboration offers another promising direction for complex graph reasoning. GraphTeam~\cite{li2024graphteam} employs multiple specialized LLM agents, each responsible for parsing, retrieval, coding, or reasoning. Each agent computes \(\mathbf{z}_i^{(k)} = \phi_k(\mathcal{G})\)
with intermediate results shared across agents for iterative refinement. GraphTool-Instruction~\cite{wang2024graphtool} extends this by explicitly incorporating external tool calls, meaning that:
\(y = \mathrm{Tool}(q)\)
where \(q\) is the subtask-specific query extracted by the model. GraphAgent-Reasoner~\cite{hu2024scalable} applies a similar agent decomposition but further breaks graph problems into node-centric subtasks, with individual agents solving \(y_i = \phi(\mathcal{N}(v_i), \mathbf{x}_i)\), with final aggregation yielding global predictions.

\subsubsection{Encoding Strategies and Algorithmic Refinement}
\noindent \textbf{Structure-Aware Encoding Approaches.} Encoding graph structures effectively for LLM reasoning is a foundational challenge. Structured JSON encoding~\cite{zhu2024investigating} serializes each node’s neighborhood into hierarchical JSON trees, preserving adjacency information \(\mathcal{N}(v_i)\) directly within the prompt. Graph Linearization~\cite{xypolopoulos2024graph} encodes \(\mathcal{G}\) into a linear sequence of tokens ordered by centrality ranking:
\(\pi = \mathrm{Order}(\mathcal{V}; \mathrm{Centrality})\).
This preserves relational salience during sequence processing. PIE~\cite{gong2025pseudocode} structures graph reasoning into staged reasoning, decomposing into:
\[
    y = \Theta_{\text{execute}}(\Theta_{\text{design}}(\Theta_{\text{understand}}(\mathcal{G}))),
\]
where each component produces structured pseudocode to guide the next stage. LogDepth Transformer Hierarchy~\cite{sanford2025understanding} theoretically demonstrates that logarithmic-depth transformers (depth \(\log N\)) efficiently capture long-range graph dependencies, outperforming standard transformers and GNNs on certain reasoning problems. This provides architectural guidance for future GFMs.

\noindent \textbf{Extended Reasoning Approaches.} LLM4Hypergraph~\cite{feng2024beyond} extends graph reasoning to hypergraphs \(\mathcal{H} = (\mathcal{V}, \mathcal{E})\), where hyperedges connect arbitrary subsets of nodes. The model applies specialized prompting strategies (Hyper-BAG, Hyper-COT) to encode these higher-order relations into:
\[
    \mathbf{z}_e = \phi(e, \{v_i : v_i \in e\}).
\]
NLGIFT~\cite{zhang2024can} measures out-of-distribution generalization by introducing diverse structural shifts (varying degree distributions, node attributes, and edge sparsity), requiring GFMs to adapt dynamically. Traversal remains central to graph math reasoning. PathCompare~\cite{agrawal2024can} enhances traversal reasoning by prompting models to compare candidate paths \(c = \mathrm{Compare}(p_1, p_2)\), where paths are sequences of edges \((v_i, v_j) \in \mathcal{E}\). TREETOP~\cite{arora2024treetop} adapts these techniques to conversation trees, where nodes represent conversational turns and edges encode reply relations.

\subsubsection{Future Directions}
Future graph math GFMs will integrate hybrid retrieval-augmented reasoning, dynamically fetching relevant subgraphs during inference to improve local-global context alignment. Inspired by Thought Propagation~\cite{yu2023thought}, future models may recursively decompose graph problems into subproblems \(y = \sum_{i} \theta_i(y_i)\), where \(y_i\) are solutions to extracted subgraphs. Combining structured retrieval, multi-agent collaboration, and external tool use will enhance adaptability and robustness across real-world tasks. Benchmarks like GraphPatternBench~\cite{dai2024large}, NLGraph~\cite{wang2023can}, and GraphArena~\cite{tang2025grapharena} underscore the need for realistic datasets spanning scientific, social, and biological domains. Tool-augmented reasoning frameworks (GraphTool-Instruction~\cite{wang2024graphtool}, PIE~\cite{gong2025pseudocode}) point toward hybrid GFMs that blend learned reasoning with structured algorithmic calls, laying the foundation for the next generation of graph math reasoning models.

\subsection{Document Network}

\begin{table*}[!t]
    \centering
    \rowcolors{2}{shadecolor}{white}
    \caption{Summary of domain-specific GFMs on document graphs. }
    \label{tab:domain gfm document}
    \resizebox{\linewidth}{!}{
        \begin{tabular}{llllp{3.5cm}p{3.5cm}p{3.5cm}p{3.5cm}l}
            \toprule
            \textbf{Method Name}                                      & \textbf{Domain}           & \textbf{Backbone} & \textbf{Pretrain} & \textbf{Adaptation}              & \textbf{Feature Align} & \textbf{Structure Align}            & \textbf{Task Align}  & \textbf{Github}                                                                \\
            \midrule
            \textbf{METAG} \cite{jin2023learning}                     & Document Graph            & LLM               & Hybrid            & In-context, Test-time Adaptation & Data - Text Attribute  & Model - Retriever                   & Explicit - QA        & \href{https://github.com/PeterGriffinJin/METAG}{Link}                          \\ 
            \textbf{TAPE} \cite{he2023harnessing}                     & Document Graph            & GNN + LLM         & Supervised        & Finetune, In-context             & Data - Text Attribute  & N/A                                 & N/A                  & \href{https://github.com/XiaoxinHe/TAPE}{Link}                                 \\ 
            \textbf{LLM4GraphTopology} \cite{sun2023large}            & Document Graph            & GNN + LLM         & Supervised        & Finetune, In-context             & Data - Text Attribute  & Data - Augment                      & N/A                  & \href{https://github.com/sunshy-1/LLM4GraphTopology}{Link}                     \\ 
            \textbf{G-Prompt} \cite{huang2023prompt}                  & Document Graph            & GNN + LLM         & Supervised        & In-context                       & Data - Text Attribute  & Model - Retriever                   & N/A                  & -                                                                              \\ 
            \textbf{ConGraT} \cite{brannon2023congrat}                & Document Graph            & GNN + LLM         & Contrastive       & Finetune                         & Data - Text Attribute  & Loss - Pretrain                     & N/A                  & \href{https://github.com/wwbrannon/congrat}{Link}                              \\ 
            \textbf{GLEM} \cite{zhao2022learning}                     & Document Graph            & GNN + LLM         & Hybrid            & Distillation, Finetune           & Data - Text Attribute  & Loss - Pretrain                     & N/A                  & \href{https://github.com/AndyJZhao/GLEM}{Link}                                 \\ 
            \textbf{PATTON} \cite{jin2023patton}                      & Document Graph            & GNN + LLM         & Hybrid            & Finetune                         & Data - Text Attribute  & Loss - Pretrain                     & N/A                  & \href{https://github.com/PeterGriffinJin/Patton}{Link}                         \\
            \bottomrule
        \end{tabular}
    }
\end{table*}

Semantic document graphs represent networks where nodes \(v_i \in \mathcal{V}\) are textual entities (e.g., documents or records) and edges \(e_{ij} \in \mathcal{E}\) capture semantic links such as citations or topical similarity. Each node and edge may carry features \(\mathbf{x}_i, \mathbf{e}_{ij} \in \mathbb{R}^D\), forming attribute matrices \(\mathbf{X} \in \mathbb{R}^{N \times D}\) and \(\mathbf{E} \in \mathbb{R}^{M \times D}\), with structure defined by adjacency \(\mathbf{A} \in \{0,1\}^{N \times N}\). Semantic document GFMs aim to jointly model textual semantics and graph structure, but face key challenges in content-structure alignment and cross-domain scalability. While LLMs encode rich text features \(\mathbf{z}_i^{\mathrm{text}}\) and GNNs propagate structural signals \(\mathbf{z}_i^{\mathrm{graph}}\), aligning these representations is difficult in heterogeneous document graphs with diverse relations and evolving content.

\noindent \textbf{Prompt-based Integration Approaches.}
To enhance multi-relation representation learning, METAG~\cite{jin2023learning} proposes a multiplex embedding framework where a language model encoder generates node embeddings \(\mathbf{z}_i\) by dynamically injecting relation-specific prior tokens from \(\mathcal{T}\). This results in relation-aware embeddings that adapt to different edge types in \(\mathbf{E}\), preserving parameter efficiency while enriching structural semantics. G-Prompt~\cite{huang2023prompt} extends task-specific adaptation by introducing a graph adapter-enhanced prompting strategy for TAGs. Given a graph \(\mathcal{G}\) with adjacency \(\mathbf{A}\), it injects task-aware prompt embeddings into \(\mathrm{LLM}(\cdot)\), where local neighborhood features from \(\mathcal{N}_v\) are fused into the prompt. This hybrid design retains both global semantics and local structure, improving few-shot and zero-shot node representation learning.

\noindent \textbf{Multi-Objective Optimization Approaches.}
Beyond pre-training, PATTON~\cite{jin2023patton} jointly optimizes two objectives: network-contextualized masked language modeling (MLM) and masked node prediction. The first reconstructs masked tokens in \(\mathbf{x}_i\) using neighborhood context \(\mathcal{N}_v\), while the second predicts node identities based on textual embeddings. This dual-task setup aligns local semantic recovery with global graph reasoning, producing node embeddings \(\mathbf{z}_i\) that capture both token-level and topological information. ConGraT~\cite{brannon2023congrat} adopts a dual-encoder design, using a GNN and a pre-trained language model to generate \(\mathbf{z}_i^{\mathrm{graph}}\) and \(\mathbf{z}_i^{\mathrm{text}}\), respectively. A cross-modal contrastive loss aligns these views by maximizing mutual information, enhancing cross-domain generalization in semantic document GFMs.

\noindent \textbf{Hybrid Training Approaches.}
LLM4GraphTopology~\cite{sun2023large} refines graph topology by prompting LLMs to assess semantic similarity between node pairs \((v_i, v_j)\), adjusting edges in \(\mathbf{A}\) based on similarity thresholds. This process is reinforced by pseudo-label propagation, where LLM-generated labels diffuse through the updated graph to enhance classification and clustering. To support scalable learning, GLEM~\cite{zhao2022learning} introduces a variational EM framework that alternates between GNN-based structural reasoning (E-step) and LLM-based semantic encoding (M-step). Node embeddings \(\mathbf{z}_i\) are iteratively updated by propagating through \(\mathrm{GNN}(\mathbf{A}, \mathbf{X})\) and conditioning language representations on aggregated graph signals. TAPE~\cite{he2023harnessing} improves interpretability by combining LLM-generated explanations \(d_{v_i}\) with node features \(\mathbf{x}_i\), forming enriched inputs \([\mathbf{x}_i \| d_{v_i}]\) for GNNs, leading to more accurate and interpretable node representations.

\noindent \textbf{Future Directions.} The evolution of semantic document graph foundation models reflects a steady progression from static graph encoders to dynamically adaptive graph-text co-modeling pipelines. Future directions include hierarchical prompting that integrates section-level and document-level contexts into \(\mathbf{z}_i\), contrastive augmentation that jointly aligns node text, edge descriptions \(d_{e_{ij}}\), and global metadata \(d_g\), and multi-lingual pre-training that generalizes GFMs across multilingual scientific corpora. These innovations will further position semantic document GFMs as central tools for scholarly discovery, legal document analysis, and large-scale retrieval across knowledge graphs.

\subsection{Heterogeneous Graph}

\begin{table*}[!t]
    \centering
    \rowcolors{2}{shadecolor}{white}
    \caption{Summary of domain-specific GFMs on heterogeneous graphs. }
    \label{tab:domain gfm hetero}
    \resizebox{\linewidth}{!}{
        \begin{tabular}{llllp{3.5cm}p{3.5cm}p{3.5cm}p{3.5cm}l}
            \toprule
            \textbf{Method Name}                             & \textbf{Domain}     & \textbf{Backbone} & \textbf{Pretrain} & \textbf{Adaptation}  & \textbf{Feature Align} & \textbf{Structure Align}        & \textbf{Task Align} & \textbf{Github}                                             \\
            \midrule
            \textbf{Heterformer} \cite{jin2023heterformer}   & Heterogeneous Graph & GNN               & Supervised        & Finetune             & Data - Text Attribute  & N/A                             & N/A                 & \href{https://github.com/PeterGriffinJin/Heterformer}{Link} \\
            \textbf{SELAR} \cite{hwang2020self}              & Heterogeneous Graph & GNN               & Supervised        & Finetune             & N/A                    & Loss - Auxiliary                & N/A                 & \href{https://github.com/mlvlab/SELAR}{Link}                \\
            \textbf{PT-HGNN} \cite{jiang2021pre}             & Heterogeneous Graph & GNN               & Contrastive       & Finetune             & N/A                    & Data - Augment, Loss - Pretrain & Explicit - Subgraph & -                                                           \\
            \textbf{HetGPT} \cite{ma2024hetgpt}              & Heterogeneous Graph & GNN               & Contrastive       & Graph Prompting      & N/A                    & Loss - Pretrain                 & Explicit - Link     & -                                                           \\
            \textbf{CrossHG-Meta} \cite{zhang2022few}        & Heterogeneous Graph & GNN               & Hybrid            & Test-time Adaptation & N/A                    & Data - Augment                  & N/A                 & -                                                           \\
            \textbf{HierPromptLM} \cite{zhu2025hierpromptlm} & Heterogeneous Graph & LLM               & Supervised        & Finetune             & Data - Text Attribute  & Loss - Multi-task               & N/A                 & \href{https://github.com/UCLA-DM/pyHGT/}{Link}              \\
            \textbf{THLM} \cite{zou2023pretraining}          & Heterogeneous Graph & GNN + LLM         & Supervised        & Finetune             & Data - Text Attribute  & Data - Augment, Loss - Pretrain & N/A                 & \href{https://github.com/Hope-Rita/THLM}{Link}              \\
            \textbf{GaLM} \cite{xie2023graph}                & Heterogeneous Graph & GNN + LLM         & Supervised        & Finetune             & Data - Text Attribute  & Loss - Auxiliary                & N/A                 & -                                                           \\
            \textbf{GHGRL} \cite{gao2024bootstrapping}       & Heterogeneous Graph & GNN + LLM         & Supervised        & Finetune             & Data - Text Attribute  & N/A                             & N/A                 & \href{https://github.com/zch65458525/GHGRL/tree/main}{Link} \\
            \textbf{HiGPT} \cite{tang2024higpt}              & Heterogeneous Graph & GNN + LLM         & Contrastive       & Finetune             & Data - Text Attribute  & N/A                             & Explicit - QA       & \href{https://github.com/HKUDS/HiGPT}{Link}                 \\
            \bottomrule
        \end{tabular}
    }
\end{table*}

A Heterogeneous Graph (HG)~\cite{wang2022survey} is denoted as $\mathcal{G} = (\mathcal{V}, \mathcal{E}, \mathcal{T}, \mathcal{R})$, comprising sets of nodes $\mathcal{V}$ and edges $\mathcal{E}$. $\mathcal{T}$ and $\mathcal{R}$ represent the types of nodes and edges, respectively, with the condition that $|\mathcal{T}| + |\mathcal{R}| > 2$. Additionally, $\tau(\cdot)$ and $\varphi(\cdot; \cdot)$ serve as mapping functions to identify the types of nodes and edges, where $\tau(i) \in \mathcal{T}$ for any $i \in \mathcal{V}$, and $\varphi(i, j) \in \mathcal{R}$ for any edge $(i,j) \in \mathcal{E}$. The central task in heterogeneous graph representation learning is to derive node embeddings that accurately reflect both structural and semantic contexts.

\noindent\textbf{Graph Model-based Approaches.} Early research efforts in heterogeneous graph foundation models predominantly leveraged self-supervised learning methodologies. SELAR \cite{hwang2020self} introduced a meta-learning approach to systematically balance auxiliary tasks, enhancing primary task performance through optimal node representations. Extending this approach, CrossHG-Meta \cite{zhang2022few} addressed the critical challenge of few-shot learning, mitigating data scarcity and emphasizing robust generalization across heterogeneous contexts. Concurrently, PT-HGNN \cite{jiang2021pre} applied contrastive learning at both node and semantic levels, significantly capturing nuanced structural information. Subsequently, HetGPT \cite{ma2024hetgpt} advanced these methodologies through graph prompting strategies, adapting pre-trained graph neural networks (GNNs) to diverse downstream tasks and thereby marking a shift toward more adaptive and flexible modeling paradigms.

\noindent\textbf{Language Model-based Approaches.} Inspired by recent advances in natural language processing, particularly the transformative successes of large language models (LLMs), researchers began integrating language encoders into heterogeneous graph modeling to address heterogeneity through textual embedding strategies. GaLM \cite{xie2023graph} modified masked language modeling by incorporating structural signals from adjacency matrices \( \mathbf{A} \in \{0,1\}^{N \times N} \) along with textual node information, enabling richer node representations. Further developments involved constructing graph-specific "tokens," analogous to language tokens, effectively encapsulating both structural and semantic attributes. For example, Higpt \cite{tang2024higpt} generated graph tokens from textual descriptions \( d_{v_i} \), processing them through a heterogeneous graph transformer (HGT) and refining the results via inference with large language models (LLMs). HierPromptLM \cite{zhu2025hierpromptlm} similarly extracted metapath-based subgraph tokens using LLMs to construct prompts tailored for downstream fine-tuning tasks. In parallel, methods like GHGRL \cite{gao2024bootstrapping} leveraged the reasoning capabilities of LLMs to discern node types from textual attributes, resulting in semantically enriched node embeddings. The creation of benchmarks such as the Heterogeneous Text Attributed Graph (HTAG) datasets further facilitates empirical evaluation and validation of these methodologies.

\noindent\textbf{Hybrid Approaches}. Most recently, hybrid models synthesizing GNNs and LLMs have achieved state-of-the-art performance by simultaneously capturing structural dependencies and semantic details. Heterformer \cite{jin2023heterformer}, for instance, integrated transformer-based neighbor aggregation with textual context, effectively unifying structural and semantic representations. Similarly, THLM \cite{zou2023pretraining} proposed a dual-encoder pre-training framework that jointly leverages GNN-based structural insights and LLM-derived semantic encodings, producing comprehensive multimodal node embeddings. These hybrid methodologies epitomize a natural evolution in graph modeling by harmonizing complementary strengths, thereby establishing foundations for robust and contextually rich graph models.

\textbf{Future Directions}. Future research in heterogeneous graph foundation models presents numerous promising directions. Key areas include exploring advanced multimodal integration strategies to enhance structural-semantic coherence, improving interpretability by explicitly aligning embeddings from distinct modalities, and developing dynamically adaptive models responsive to evolving graph structures. Moreover, research focusing on computational scalability and standardizing evaluation frameworks will be pivotal for advancing robust and contextually comprehensive graph foundation models.

\begin{table*}[!t]
    \centering
    \rowcolors{2}{shadecolor}{white}
    \caption{Summary of domain-specific GFMs on knowledge graphs, academic networks, temporal graphs, casual graphs. }
    \label{tab:domain gfm others}
    \resizebox{\linewidth}{!}{
        \begin{tabular}{llllp{3.5cm}p{3.5cm}p{3.5cm}p{3.5cm}l}
            \toprule
            \textbf{Method Name}                            & \textbf{Domain}  & \textbf{Backbone} & \textbf{Pretrain} & \textbf{Adaptation} & \textbf{Feature Align} & \textbf{Structure Align} & \textbf{Task Align} & \textbf{Github}                                          \\
            \midrule
            \textbf{MOTIF} \cite{huang2025expressive}       & Knowledge Graph  & GNN               & Supervised        & Finetune            & Data - Text Attribute  & Model - Codebook         & Explicit - Link     & -                                                        \\
            \textbf{UltraQuery} \cite{galkin2024foundation} & Knowledge Graph  & GNN               & Supervised        & Finetune            & Data - Text Attribute  & Model - Codebook         & Explicit - Link     & \href{https://github.com/DeepGraphLearning/ULTRA}{Link}  \\
            \textbf{ULTRA} \cite{galkin2023towards}         & Knowledge Graph  & GNN               & Supervised        & Finetune            & Data - Text Attribute  & Model - Codebook         & Explicit - Link     & \href{https://github.com/DeepGraphLearning/ULTRA}{Link}  \\
            \textbf{KG-ICL} \cite{cui2025prompt}            & Knowledge Graph  & GNN               & Supervised        & Graph Prompting     & Data - Text Attribute  & Model - Codebook         & Explicit - Link     & \href{https://github.com/nju-websoft/KG-ICL}{Link}       \\ \midrule
            \textbf{LitFM} \cite{zhang2024litfm}            & Academic Network & GNN               & Hybrid            & Finetune            & Data - Text Attribute  & Loss - Multi-task        & Explicit - QA       & -                                                        \\ \midrule
            \textbf{MiNT} \cite{shamsi2024mint}             & Temporal Graph   & GNN + LLM         & Supervised        & Finetune            & N/A                    & N/A                      & N/A                 & \href{https://anonymous.4open.science/r/MiNT}{Link}      \\ \midrule
            \textbf{ZCG} \cite{antonucci2023zero}           & Causal Graph     & LLM               & Hybrid            & In-context          & Data - Text Attribute  & N/A                      & Explicit - QA       & \href{https://github.com/IDSIA-papers/causal-llms}{Link} \\
            \bottomrule
        \end{tabular}
    }
\end{table*}

\subsection{Knowledge Graph}

Designing graph foundation models specifically for knowledge graphs (KGs) presents unique challenges, primarily due to the necessity of effectively inferring complex relationships and capturing implicit correlations among triplets (entity-relation-entity). Unlike general graphs, knowledge graphs require models to handle compositional generalization, logical consistency, and inductive inference, particularly for unseen entities and relations. Current research in graph foundation models for knowledge graphs predominantly addresses two key tasks: logical query answering and inductive link prediction.

\textbf{Logical Query Answering}. In logical query answering task, the objective is to accurately respond to structured queries involving multi-hop reasoning and logical operators (e.g., intersection, union, negation). Formally, given a knowledge graph \( G = (\mathcal{E}, \mathcal{R}, \mathcal{T}) \) with entities \( \mathcal{E} \), relations \( \mathcal{R} \), and triplets \( \mathcal{T} \subseteq \mathcal{E} \times \mathcal{R} \times \mathcal{E} \), models aim to generalize compositionally and dynamically aggregate reasoning signals across queries, including those involving unseen entities and relations. For instance, UltraQuery \cite{galkin2024foundation} addresses this by defining both projection and logical operators as vocabulary-independent functions, achieving zero-shot generalization through inductive link prediction. Similarly, KG-ICL \cite{cui2025prompt} utilizes prompt-based reasoning, dynamically constructing query-specific prompt graphs encoded by distinct message-passing networks, thus demonstrating robust generalization capabilities across various KGs. KICGPT \cite{wei2024kicgpt} further incorporates LLMs in knowledge graph completion, leveraging the textual understanding of LLMs to enhance the model capacity.

\textbf{Inductive Link Prediction}. Inductive link prediction task focuses on predicting missing relations or links without relying on learned entity embeddings. The task can be formulated as learning transferable relation representations \( r \in \mathcal{R} \) to infer missing links \((e_h, r, e_t)\), even when entities \( e_h, e_t \) are unseen during training. ULTRA \cite{galkin2023towards} addresses this by introducing a relation-based meta-graph structure, propagating information across relations rather than entities, thus enhancing transferability to novel knowledge graphs. Higher-order KGFM \cite{huang2025expressive} extends this by modeling multi-relation motifs to capture complex relational dependencies, significantly improving expressive power for inductive reasoning. 

\textbf{Future Directions}. Future research in graph foundation models for knowledge graphs should prioritize three main directions. Firstly, expanding beyond binary relations to handle complex, higher-order relational interactions such as temporal or n-ary relationships is critical. Secondly, integrating multi-modal data - including textual, visual, and numerical information - into KGs will enable comprehensive multi-modal reasoning capabilities. Lastly, improving model interpretability, controllability, and computational scalability remains essential, particularly for applications in sensitive domains such as healthcare and finance, where transparency and computational efficiency are paramount.

\subsection{Temporal Graph}

Temporal graphs, denoted as a sequence of evolving graph snapshots \(\{\mathcal{G}^t = (\mathcal{V}^t, \mathcal{E}^t)\}_{t=1}^T\), capture dynamic node interactions over time. Each node \(v_i \in \mathcal{V}^t\) is associated with a time-dependent feature vector \(\mathbf{x}_i^t \in \mathbb{R}^D\), while each edge \(e_{ij}^t \in \mathcal{E}^t\) carries an evolving feature \(\mathbf{e}_{ij}^t \in \mathbb{R}^D\). Modeling such structures requires simultaneously encoding spatial dependencies within each snapshot \(\mathcal{G}^t\) and capturing temporal dependencies across graph states over time. Existing temporal GNNs rely heavily on handcrafted diffusion mechanisms to propagate node features across timestamps, which often struggle to generalize to unseen temporal patterns, especially when nodes and edges are associated with rich text descriptions \(d_{v_i}^t\) and \(d_{e_{ij}}^t\) that evolve alongside the graph.

\noindent\textbf{Language Models-based Approaches}.
Recent works have explored the potential of large language models (LLMs) in reasoning over temporal graphs. LLM4DyG~\cite{zhang2024llm4dyg} represents a pioneering effort in benchmarking LLMs for spatial-temporal graph tasks, treating dynamic graphs as serialized sequences of adjacency matrices \(\{\mathbf{A}^t\}_{t=1}^T\) with evolving node descriptions \(\{d_{v_i}^t\}\). The study highlights that as graph size \(N\) and temporal density \(T\) increase, LLM performance declines, emphasizing the challenge of maintaining spatial-temporal consistency at scale. To address this, Disentangled Spatial-Temporal Thoughts (DST2) was introduced, explicitly separating spatial reasoning within \(\mathcal{G}^t\) from temporal reasoning across successive snapshots \(\mathcal{G}^{t-1} \to \mathcal{G}^t\), improving interpretability and prediction accuracy. A related line of work investigates how LLMs can model graph \textit{flow dynamics}, where node states \(y_i^t\) evolve based on local neighborhoods and historical attributes. FlowGPT~\cite{zhang2024flowgpt} introduces a benchmark assessing LLMs’ ability to capture diffusion processes such as SIR (Susceptible-Infected-Removed). By serializing dynamic graphs into time-ordered sequences of node states and adjacency matrices, FlowGPT evaluates how well LLMs trace propagation patterns and identify influential nodes. Together, these works highlight the challenges and potential of LLM-based reasoning in evolving temporal graphs.

\noindent\textbf{Transfer Learning Approaches}.
To enhance the adaptability of foundation models across diverse temporal graphs, MiNT~\cite{shamsi2024mint} proposes a multi-network pretraining strategy. Instead of training on a single dynamic graph, MiNT learns from a collection of networks \(\{\mathcal{G}^{(k)}\}_{k=1}^K\), each spanning its own time horizon. The approach uses a single encoder to generate temporally-aware node representations \(\mathbf{z}_i^{(k,t)} = \mathrm{GNN}(\mathbf{x}_i^{(k,t)}, \mathbf{A}^{(k,t)})\) that generalize to unseen graphs. By leveraging structural and temporal diversity, MiNT outperforms traditional temporal GNNs trained on individual datasets, demonstrating the promise of cross-network pretraining for temporal graph foundation models.

\noindent\textbf{Future Directions}. These advances highlight the unique challenges of temporal graph reasoning, including disentangled spatial-temporal modeling, cross-graph generalization, and flow-aware sequence modeling. Future work may explore hierarchical temporal abstraction to enrich node representations \(\mathbf{z}_i^t\) with multi-scale embeddings, enabling reasoning over both short- and long-term patterns. Adaptive serialization that adjusts input granularity based on graph density or event frequency could enhance LLM adaptability. Lastly, hybrid architectures combining GNN-based spatial encoding with LLM-based temporal reasoning show promise for building expressive, interpretable temporal graph foundation models.

\subsection{Academic Network}

Academic citation graphs \(\mathcal{G} = (\mathcal{V}, \mathcal{E})\) encode structural dependencies between research papers, where nodes \(v_i \in \mathcal{V}\) represent papers with metadata and text \(\mathbf{x}_i \in \mathbb{R}^D\), and edges \(e_{ij} \in \mathcal{E}\) capture citation links. These graphs require modeling both semantic relevance and citation dynamics while balancing local consistency with global citation flow. Unlike static retrieval, citation graphs reflect evolving scientific discourse, demanding embeddings \(\mathbf{z}_i\) that integrate node content, citation context, and temporal relevance. This necessitates dynamically contextualized retrieval, where citation edges complement textual similarity.

\noindent \textbf{Graph-Augmented Retrieval and Generation.} LitFM~\cite{zhang2024litfm} pioneers this space by proposing the first literature foundation model explicitly designed to integrate academic citation graphs into LLM workflows. The framework uses a graph retriever that retrieves structurally relevant papers based on graph proximity in adjacency matrix \(\mathbf{A}\) and citation-aware embeddings \(\mathbf{z}_i\), mitigating common LLM failures such as citation hallucination and knowledge incompleteness. Beyond simple graph-enhanced retrieval, LitFM also employs instruction tuning over domain-specific citation graphs, enabling the model to generalize across citation prediction, related work generation, and literature review summarization tasks. The development of LitFM highlights the potential of academic graph foundation models to reshape how scientific literature is processed, summarized, and cited within LLM ecosystems.

\noindent \textbf{Future Directions.} Future directions could focus on enriching paper embeddings \(\mathbf{z}_i\) by incorporating temporal citation patterns and multi-modal signals such as figures, tables, and equation graphs. Additionally, expanding instruction tuning to include task compositions could enable more complex scholarly reasoning tasks. As citation graphs continue to grow in scale and complexity, the fusion of LLM semantic capabilities with structural insights from citation graphs will remain central to building robust academic graph foundation models.

\subsection{Causal Graph}

Causal graphs, represented as directed acyclic graphs (DAGs) \(\mathcal{G} = (\mathcal{V}, \mathcal{E})\), encode cause-effect relationships, where nodes \(v_i\) denote variables and directed edges \(e_{ij}\) capture causal links. Node features \(\mathbf{x}_i \in \mathbb{R}^D\) may include contextual data like text or metadata, forming hybrid semantic-causal structures. Designing causal GFMs poses unique challenges: the graphs are sparse and directional, requiring inference of asymmetric dependencies, and often originate from noisy, text-heavy data. This demands joint reasoning over textual evidence and graph structure, making semantic grounding and structural consistency critical but difficult to achieve.

\noindent \textbf{Causal Reasoning with Language Models.}
To address causal graph reasoning, Causal-LLM~\cite{antonucci2023zero} introduces a zero-shot approach that constructs pairwise queries like “Does \(v_i\) cause \(v_j\)?” from unstructured text, iteratively building a global causal graph \(\mathcal{G}\). Without explicit supervision, it leverages the causal reasoning abilities of LLMs for scalable discovery, though it struggles with indirect chains (e.g., \(v_i \to v_k \to v_j\)), exposing limitations in prompt design. Complementing this, CLEAR~\cite{chen2024clear} proposes a benchmark to evaluate LLMs on twenty causal tasks, including D-separation, backdoor adjustment, and effect estimation. Models are tested on structural and textual views of \(\mathcal{G}\), revealing performance drops on graphs with high-degree nodes or long paths. CLEAR also shows that phrasing and context order significantly affect results, highlighting the fragility of current prompting methods in causal reasoning.

\noindent \textbf{Future Directions.} The emergence of causal GFMs presents a promising path toward scalable, interpretable causal discovery across diverse scientific fields. Future research may explore structured causal prompting where queries incorporate domain-specific causal priors such as preferred causal ordering (e.g., temporal precedence). Another promising direction is to augment node features \(\mathbf{x}_i\) with confidence-aware embeddings derived from multiple noisy sources, enabling the model to express uncertainty-aware causal graphs. Moreover, integrating causal graph learning into multi-modal GFMs, where text, tables, and graphs jointly contribute to causal inference, could significantly enhance causal discovery in data-rich scientific domains. These innovations will further solidify causal graph foundation models as essential tools for automated scientific reasoning and knowledge discovery.

\newpage

\section{Theoretical Understandings}
\label{sec:theory}

\subsection{Emergence and Scaling Law}

Foundation models exhibit emergence \cite{wei2022emergent}, where increasing model size, data availability, and total compute leads to significant improvements in performance. These phenomena are characterized through neural scaling laws \cite{kaplan2020scaling,snell2024scaling}, which provide quantitative insights into model behavior under resource expansion. While scaling laws have been extensively studied in LLMs, replicating these observations in graph-based models remains an ongoing challenge.

\subsubsection{Well-Structured Graphs}

Current studies on graph scaling laws primarily focus on well-structured graph domains, such as molecular and atomic systems \cite{frey2023neural, chen2023uncovering, shoghi2023molecules, zhang2024dpa, batatia2023foundation}. In these settings, graphs exhibit naturally predefined structures where individual components carry intrinsic semantics, akin to the structured nature of language with its grammatical rules. This structural consistency facilitates the emergence of scaling properties in graph learning.

Empirical evidence supporting the existence of scaling laws in domain-specific graphs has emerged from molecular and atomic modeling. For instance, JMP \cite{shoghi2023molecules} demonstrated improvements in atomic property prediction across diverse chemical domains by pretraining on large-scale datasets. The pretraining corpus comprised approximately 120 million systems from four distinct sources: OC20 (100M examples), OC22 (8M examples), ANI-1x (2M examples), and Transition-1x (10M examples). These datasets included both equilibrium and non-equilibrium atomic structures, with energy and force labels serving as primary supervision signals. Notably, this work illustrated that a pretrained model could acquire transferable knowledge, requiring minimal labeled data for downstream adaptation. Following a similar paradigm, DPA-2 \cite{zhang2024dpa} leveraged large-scale multi-task pretraining over 10 million atomic structures, facilitating knowledge transfer to unseen tasks. These studies \cite{shoghi2023molecules, zhang2024dpa, batatia2023foundation} reinforce the understanding that scaling graph models with natural structures enhances performance, mirroring trends observed in LLMs and vision models.

To further quantify emergence in molecular graph, existing studies have sought to determine the critical thresholds of data and model size required to achieve scaling laws \cite{frey2023neural, chen2023uncovering}. Frey et al. \cite{frey2023neural} investigated neural scaling behaviors in large-scale graph-based chemical models, identifying key scaling exponents that capture performance improvements as model capacity and dataset size increase. Their findings revealed a scaling exponent of 0.17 for the largest dataset and 0.26 for equivariant GNN-based interatomic potential models, indicating measurable gains in pretraining loss with increased resources. Similarly, Chen et al. \cite{chen2023uncovering} conducted an extensive analysis of scaling laws in molecular graphs, examining the impact of data modality, dataset partitioning strategies, pretraining paradigms, and model capacity constraints. Their key insights include: \textit{Modality Dependence:} Different molecular representations exhibit distinct scaling behaviors. Graph-based and fingerprint-based encodings demonstrate the highest data efficiency, whereas SMILES-based representations exhibit diminished performance improvements as dataset size increases. \textit{Pretraining Efficacy:} Pretraining provides significant benefits in low-data regimes but exhibits diminishing returns in high-data scenarios, where negative transfer effects may arise. \textit{Dataset Partitioning:} The efficiency of data utilization varies based on partitioning strategies. Random splits yield the highest efficiency, while scaffold-based and imbalanced splits introduce distribution shifts that lower learning effectiveness. \textit{Model Capacity Trade-offs:} No straightforward relationship exists between dataset size and optimal model capacity. In some cases, smaller datasets necessitate larger models to achieve peak performance.

Beyond molecular graphs, scaling laws have also been explored in temporal graphs. MiNT \cite{shamsi2024mint} analyzed scaling behaviors in dynamic transaction networks, leveraging a collection of 84 temporal graphs. By pretraining on 64 networks and evaluating transferability on 20 unseen networks, MiNT achieved superior zero-shot performance, surpassing models trained on individual datasets. Crucially, their study demonstrated a consistent improvement in performance with increasing numbers of training networks.

\subsubsection{General Graphs}

While scaling laws have been studied in structured domains, such as molecular and atomic graphs, their applicability to general-purpose GFMs remains largely unexplored. Recent efforts have attempted to investigate scaling behavior from both supervised learning \cite{liu2024towards} and self-supervised learning \cite{ma2024neural} perspectives. Under supervised learning settings, Liu et al. \cite{liu2024towards} analyzed scaling behavior in GFMs trained on up to 100 million parameters and 50 million samples. Their findings indicate that model depth plays a pivotal role in scaling performance. Furthermore, they observed that traditional measures of data volume, such as the number of graph instances, are ineffective due to the irregular size of individual graphs. Instead, they proposed using the number of nodes or edges as a more reliable metric for defining scaling laws in graph data. From a self-supervised learning perspective, Ma et al. \cite{ma2024neural} examined whether existing graph SSL techniques exhibit consistent neural scaling behavior. Their analysis revealed that while SSL loss continues to improve with increasing data and model sizes, downstream performance remains highly sensitive to architectural choices and pretext task design. Unlike in other domains, where larger datasets and models consistently yield better performance, graph SSL methods do not exhibit clear scaling trends. Consequently, they argue that existing graph SSL frameworks may not yet be suitable for training scalable GFMs. In addition, several studies have proposed advanced models \cite{xia2024anygraph, wang2024learning} and evaluated their scaling behavior. However, these analyses are often constrained to small graph datasets \cite{xia2024anygraph} or focus solely on model scaling rather than data scaling effects \cite{wang2024learning}.

These observations raise a fundamental question: \textit{Do scaling laws inherently exist in graph learning?} If not, \textit{what requirements must be satisfied to achieve a true scaling law for GFMs?} In Section \ref{sec:open question scalability}, we explore these open questions and propose potential directions for future research.

\subsection{Transferability}

Transferability refers to a model's capability to extract patterns from source tasks and apply this knowledge to enhance performance on related target tasks \cite{bengio2012deep,jiang2022transferability}. The principles behind the transferability is that the pretrained models capture the general, transferable patterns across domains. For example, on textual data, the transferable patterns can be treated as tokens, words, and phrases; on image data, the transferable patterns can be treated as contours, colors, textures, and edges of an image. Understanding transferable patterns is essential for developing graph foundation models. We discuss the transferability of GFMs from the perspective of single-task and cross-task in the following, respectively.

\subsubsection{Single-Task Transferability}

\noindent\textbf{Node-Level Tasks.} Node-level tasks focus on understanding the properties of individual nodes within a graph, where these properties are typically influenced by the attributes of neighboring nodes. Depending on the nature of node interactions, connections can follow one of two fundamental principles: "similarity attracts" (homophily) or "opposite attracts" (heterophily). Specifically, homophily describes that connected nodes exhibit similar characteristics, whereas heterophily refers to that connected nodes possess dissimilar attributes. A primary challenge in ensuring the transferability of node-level tasks lies in designing a unified model capable of capturing both homophily and heterophily patterns. Standard GNN architectures struggle to generalize across both types of graphs, often exhibiting poor performance when applied jointly \cite{mao2023demystifying}. To address this issue, recent approaches incorporate additional textual descriptions to provide contextual information \cite{li2024zerog} or employ learnable prediction aggregation mechanisms \cite{zhao2025fully} to adaptively model different node interaction patterns.

\noindent\textbf{Link-Level Tasks.} Link-level tasks require models to capture relational structures between node pairs, often relying on proximity-based measures to determine the likelihood of a connection. More precisely, if two nodes share common neighborhoods, they are more likely to be linked. Depending on the extent of neighborhood overlap, proximity can be classified into two levels: (1) \textit{local proximity}, where nodes share direct neighbors, and (2) \textit{global proximity}, where nodes exhibit high-order neighborhood relationships. Effectively modeling these proximity patterns is challenging, as traditional message-passing GNNs lack the necessary expressiveness to capture link isomorphism \cite{zhang2021labeling}. To enhance relational modeling capabilities, advanced techniques such as labeling tricks \cite{zhang2021labeling} introduce additional structural knowledge or positional embeddings to enrich link representations, enabling improved expressiveness.

\noindent\textbf{Graph-Level Tasks.} Graph-level tasks involve learning representations that capture distinctive substructures known as graph motifs - small, recurring patterns that define structural properties within a graph. The complexity of graph-level transferability arises from that motif distributions vary across different graphs, requiring models to identify shared motifs across diverse motif sets, which potentially can be satisfied via disentangled representation learning \cite{tran2017disentangled} or invariance learning \cite{bardes2021vicreg}. However, another fundamental challenge lies in the expressiveness limitations of message-passing GNNs, which are inherently constrained by the 1-WL test \cite{zhang2024beyond}. Even if shared motif sets can be identified, standard GNN architectures may fail to encode them effectively. Addressing this limitation, expressive GNNs \cite{morris2019weisfeiler} have been proposed to enhance motif encoding capabilities. For a more detailed discussion, including network analysis, model expressiveness, and stability, we refer readers to a recent study \cite{mao2024graph} that analyzes these challenges in depth.

\subsubsection{Cross-Task Transferability}

\noindent\textbf{Graphon.} Graphon theory \cite{ruiz2020graphon, cao2023pre, sun2024fine} has been explored as a theoretical foundation for identifying transferable patterns in graphs. If two graphs are generated from the same graphon, they are expected to share similar topological properties, which can result in high transferability. Ruiz et al. \cite{ruiz2020graphon} established theoretical bounds on the embeddings of two graphs sampled from the same graphon, while Cao et al. \cite{cao2023pre} employed graphon theory to analyze transferability in pretraining and fine-tuning settings. By mapping pre-trained graphs into a graphon space, they ensured transferability if the target graph could be generated within this space. Extending this work, Sun et al. \cite{sun2024fine} proposed a fine-tuning strategy based on graphon theory. Despite its potential, the applicability of graphon theory to real-world graphs is constrained by its strong underlying assumptions \cite{levie2021transferability}. Additionally, even when these assumptions hold, identifying a shared graphon from a large set of cross-domain graphs remains a significant challenge, limiting the practical utility of graphon-based methods in GFM design.

\noindent\textbf{Substructures.} An alternative approach to defining transferable patterns involves leveraging recurring substructures such as triangles, stars, and $k$-cliques \cite{zhao2023gimlet, mao2024graph}. These motifs frequently appear across different graph domains but may carry different semantic meanings. For instance, triangles commonly occur in citation networks, social networks, and molecular graphs, albeit with domain-specific interpretations. Building on this observation, recent studies \cite{sun2023all, liu2024one} have proposed subgraph-based learning frameworks, where sampled subgraphs containing structures are encoded using GNNs for prediction. From a theoretical perspective, these methods leverage graph spectrum analysis to quantify transferability. Levie et al. \cite{levie2019transferability} analyzed transferability through stability, asserting that effective transfer should minimize sensitivity to small perturbations. Similarly, Levie et al. \cite{levie2021transferability} demonstrated that transferability is feasible when different graphs discretize the same underlying space. Zhu et al. \cite{zhu2021transfer} further reinforced this perspective by showing that higher similarity in ego-graph distributions correlates with better transferability.

\noindent\textbf{Tree Structures.} While substructures (motifs) provide a promising foundation for transferable patterns, they are not always learnable within GNNs. Traditional message-passing GNNs, constrained by the 1-WL test \cite{xu2018how, morris2019weisfeiler}, struggle to distinguish certain motifs \cite{garg2020generalization, chen2020can, zhang2024beyond}, such as stars, conjoint cycles, and $k$-cliques. To address this limitation, recent works have explored subtree-based representations to define transferable patterns. Wang et al. \cite{wang2024gft} were the first to propose subtree structures as fundamental transferable patterns, evaluating their effectiveness both empirically and theoretically. Building on this, Wang et al. \cite{wang2024learning} further established theoretical guarantees on the stability, transferability, and generalization of tree-based patterns. The key advantage of subtree-based methods is that message-passing GNNs can fully capture subtree structures \cite{xu2018how}. However, a major limitation is that tree-based representations inherently discard certain structural dependencies, potentially leading to information loss.

\newpage

\section{Dataset Resources}
\label{sec:resource}



Graph-structured data is prevalent across domains, leading to diverse benchmarks for evaluating graph learning methods, such as e-commerce, academic citation networks, knowledge bases, molecular science, temporal graphs, social network, brain map and images, which collectively cover various scales, features, and graph-based tasks.
\subsection{Tasks and Domains Overview}
As shown in Table~\ref{tab:datasets}, the datasets span various domains, characterized by distinct structural and task-related properties. This diversity enables comprehensive evaluation of graph learning methodologies across multiple application scenarios, providing insights into both generalizability and domain-specific performance characteristics.

\begin{table*}[!t]
    \centering
    \rowcolors{2}{shadecolor}{white}
    \caption{Overview of benchmark datasets used in graph learning research. The table categorizes datasets based on their domain, task type, and structural properties, including the number of nodes, edges, and classes. It also indicates whether datasets contain text attributes.}
    \label{tab:datasets}
    \resizebox{\linewidth}{!}{
        \begin{tabular}{lllrrrrcl}
            \toprule
            \textbf{Dataset}  & \textbf{Domain} & \textbf{Task} & \textbf{\# Nodes} & \textbf{\# Edges} & \textbf{\# Classes} & \textbf{\# Graphs} & \textbf{Text-Attributed} & \textbf{Source}          \\ \midrule
            Products          & E-commerce      & Node, Link    & 316,513           & 19,337,745        & 39                  & 1                  & Yes                      & \cite{chen2024text}      \\
            History           & E-commerce      & Node, Link    & 41,551            & 503,180           & 12                  & 1                  & Yes                      & \cite{chen2024text}      \\
            Children          & E-commerce      & Node, Link    & 76,875            & 2,325,044         & 24                  & 1                  & Yes                      & \cite{chen2024text}      \\
            Computer          & E-commerce      & Node, Link    & 87,229            & 1,256,548         & 10                  & 1                  & Yes                      & \cite{chen2024text}      \\
            Photo             & E-commerce      & Node, Link    & 48,362            & 873,793           & 12                  & 1                  & Yes                      & \cite{chen2024text}      \\
            Sportsfit         & E-commerce      & Node, Link    & 173,055           & 3,020,134         & 13                  & 1                  & Yes                      & \cite{chen2024text}      \\
            Ratings           & E-commerce      & Node, Link    & 24,492            & 186,100           & 5                   & 1                  & Yes                      & \cite{chen2024text}      \\ \midrule
            Arxiv             & Academia        & Node, Link    & 169,343           & 2,315,598         & 40                  & 1                  & Yes                      & \cite{chen2024text}      \\
            Cora              & Academia        & Node, Link    & 2,708             & 10,556            & 7                   & 1                  & Yes                      & \cite{chen2024text}      \\
            Citeseer          & Academia        & Node, Link    & 3,186             & 8,450             & 6                   & 1                  & Yes                      & \cite{chen2024text}      \\
            Pubmed            & Academia        & Node, Link    & 19,717            & 88,648            & 3                   & 1                  & Yes                      & \cite{chen2024text}      \\
            Arxiv 23          & Academia        & Node, Link    & 46,198            & 77,726            & 40                  & 1                  & Yes                      & \cite{chen2024text}      \\
            DBLP              & Academia        & Node, Link    & 14,376            & 431,326           & 4                   & 1                  & Yes                      & \cite{chen2024text}      \\ \midrule
            WN18RR            & knowledge Base  & Link          & 40,943            & 93,003            & 11                  & 1                  & Yes                      & \cite{galkin2023towards} \\
            FB15K237          & knowledge Base  & Link          & 14,541            & 310,116           & 237                 & 1                  & Yes                      & \cite{galkin2023towards} \\
            Codex Small       & knowledge Base  & Link          & 2,034             & 36,543            & 42                  & 1                  & Yes                      & \cite{galkin2023towards} \\
            Codex Median      & knowledge Base  & Link          & 17,050            & 206,205           & 51                  & 1                  & Yes                      & \cite{galkin2023towards} \\
            Codex Large       & knowledge Base  & Link          & 77,951            & 612,437           & 69                  & 1                  & Yes                      & \cite{galkin2023towards} \\
            NELL995           & knowledge Base  & Link          & 74,536            & 153,039           & 200                 & 1                  & Yes                      & \cite{galkin2023towards} \\
            GDELT             & knowledge Base  & Link          & 5,849             & 943,956           & 237                 & 1                  & Yes                      & \cite{zhang2024dtgb}     \\
            ICEWS1819         & knowledge Base  & Link          & 31,796            & 1,100,071         & 266                 & 1                  & Yes                      & \cite{zhang2024dtgb}     \\ \midrule
            Chemblpre         & Molecule        & Graph         & 8,845,648         & 19,123,034        & 1,295               & 341,952            & Yes                      & \cite{feng2024taglas}    \\
            PCBA              & Molecule        & Graph         & 11,349,235        & 24,566,048        & 128                 & 437,092            & Yes                      & \cite{feng2024taglas}    \\
            HIV               & Molecule        & Graph         & 1,049,163         & 2,259,376         & 1                   & 41,127             & Yes                      & \cite{feng2024taglas}    \\
            BBBP              & Molecule        & Graph         & 49,068            & 105,842           & 1                   & 2,039              & Yes                      & \cite{feng2024taglas}    \\
            BACE              & Molecule        & Graph         & 51,577            & 111,536           & 1                   & 1,513              & Yes                      & \cite{feng2024taglas}    \\
            TOXCAST           & Molecule        & Graph         & 161,002           & 330,180           & 588                 & 8,575              & Yes                      & \cite{feng2024taglas}    \\
            CYP450            & Molecule        & Graph         & 414,367           & 895,886           & 5                   & 16,896             & Yes                      & \cite{feng2024taglas}    \\
            TOX21             & Molecule        & Graph         & 145,459           & 302,190           & 12                  & 7,831              & Yes                      & \cite{feng2024taglas}    \\
            MUV               & Molecule        & Graph         & 2,255,846         & 4,892,252         & 17                  & 93,087             & Yes                      & \cite{feng2024taglas}    \\
            Alkane-Carbonyl   & Molecule        & Graph         & $\sim$41          & $\sim$43          & 2                   & 4,326              & No                       & \cite{dai2024large}   \\
            ZINC              & Molecule        & Graph         & $\sim$20          & $\sim$22          & 2                   & 12,000             & No                       & \cite{dai2024large}   \\
            Fluoride-Carbonyl & Molecule        & Graph         & 8,671             & $\sim$23          & 2                   & 8,671              & No                       & \cite{dai2024large}   \\
            ENZYMES           & Molecule        & Graph         & $\sim$33          & $\sim$64          & 6                   & 600                & No                       & \cite{dai2024large}   \\
            MUTAG             & Molecule        & Graph         & $\sim$16          & $\sim$17          & 2                   & 188                & No                       & \cite{dai2024large}   \\
            ToxCast           & Molecule        & Graph         & $\sim$19          & $\sim$39          & 588                 & 8,575              & Yes                      & \cite{ToxCast2024}       \\ \midrule
            Enron             & Temporal        & Link          & 42,712            & 797,907           & 10                  & 1                  & Yes                      & \cite{zhang2024dtgb}     \\
            Googlemap CT      & Temporal        & Link          & 111,169           & 1,380,623         & 5                   & 1                  & Yes                      & \cite{zhang2024dtgb}     \\ \midrule
            IMDB-BINARY       & Social Network  & Graph         & $\sim$20          & $\sim$96          & 2                   & 1,000              & No                       & \cite{dai2024large}   \\
            IMDB-MULTI        & Social Network  & Graph         & $\sim$13          & $\sim$65          & 3                   & 1,500              & No                       & \cite{dai2024large}   \\
            MovieLens         & Social Network  & Link          & 2,625             & 100,000           & 5                   & 1                  & Yes                      & \cite{lin2024langgfm}       \\
            Flickr            & Social Network  & Node          & 89,250            & 899,756           & 7                   & 1                  & Yes                      & \cite{xu2024graphfm}       \\
            Reddit            & Social Network  & Node          & 232,965           & 11,606,919        & 41                  & 1                  & Yes                      & \cite{xu2024graphfm}       \\ \midrule
            ogbl-vessel       & Brain Graph     & Link          & 3,538,495         & 5,345,897         & 2                   & 1                  & No                       & \cite{hu2020open}  \\ \midrule
            Fingerprint       & Images          & Graph         & 2,149             & $\sim$2           & 3                   & 2,149              & No                       & \cite{dai2024large}   \\
            \bottomrule
        \end{tabular}
    }
\end{table*}

\subsubsection{Tasks}

\noindent\textbf{Node Classification.} Node classification involves predicting labels for nodes within a graph, such as categorizing academic papers based on research areas or classifying products into specific categories. The task necessitates effective encoding of both node-specific features and neighborhood structural information to achieve optimal performance.

\noindent\textbf{Link Prediction.} Link prediction aims to forecast whether an edge exists between pairs of nodes or predict future edges in evolving networks. It is critical in recommendation systems and knowledge base completion. The fundamental challenge lies in effectively modeling node similarity based on both structural proximity and feature compatibility, often requiring sophisticated embedding techniques that capture higher-order connectivity patterns.

\noindent\textbf{Graph Classification.} Graph classification tasks predict labels for entire graphs, widely applied in molecular property prediction and social network analysis. This paradigm necessitates the development of effective graph-level representations that preserve both local substructure information and global topological characteristics. Graph classification frameworks typically implement hierarchical pooling mechanisms to progressively coarsen graph representations, analogous to the pooling operations in convolutional neural networks for computer vision tasks. The efficacy of such approaches often depends on their ability to identify discriminative subgraphs within the broader graph structure.

\subsubsection{Domains}

\noindent\textbf{E-commerce.} Datasets from e-commerce platforms typically feature large-scale product graphs, where nodes represent products and edges represent user interactions or product similarities. These graphs frequently exhibit complex heterogeneous structures with multiple edge types representing diverse interaction modalities such as co-purchasing, co-viewing, and semantic similarity.

\noindent\textbf{Academia.} Academic datasets model citation networks, where nodes represent papers and edges represent citation relationships, enabling research area classification and citation prediction tasks. Such networks typically manifest temporal evolution characteristics, as citations accumulate over time and research trajectories evolve, presenting opportunities for dynamic graph modeling approaches.

\noindent\textbf{Knowledge Bases.} Knowledge base datasets comprise structured entities and their interrelations, predominantly employed in link prediction tasks to infer missing relationships or validate entity links. These datasets often incorporate ontological constraints and hierarchical structures that introduce inductive biases beneficial for reasoning tasks. 

\noindent\textbf{Molecular Science.} Molecular datasets consist of numerous small graphs representing molecular structures, primarily used in property prediction, drug discovery, and biochemical analysis tasks. These graphs exhibit strong regularity in node degree distributions and edge formations, reflecting the physical constraints of chemical bonding principles.

\noindent\textbf{Temporal Graphs.} Temporal datasets feature dynamic graphs evolving over time, suitable for tasks such as temporal link prediction and anomaly detection in evolving networks. These datasets capture longitudinal structural transitions, enabling the modeling of evolutionary patterns and temporal dependencies in graph structures.

\noindent\textbf{Social Networks.} Social network datasets represent interactions among users, facilitating tasks like community detection, influence prediction, and content recommendation. These networks typically exhibit distinctive properties such as high clustering coefficients, small-world phenomena, and scale-free degree distributions that influence algorithm design considerations.

\noindent\textbf{Brain Graphs.} Brain graph datasets model neuronal connectivity or vascular structures, supporting tasks such as neurological disorder diagnosis and anatomical studies. These graphs present unique challenges due to their inherent multi-scale organization, from microscopic neuronal circuits to macroscopic brain regions connected through white matter tracts.

\noindent\textbf{Images.} Image-based graph datasets represent visual structures such as fingerprints, enabling tasks related to visual pattern recognition and image classification. These graphs typically encode spatial relationships between visual elements, transforming grid-structured image data into irregular graph structures that capture meaningful object-part relationships.

\subsection{Benchmark Descriptions}

Recent graph learning research has introduced diverse benchmarks across multiple domains. \textit{Text-space Graph Foundation Models}~\cite{chen2024text} provides 13 text-attributed benchmarks ranging from large-scale e-commerce networks (Products: 316K nodes) to smaller academic graphs (Cora), focusing on node and link prediction. \textit{Knowledge Base Benchmarks}~\cite{galkin2023towards} offers text-enhanced versions of standard knowledge graphs (WN18RR, FB15K237) for semantic link prediction tasks. \textit{Temporal Graph Benchmarks}~\cite{zhang2024dtgb} includes evolving datasets like ICEWS1819 for temporal link prediction in dynamic structures. \textit{TAGLAS}~\cite{feng2024taglas} introduces text-attributed molecular datasets (Chemblpre, PCBA) with hundreds of thousands of graphs for property prediction in drug discovery. \textit{Graph Pattern Recognition}~\cite{dai2024large} presents structural-only benchmarks (ENZYMES, MUTAG) to evaluate topology comprehension without text attributes. \textit{Special-Purpose Benchmarks} cover domain-specific tasks including brain connectivity (ogbl-vessel~\cite{hu2020open}), recommendation systems (MovieLens~\cite{lin2024langgfm}), and social networks (Reddit, Flickr~\cite{xu2024graphfm}).

\newpage

\section{Open Questions}
\label{sec:open question}

\subsection{How to Enhance Scalability?}
\label{sec:open question scalability}

LLMs achieve scalability through the scaling law, where larger models and more training data lead to significant performance improvements. However, such a trend has not yet to emerge in existing GFMs. To establish a true scaling law on graphs, we highlight three key aspects.

\noindent\textbf{Better Graph Backbones.} GNNs suffer from intrinsic limitations, including over-smoothing \cite{li2019deepgcns}, over-squashing \cite{alon2021on}, inadequate long-range dependency modeling \cite{dwivedi2022long}, and limited expressiveness in capturing graph substructures \cite{xu2018how}. These limitations hinder both model scalability \cite{morris2024position} and efficient multi-GPU training \cite{cai2021dgcl}. To overcome these challenges, designing a more scalable graph backbone is imperative. Inspired by the success of transformer architectures in NLP and CV foundation models \cite{brown2020language,dosovitskiy2021an}, transformer-based architectures have emerged as potential candidates for graph learning \cite{dwivedi2020generalization,rampasek2022recipe}. Basic graph transformers tokenize graphs into node sequences, but their applicability is constrained to small-scale graphs due to quadratic complexity \cite{wu2023simplifying}. Recent advancements propose leveraging substructures as graph tokens \cite{wang2025gpm}, where sequences of substructure patterns serve as representations for nodes, edges, and entire graphs. This method significantly reduces encoding complexity and improves scalability \cite{wang2025gpm}, making it a promising direction for future GFMs.

\noindent\textbf{Better Pretraining Objectives.} A well-designed pretraining objective is essential for extracting transferable knowledge from large-scale datasets. In NLP and vision, foundation models predominantly employ generative pretraining, such as next-token prediction \cite{brown2020language}, to capture meaningful semantics. In contrast, most graph self-supervised learning approaches rely on contrastive pretraining, which has shown limited effectiveness compared to generative pretraining in other domains \cite{he2022masked}. While some studies have explored reconstruction-based objectives for graphs \cite{hou2022graphmae}, they primarily focus on low-level semantics, such as nodes and edges. This differs from NLP and CV, where models reconstruct word tokens and image patches, preserving high-level semantics \cite{assran2023self}. As a result, existing generative graph methods fail to outperform contrastive learning. To achieve meaningful pretraining for GFMs, it is crucial to shift toward reconstructing high-level semantic structures.

\noindent\textbf{Better Learning Instances.} In LLMs, sentences composed of word tokens serve as basic learning instances, while in VLMs, images consisting of visual patches act as primary learning instances \cite{wang2024gft}. Training from these instances allows foundation models to acquire transferable and scalable knowledge. However, it remains unclear which learning instances—nodes, edges, or entire graphs—should be scaled in GFMs \cite{mao2024graph}. Moreover, different graph-based tasks rely on distinct learning instances; node-level tasks focus on nodes, whereas graph-level tasks emphasize graphs. To align it across tasks, unified learning instances such as subgraphs \cite{sun2023all} and trees \cite{wang2024gft} have been proposed. However, a critical question remains: can scaling these unified learning instances facilitate the acquisition of cross-task knowledge?

\subsection{How to Mitigating Data Scarcity?}

The effectiveness of existing foundation models, such as LLMs and LVMs, is largely attributed to their data-driven learning paradigm. Unlike textual and image data, which are readily accessible from online sources, obtaining graph datasets presents a significant challenge. To address the issue of data scarcity in graph learning, we outline three promising directions.

\noindent\textbf{Automated Graph Collection Pipelines.} The success of LLMs and LVMs has been facilitated by extensive datasets curated through automated web scraping techniques. Unlike text or images, which can be extracted from publicly available sources, graph construction often requires explicit human curation, as graphs encode semantic and domain-specific relationships. Recent advances in LLMs suggest that automated dataset construction \cite{crawl4ai} could serve as a viable solution for acquiring graph data. Leveraging such techniques, it is possible to systematically extract structured relationships from diverse sources, including academic repositories, biomedical databases, and online knowledge graphs.

\noindent\textbf{Synthetic Data Generation.}
Beyond directly collecting new datasets, data augmentation and synthesis techniques have been widely adopted in other domains to mitigate data scarcity. In LLMs, knowledge distillation and data synthesis \cite{long2024llms} have emerged as prominent strategies for enhancing model generalization. Similarly, in graph learning, recent advances in generative models, such as graph diffusion models \cite{vignac2022digress}, have enabled the augmentation of graph structures with synthetic data. For example, diffusion models have been employed to enhance structural diversity in graph datasets \cite{tang2024cross}, while LLMs have been utilized to generate synthetic text attributes for text-attributed graphs \cite{wang2025can}. These techniques provide a pathway to augmenting existing datasets with novel graph instances. Developing robust graph generation techniques that preserve the structural and semantic integrity of real-world graphs remains an important direction for future research.

\noindent\textbf{Acquiring High-Quality Graph Data.}
While increasing dataset size is a common strategy for improving model performance, recent studies in LLM pretraining \cite{lin2024rho} and instruction tuning \cite{xia2024less} have demonstrated that high-quality data can be more valuable than sheer data volume. In other words, models trained on a small set of high-quality data can achieve performance comparable to or even superior to those trained on large-scale low-quality datasets. Applying this principle to graph learning suggests that curating high-quality graph datasets may yield greater benefits than simply scaling dataset size. However, graph datasets are often incomplete, and evaluating their quality is inherently challenging. Although various data valuation techniques \cite{winter2002shapley,jia2019towards} have been proposed to assess data contributions in machine learning, defining quality metrics for graph data remains an open problem. The quality of graph data is highly dependent on the chosen backbone architecture and pretraining strategy, necessitating further research into effective graph data valuation methodologies.

\subsection{How to Better Evaluate GFMs?}

Evaluating the effectiveness of GFMs is crucial. However, due to their broad applicability across diverse domains, traditional evaluation approaches using small-scale benchmarks are often insufficient. In this section, we discuss two aspects necessary for advancing the evaluation framework.

\noindent\textbf{Developing Advanced Benchmarks.} Assessing the power of foundation models requires well-designed benchmarks. However, existing graph benchmarks suffer from several limitations: (1) they often lack transformative real-world applications, (2) they are constructed in ways that do not meaningfully reflect practical use cases, and (3) the benchmarking culture in the graph community has been criticized for its inconsistencies and reproducibility issues \cite{bechler2025position}. Given these challenges, constructing high-quality, large-scale benchmarks that align with real-world scenarios is essential for evaluating GFMs effectively. Such benchmarks should encompass diverse graph structures, multiple learning tasks, and varying levels of supervision to provide a comprehensive assessment of model capabilities.

\noindent\textbf{Beyond Accuracy: Evaluating Generalization, Robustness, and Trustworthiness.} Traditional evaluation metrics, such as accuracy, are insufficient to capture the full potential of GFMs. Beyond raw performance on benchmark datasets, it is critical to assess their generalization ability across different domains, their robustness against adversarial and noisy data, and their trustworthiness in high-stakes applications. Developing novel evaluation metrics and benchmarks that explicitly test these dimensions is necessary for a more holistic understanding of GFM capabilities. Future work should explore methodologies for measuring domain adaptation, reliability under distribution shifts, interpretability, and ethical considerations in graph-based decision-making.

\subsection{How to Better Utilize GFMs?}

Effectively leveraging GFMs is crucial for maximizing their impact across various domains. In this section, we discuss three key aspects that can enhance the adaptability, applicability, and multimodal integration.

\noindent\textbf{Advanced Adaptation Methods.} LLMs have demonstrated remarkable capabilities in in-context learning and zero-shot generalization, allowing models to adapt to new tasks without fine-tuning. While recent advancements in graph prompt learning \cite{sun2023graph} have introduced similar adaptation techniques for GFMs, these approaches still require fine-tuning of prompt tokens to align with downstream tasks. An alternative direction follows the paradigm of VLMs, where LLMs are used to reason over structured graph data \cite{liu2024can}. However, efficiently incorporating graph knowledge into this framework remains a significant challenge. This raises an important question: can we develop methods that enable seamless adaptation of GFMs without fine-tuning? Recently, large visual models (LVMs) \cite{bai2024sequential} have demonstrated the feasibility of handling diverse visual tasks in an autoregressive manner without relying on language-based interfaces. Extending this concept to graphs could enable GFMs to perform true zero-shot and in-context learning, eliminating the need for explicit task-specific adaptations.

\noindent\textbf{Killing Applications.} LLMs have demonstrated their capability to handle complex, domain-specific problems. While GFMs have shown success in applications such as social network analysis, drug property prediction, and recommender systems, these tasks can often be addressed effectively using traditional graph learning models, reducing the necessity for a GFM. Thus, it is essential to identify high-impact applications where simple graph learning approaches fall short. These applications should be sufficiently complex and demand capabilities that only GFMs can provide. Potential directions include: \textit{Chip design}: Optimizing circuit layouts through learned structural representations \cite{mirhoseini2021graph}. \textit{Combinatorial optimization}: Addressing NP-hard graph problems using scalable GFM-based solutions \cite{wenkel2024towards}. \textit{Relational databases}: Enhancing query optimization and knowledge extraction from structured database systems \cite{fey2024position}.

\noindent\textbf{Integrating Multimodal Knowledge.} Graph data inherently encapsulates structured knowledge spanning multiple modalities. For instance, molecular data can be represented as graphs, sequences, textual descriptions, or even photographic images; social networks involve diverse data types, including user names, job positions, profile images, structured activity logs, and historical interactions. Designing models that effectively integrate and process multimodal graph information remains a challenge. One crucial question is whether a unified model should be designed to handle different modalities collectively, or if modality-specific models should be employed to leverage their unique strengths.

\noindent\textbf{Human-in-the-Loop.} While GFMs offer strong generalization, many real-world scenarios---such as drug discovery, recommendation, and scientific workflows---require iterative human feedback or domain-specific supervision. Integrating human-in-the-loop mechanisms enables model correction, prompt refinement, and interactive adaptation, improving alignment with expert knowledge and practical needs. As GFMs scale toward broader deployment, human involvement will be essential for enhancing interpretability, control, and real-world reliability.

\subsection{Advanced Theoretical Understandings}

A deeper theoretical understanding of GFMs is essential for improving their effectiveness, reliability, and generalizability. In this section, we discuss three areas of theoretical investigation that remain open challenges.

\noindent\textbf{Transferability.} Graphs encode complex, often manually defined relationships, making their transferable knowledge less intuitive compared to text or images. While some empirical studies suggest the possibility of transferability even across seemingly unrelated domains, there remains a lack of both theoretical and intuitive explanations for such phenomena. For instance, it is difficult to conceptualize shared transferable patterns between social networks and molecular graphs due to the stark differences in their structural distributions. Recent works \cite{wang2024gft,wang2024learning} have attempted to characterize transferability between graphs from different domains by analyzing subtree distributions. While these studies provide valuable insights, treating graphs solely as compositions of trees inevitably results in the loss of structural information \cite{xu2018how}. A comprehensive theoretical framework is needed to define and quantify the transferable knowledge within and across graph domains, paving the way for more effective transfer learning strategies in GFMs.

\noindent\textbf{Pattern Conflict Issue.} It is even more challenging in identifying cross-task transferable patterns in GFMs due to the pattern conflicts. This issue arises when the same structural pattern carries different semantic meanings across diverse domains, leading to potential inconsistencies in learned representations. Consider a pretraining scenario where the model is trained on datasets spanning multiple domains, such as social networks and molecular networks. Suppose the model learns to recognize and leverage triangle structures during pretraining. In social networks, triangles often represent \textit{stability}, following the principle of "the friend of my friend is my friend." However, in molecular graphs, triangular patterns may indicate \textit{instability} due to specific chemical constraints. This fundamental discrepancy in interpretation can severely degrade model performance \cite{cao2023pre,mao2024graph}. Existing methods \cite{wang2024gft,liu2024one} either fail to address or only partially mitigate \cite{wang2024learning} the pattern conflict issue. Developing strategies to effectively resolve this challenge is crucial for constructing truly generalizable GFMs.

\noindent\textbf{Robustness and Trustworthiness.} Real-world graphs exhibit various undesirable properties, including long-tail distributions, incompleteness, class imbalances, limited labeled data, and structural alterations. To develop robust and trustworthy GFMs, it is essential to understand how these models respond to such challenges. A promising direction involves analyzing the stability of GFMs under structural distribution shifts \cite{wang2024learning}. This includes studying their resilience to adversarial perturbations, handling missing or noisy data, and ensuring fairness in decision-making. Establishing theoretical guarantees for robustness will be crucial for deploying GFMs in high-stakes applications such as healthcare, finance, and cybersecurity.

\noindent\textbf{Generalization.} Balancing model fitting and generalization is fundamental in machine learning. Overemphasizing fitting capacity can lead to overfitting on specific datasets, impairing performance on unseen graphs, whereas excessive focus on generalization may compromise predictive accuracy on in-distribution tasks. Understanding the generalization of GFMs is crucial to optimizing this trade-off. A pioneering work \cite{wang2024learning} established generalization bounds for GFMs using subtree-based learning tokens. However, a broader generalization analysis that extends beyond tree structures is necessary to derive more applicable insights. 

\newpage

\section{Conclusion}
\label{sec:conclusion}

\textit{Graph Foundation Models} represent a transformative paradigm in graph machine learning, aspiring to replicate the success of foundation models in natural language processing and computer vision within the structured domain of graphs. In this survey, we present a comprehensive and systematic review of the emerging landscape of GFMs. We begin by contextualizing their development, outlining the fundamental challenges that arise from graph heterogeneity, non-Euclidean structure, and cross-domain transferability. To unify the diverse body of work, we propose a general framework that decomposes GFMs into modular components—encompassing backbone architectures, pretraining strategies, and adaptation mechanisms.

We categorize GFMs into three major classes: \textit{universal GFMs}, which aim for broad generalization across tasks and domains; \textit{task-specific GFMs}, which prioritize performance on focused objectives like link prediction or node classification; and \textit{domain-specific GFMs}, which target specialized applications such as molecules, knowledge graphs, and computational graphs. For each category, we analyze core design principles, review representative methods, and conduct comparative analyses to highlight their relative strengths and limitations. Beyond empirical trends, we further investigate the theoretical underpinnings of GFMs, offering insights into expressiveness, transferability, and generalization guarantees. Despite notable progress, the field faces several open challenges. These include the scalability of GFMs to massive graphs; the integration of multimodal signals; the development of principled evaluation protocols; and the formulation of theoretical foundations that explain the transferability and generalization.

Looking ahead, we envision GFMs as foundational infrastructure for general-purpose graph intelligence. Future research should focus on building more scalable, interpretable, and adaptable architectures, expanding graph pretraining corpora across real-world domains, and advancing theoretical frameworks that explain their behavior. By bridging structural inductive biases, GFMs hold immense promise for enabling new potentials in scientific discovery, industrial systems, and decision-making over structured data.

\newpage


\bibliographystyle{unsrt}
\bibliography{citation}

\newpage
\appendix

\end{document}